\documentclass{article} 

\usepackage{times}

\usepackage{booktabs}
\usepackage{graphicx,caption}
\usepackage{authblk}

\usepackage{amsmath,amsthm,amsfonts,amssymb,amscd,pifont}
\usepackage{verbatim}
\usepackage{url}
\usepackage{natbib}

\usepackage{makecell}

\usepackage{mathrsfs}
\usepackage[ansinew]{inputenc}
\usepackage{graphicx}
\usepackage{a4wide}
\usepackage{subfigure}
\usepackage{graphics}

\usepackage{tabularx}
\usepackage{booktabs}

\usepackage{graphicx}
\usepackage{hyperref}
\usepackage{xfrac}
\usepackage{array}
\usepackage[all]{xy}
\usepackage{epic}
\usepackage{latexsym}
\usepackage{enumerate}
\usepackage{appendix}
\newtheorem{theorem}{Theorem}

\newtheorem*{Theorem*}{Theorem}

\newtheorem{maintheorem}{Theorem}

\newcommand{\cmt}{\begin{maintheorem}}
\newcommand{\fmt}{\end{maintheorem}}

\newtheorem{maincorollary}[maintheorem]{Corollary}

\newcommand{\cmc}{\begin{maincorollary}}
\newcommand{\fmc}{\end{maincorollary}}

\newtheorem{T}{Theorem}[section]
\newcommand{\ct}{\begin{T}}
\newcommand{\ft}{\end{T}}
\newtheorem{D}{Definition}[section]
\newcommand{\cd}{\begin{D}}
\newcommand{\fd}{\end{D}}

\newtheorem{Corollary}[T]{Corollary}
\newcommand{\cco}{\begin{Corollary}}
\newcommand{\fco}{\end{Corollary}}

\newtheorem{Proposition}[T]{Proposition}
\newcommand{\cpr}{\begin{Proposition}}
\newcommand{\fpr}{\end{Proposition}}

\newtheorem{Lemma}[T]{Lemma}
\newcommand{\cle}{\begin{Lemma}}
\newcommand{\fle}{\end{Lemma}}

\newtheorem{Sublemma}[T]{Sublemma}
\newcommand{\csle}{\begin{Sublemma}}
\newcommand{\fsle}{\end{Sublemma}}

\usepackage{graphicx}
\usepackage{xcolor}
\usepackage{mathtools}
\usepackage{mathrsfs}

\newcommand{\be}{\begin{eqnarray}}
\newcommand{\ee}{\end{eqnarray}}

\usepackage{caption}  
\usepackage{textcomp} 
\usepackage{multirow}  
\usepackage{amsmath}

\usepackage{tikz}
\usetikzlibrary{positioning, shapes.geometric}

\usepackage{pgfplots}
\usepackage{pgfplotstable}
\usepackage{tikz}
\usetikzlibrary{shapes.geometric, arrows.meta, positioning}

\usepackage{pgfplots}
\usepgfplotslibrary{polar}
\pgfplotsset{compat=1.18}

\usepackage{hyperref}
\usepackage{url}

\title{Enhancing Neural Function Approximation: The XNet Outperforming KAN}

\author[1,2]{Xin Li\thanks{Email: \texttt{xinli2023@u.northwestern.edu}}}
\author[3]{Xiaotao Zheng\thanks{Email: \texttt{20234013002@stu.suda.edu.cn}}}
\author[4,5]{Zhihong Xia\thanks{Corresponding author: \texttt{xia@math.northwestern.edu}}}

\affil[1]{Oxford Suzhou Centre for Advanced Research, Suzhou, China}
\affil[2]{Department of Computer Science, Northwestern University, Evanston, IL, USA}
\affil[3]{Center for Financial Engineering, Soochow University,Suzhou, China}
\affil[4]{School of Natural Science, Great Bay University, Guangdong, China}
\affil[5]{Department of Mathematics, Northwestern University, Evanston, IL, USA}

\date{}
\begin{document}

\maketitle

\begin{abstract}
XNet is a single-layer neural network architecture that leverages Cauchy integral-based activation functions for high-order function approximation. Through theoretical analysis, we show that the Cauchy activation functions used in XNet can achieve arbitrary-order polynomial convergence, fundamentally outperforming traditional MLPs and Kolmogorov-Arnold Networks (KANs) that rely on increased depth or B-spline activations. Our extensive experiments on function approximation, PDE solving, and reinforcement learning demonstrate XNet's superior performance - reducing approximation error by up to 50000 times and accelerating training by up to 10 times compared to existing approaches. These results establish XNet as a highly efficient architecture for both scientific computing and AI applications.
\end{abstract}


\vskip 0.3in

\section{Introduction}
A novel method for constructing real networks from the complex domain using the Cauchy integral formula was introduced in \cite{LXZ24}, utilizing Cauchy kernels as basis functions. This approach enables a significant extension of traditional neural network architectures. This work comprehensively compares these networks with Kolmogorov-Arnold Networks (KANs), which use B-splines as basis functions \cite{liu2024kan}, and Multi-layer Perceptrons (MLPs), highlighting substantial improvements in various tasks.

Multi-layer perceptrons (MLPs) (\cite{Haykin94, Cybenko89, Hornik89}),
recognized as fundamental building blocks in deep learning, have their
limitations despite their wide use, particularly in their accuracy, and
the large number of parameters needed in structures such as transformers
(\cite{Vaswani17}), as well as their lack of interpretability without post-analysis
tools (\cite{Cunningham23}). Kolmogorov-Arnold Networks (KANs) were
introduced as a potential alternative, drawing on the Kolmogorov-Arnold
representation theorem (\cite{Kolmogorov56, Braun09}), and have demonstrated
efficiency and accuracy in computational tasks, especially in
solving PDEs and function approximation (\cite{Sprecher02,Koppen02,Lin93,Lai21,Leni13,Fakhoury22}).

In the swiftly advancing domain of deep learning, the continuous
search for novel neural network designs that deliver superior accuracy
and efficiency is pivotal. While traditional activation functions such
as the Rectified Linear Unit (ReLU) (\cite{Hinton2010}) have been widely
adopted due to their straightforwardness and efficacy in diverse
applications, their shortcomings become evident as the complexity of
challenges escalates. This is particularly true in areas that demand
meticulous data fitting and the solutions of intricate partial
differential equations (PDEs). These limitations have paved the way
for architectures that merge neural network techniques with PDEs,
significantly enhancing function approximation capabilities in
high-dimensional settings (\cite{Sirignano2018, Raissi2019, Jin2021,
  Transolver2024, PINNsFormer2023}).

Reinforcement learning (RL) \cite{sutton2018reinforcement} is widely applied in robotics \cite{nguyen2019review}, healthcare \cite{yu2021reinforcement}, and gaming \cite{hafner2023mastering}. Proximal Policy Optimization (PPO) \cite{schulman2017proximal} is a standard approach balancing exploration and exploitation. KAN models have also been explored in RL, demonstrating promising results \cite{kich2024kolmogorov, genet2024tkan}. This study systematically compares PPO implementations using MLP, KAN, and XNet, revealing XNet’s advantages in handling complex decision-making tasks.

Inspired by the mathematical precision of the Cauchy integral theorem,
\cite{LXZ24} introduced the XNet architecture, a novel neural network
model that incorporates a uniquely designed Cauchy activation
function. This function is mathematically expressed as:

$$\phi_a(x) = \frac{\lambda_1 * x}{x^2+d^2}+
\frac{\lambda_2}{x^2+d^2},$$ where $\lambda_1$, $\lambda_2$, and $d$
are parameters optimized during training. This design is not only a
theoretical advancement but also empirically advantageous, offering a
promising alternative to traditional models for many applications. By
integrating Cauchy activation functions, XNet demonstrates superior
performance in function approximation tasks and in solving
low-dimensional PDEs compared to its contemporaries, namely Multilayer
Perceptrons (MLPs) and Kolmogorov-Arnold Networks (KANs). This paper
systematically compares these architectures, highlighting XNet's
advantages in terms of accuracy, convergence speed, and computational
demands.

Furthermore, empirical evaluations reveal that the Cauchy activation
function possesses a localized response with decay at both ends,
significantly benefiting the approximation of localized data
segments. This capability allows XNet to fine-tune responses to
specific data characteristics, a critical advantage over the globally
responding functions like ReLU.

The implications of this research are significant. It has been
demonstrated that XNet can serve as an effective foundation for
general AI applications, and findings indicate that it can even
outperform meticulously designed networks tailored for specific
purposes.

Table~\ref{tab:arch_comparison} summarizes the key differences among XNet, MLPs, and KANs. Notably, XNet maintains high accuracy with minimal computational overhead, providing a practical alternative to deep networks.

\begin{table*}[t]
\caption{Comparison of Neural Network Architectures}
\label{tab:arch_comparison}
\begin{center}
\small
\begin{tabular}{lccc}
\toprule
Architecture & Activation & Approx. Rate & Structure \\
\midrule
MLP  & Fixed (ReLU, sigmoid) & $O(1/N)$  & Deep, many parameters \\
KAN  & Learnable B-splines   & $O(N^{-k})$  & Hierarchical, structured \\
XNet & Cauchy kernels  & $O(N^{-p})$, any $p>0$ & \textbf{Single-layer, compact} \\
\bottomrule
\end{tabular}
\end{center}
\end{table*}

\textbf{Principal Contributions}

Our study advances the field of neural network architectures through several key contributions:

$(i)$ \textbf{Enhanced Function Approximation Capabilities:} Through comprehensive comparative analysis, we demonstrate XNet's superior performance over KAN in function approximation, particularly excelling in handling the Heaviside step function, complex high-dimensional scenarios, and noisy data. Sections \ref{sec:1d} through \ref{sec:functions_noisy} provide detailed empirical validations of these capabilities.

$(ii)$ \textbf{Superiority in Physics-Informed Neural Networks:} Using the Poisson equation and Heat equation as a benchmark, we establish XNet's enhanced efficacy within the PINN framework. Our results, detailed in Section \ref{sec:application}, demonstrate significant performance improvements over both MLPs and KANs, setting new standards for accuracy and efficiency in PDE solving.

(iii) \textbf{Enhanced Reinforcement Learning Performance}: The effectiveness of XNet in continuous control tasks is demonstrated in Section \ref{online-RF}. Our experiments on DeepMind Control Suite benchmarks show that XNet significantly outperforms traditional architectures, achieving superior scores on HalfCheetah-v4 (3298.52) and Swimmer-v4 (100.38) environments, marking substantial improvements over both MLP-based and KAN-based approaches.

These contributions collectively establish XNet as a versatile and powerful architecture, capable of advancing performance across multiple domains in machine learning and artificial intelligence. The consistent superior results across diverse applications demonstrate XNet's potential to significantly impact future developments in the field.

Table~\ref{tab:model_comparison} demonstrates the superior performance of XNet architectures compared to traditional neural networks in solving the Heat equations within the PINNs framework. The error reduction and time reduction metrics highlight XNet's advantages in both accuracy and computational efficiency.

\begin{table}
\caption{Performance Comparison of Neural Architectures for Solving Heat Equations via PINNs}
\label{tab:model_comparison}
\begin{center}
\resizebox{0.75\textwidth}{!}{ 
\begin{tabular}{lcccc}
\toprule
Architecture & MSE & Training Time (s) & Error Reduction$^*$ & Time Speedup$^\dagger$ \\
\midrule
MLP [2, 20, 1] & \underline{2.01 $\times$10$^{-4}$} & 21.2 & 1$\times$ & 18$\times$ \\
MLP [2, 200, 1] & 1.16 $\times$10$^{-4}$ & 37.6 & 2$\times$ & 10$\times$ \\
MLP [2, 20, 20, 1] & 2.45 $\times$10$^{-5}$ & 43.8 & 8$\times$ & 9$\times$ \\
MLP [2, 200, 200, 1] & 7.89 $\times$10$^{-6}$ & 108.7 & 25$\times$ & 4$\times$ \\
KAN [2, 10, 1] & 1.51 $\times$10$^{-7}$ & 254.6 & 1,331$\times$ & 2$\times$ \\
KAN [2, 20, 1] & 4.53 $\times$10$^{-8}$ & \underline{384.2} & 4,436$\times$ & 1$\times$ \\
XNet [2, 20, 1] & 3.89 $\times$10$^{-8}$ & 43.5 & 5,166$\times$ & 9$\times$ \\
XNet [2, 200, 1] & \textbf{3.69 $\times$10$^{-9}$} & 108.3 & \textbf{54,493}$\times$ & 4$\times$ \\
\bottomrule
\end{tabular}
} 
\end{center}
\footnotesize{$^*$Error Reduction = MSE$_{\text{MLP [2, 20, 1]}}$/MSE$_{\text{model}}$. $^\dagger$Time Speedup = Time$_{\text{KAN [2, 20, 1]}}$/Time$_{\text{model}}$, higher is better.}
\vspace{1mm}
\small{Note: Bold and underlined values indicate the best and second-best performance respectively.}
\end{table}

\section{Theoretical Comparison Between Cauchy Kernels and B-splines}

\label{sec:compare_activation_functions}

\begin{theorem}{Cauchy Approximation Theorem (from \cite{LXZ24}).} \label{Cauchy_approximation_theorem}
Let $f(z^1,\dots,z^d)$ be an analytic function on an open set \( U \subset \mathbb{C}^d \). It was shown in \cite{LXZ24} that for any desired accuracy \( \varepsilon > 0 \), $f$ can be approximated in the \(L^\infty\)-norm using a finite sum of Cauchy kernels:
\begin{equation}\label{cauchy_th}
\sup\limits_{z \in U} \left|f(z^1,\dots,z^d) - \sum_{k=1}^N \frac{\lambda_k}{(\xi_k^1 - z^1)\cdots(\xi_k^d - z^d)}\right| < \varepsilon.
\end{equation}
Furthermore, the approximation error satisfies \( \varepsilon = O(N^{-p}) \) for any fixed integer \( p \) and sufficient smoothness of functions $f$.

\end{theorem}

Cauchy activation functions are derived from the above approximation theorem. This demonstrates that XNet can approximate analytic functions with error decreasing at an arbitrary polynomial rate, whereas traditional neural networks typically require increased width and depth to achieve comparable accuracy. Structurally, XNet can be interpreted as a single-layer MLP with Cauchy activation functions, while KAN utilizes B-splines as basis functions.

Cauchy kernels and B-splines provide distinct strategies for function approximation. The following section explores their differences from three perspectives: 1. Approximation Power; 2. Computational Efficiency; 3. Numerical Properties.

\paragraph{Approximation Rate} 
Given an analytic function \( f \), the Cauchy kernel approximation satisfies the following error bound:
\begin{equation}
\| f - f_N \| = O(N^{-p}), \quad \forall p > 0.
\end{equation}
In contrast, B-spline-based approximation satisfies:
\begin{equation}
\| f - f_N \| = O(N^{-k}),
\end{equation}
where \( k \) is the degree of the B-spline basis. This highlights that Cauchy kernels can achieve arbitrarily fast convergence, whereas B-splines are inherently constrained by their polynomial degree. The complete proof is provided in Appendix \ref{ref:th_proof}.

\paragraph{Computational Efficiency}
The Cauchy Approximation Theorem indicates that to achieve a specified approximation error \( \epsilon \), the number of required basis functions for Cauchy kernels is given by:
\begin{equation}
N_{\text{Cauchy}} = O(\epsilon^{-1/p}),
\end{equation}
where \( p \) is any large positive integer. This flexibility allows for the convergence rate to be tailored very aggressively, making it possible to enhance the precision of approximation significantly by increasing \( p \).

Contrastingly, the use of B-splines involves a fixed degree \( k \), which generally does not exceed practical limits due to computational and numerical stability considerations:
\begin{equation}
N_{\text{B-spline}} = O(\epsilon^{-1/k}).
\end{equation}
While B-splines are powerful and flexible within their operational scope, they are inherently limited by the highest degree \( k \) that can be practically utilized, which may not offer the same level of aggressive error reduction as Cauchy kernels.

Furthermore, the structural simplicity of Cauchy-based systems, which can be efficiently implemented using a single-layer network architecture, provides significant advantages in terms of computational speed and resource usage. This simplicity allows for rapid adjustments and learning, particularly advantageous in scenarios requiring real-time processing and high responsiveness. In contrast, networks based on B-splines or other more complex architectures often require multiple layers or more intricate setups, which can increase computational time and delay convergence.

\textbf{Example in Practice:}
In real-world applications, especially those requiring quick response times and high precision, the single-layer Cauchy network can outperform more complex multi-layer networks. For instance, in tasks involving real-time image or signal processing, the ability to compute approximations quickly and accurately is crucial, and the Cauchy kernel's properties directly contribute to superior performance in these cases.

\paragraph{Numerical Characteristics}
For function approximation with Cauchy kernels and B-splines, the following aspects are crucial for assessing their practical application:

\begin{itemize}
    \item \textbf{Matrix Structure and Stability:}
    Cauchy kernel interpolation results in dense matrices, which might lead to large condition numbers as the problem size increases. This is analyzed using Gershgorin's Circle Theorem:
    \begin{equation}
    |\lambda - A_{ii}| \leq \sum_{j \neq i} |A_{ij}| = \sum_{j \neq i} \left|\frac{1}{\xi_i - z_j}\right|,
    \end{equation}
    indicating that the condition number \( \kappa(A) = O(N) \) as \( N \) increases. Although this suggests potential issues in numerical stability, the single-layer structure of networks using Cauchy kernels allows for rapid computation and effective management of these matrices, particularly when combined with modern computational techniques to mitigate numerical instability.

    B-splines, benefiting from local support, generate sparse banded matrices that inherently have better stability and lower condition numbers, typically \( O(1) \).

   \item \textbf{Derivative Properties:}
    Cauchy kernels allow precise and straightforward derivative computations due to their explicit formulas:
    \begin{equation}
    \frac{d}{d(d^2)} \left( \frac{1}{x^2 + d^2} \right) = -\frac{1}{(x^2 + d^2)^2}.
    \end{equation}
    This property is particularly useful in applications requiring parameter sensitivity analysis, as it enables efficient computation of derivatives with respect to \( d^2 \), which can play a role in optimization and function fitting.

    In contrast, B-splines, despite their robustness, require piecewise polynomial differentiation, which increases computational complexity and may reduce precision when handling high-order derivatives.

\end{itemize}

The single-layer architecture of Cauchy-based networks not only simplifies the computational model but also enhances the potential for achieving high precision in function approximation. This structure is particularly effective in scenarios requiring real-time processing and high responsiveness, where the ability to quickly adjust and compute with high accuracy is essential. Detailed proofs and further theoretical elaborations on these properties are presented in Appendix \ref{ref:th_proof}.

\section{Experimental Setup and Results}

To comprehensively evaluate XNet's performance compared to KAN and traditional MLPs, we conduct experiments across three domains: function approximation, PDE solving, and reinforcement learning. Each domain is designed to test different aspects of the networks' capabilities.

\subsection{Function Approximation Experiments}

\subsubsection{Study Design}
We evaluate the networks on three categories of functions with increasing complexity:
\begin{itemize}
    \item \textbf{Low-Dimensional Tasks:}
    \begin{itemize}
        \item Discontinuous: Heaviside step function
    \end{itemize}
    
    \item \textbf{High-Dimensional Tasks:}
    \begin{itemize}
        \item 4D nonlinear: 
        $\exp\left(\frac{1}{2} \left(\sin\left(\pi(x_{1}^{2}+x_{2}^{2})\right) + x_{3}x_4\right)\right)$
        \item 100D periodic:
        $\exp\left(\frac{1}{100}\sum_{i=1}^{100}\sin^2\left(\frac{\pi x_i}{2}\right)\right)$
    \end{itemize}
    
    \item \textbf{Noisy Data Tasks:}
    \begin{itemize}
        \item Time series system with varying noise levels (0\%, 5\%, 10\%)
    \end{itemize}
\end{itemize}

The experimental comparison between XNet, B-spline, and KAN demonstrates XNet's superior approximation ability. Except for the first function  example and the final task, all other examples are from the referenced article, with KAN settings matching those from the original experiments. This ensures a fair comparison, fully proving that XNet has stronger approximation capabilities in various benchmarks.

\begin{table}[htbp]
\caption{Low-dimensional and High-dimensional Functions Examples}\label{tab:function_examples}
\centering
\resizebox{0.75\textwidth}{!}{ 
\begin{tabular}{c}  
\toprule
\textbf{Heaviside step function} \\
\midrule
\begin{minipage}[b]{0.9\textwidth}  
\centering
$f(x) = \begin{cases} 
1, & x > 0 \\
0, & \text{otherwise}
\end{cases}$
\\[10pt] 
\includegraphics[width=0.5\textwidth]{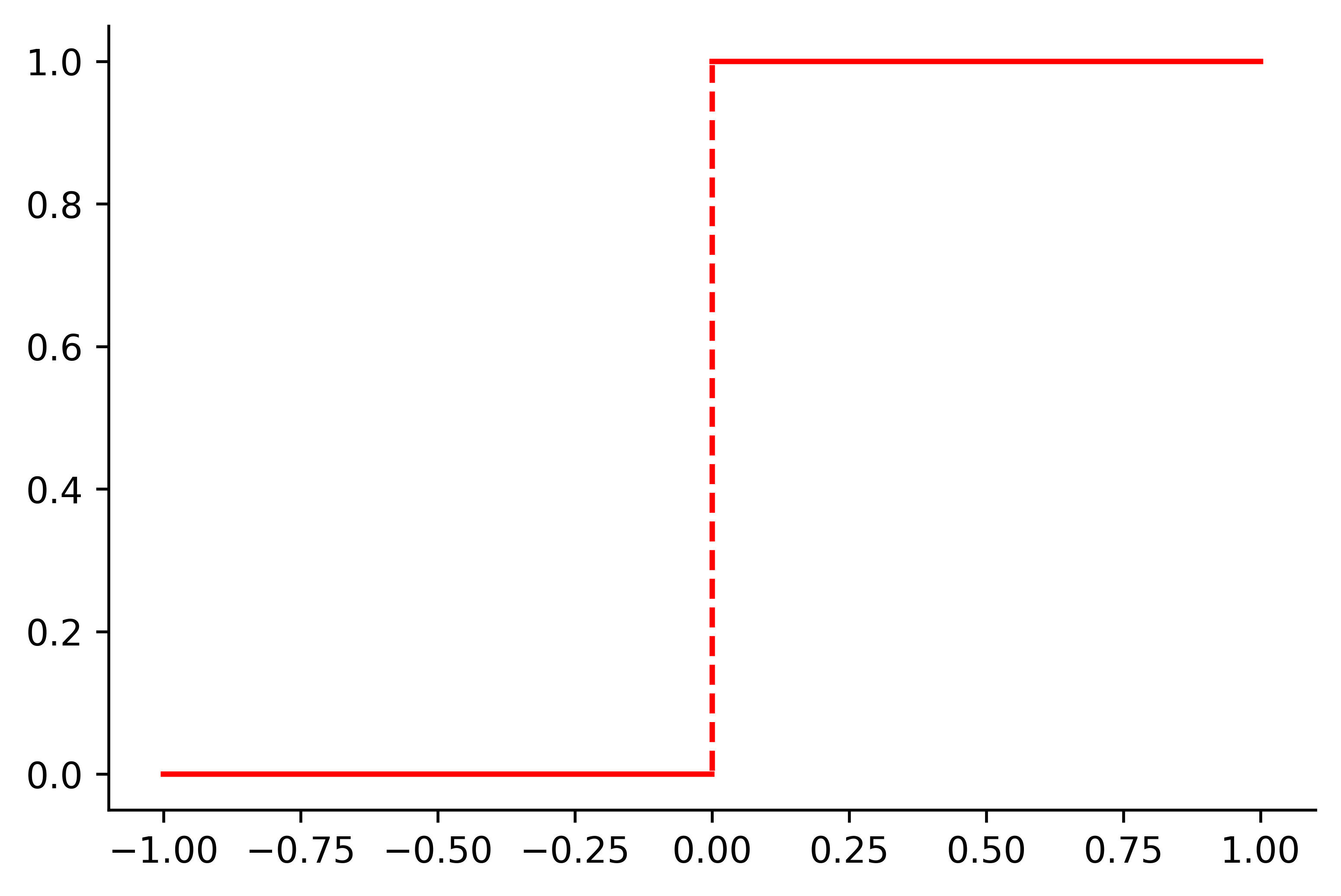}  
\end{minipage} \\
\midrule
\textbf{High-dimensional Functions} \\
\midrule
\begin{minipage}[c]{0.9\textwidth}
\centering
\begin{equation*}
f(x_1,x_2,x_3,x_4) = \exp\left(\frac{1}{2}\left(\sin\left(\pi(x_{1}^{2}+x_{2}^{2})\right) + x_{3}x_{4}\right)\right)
\end{equation*}
\begin{equation*}
f(x_1, \dots, x_{100}) = \exp\left(\frac{1}{100}\sum_{i=1}^{100}\sin^2\left(\frac{\pi x_i}{2}\right)\right)
\end{equation*}
\end{minipage} \\
\bottomrule
\end{tabular}
} 
\end{table}

\subsubsection{Experimental Protocol}
\begin{itemize}
    \item \textbf{Network Configurations} \, \\
        - XNet: 64 or 5000 basis functions \\
        - KAN: Hierarchical structure with optimal grid points\\
        - MLP: Comparable parameter count
        
    \item \textbf{Training Setup}\, \\
        - Low-D: 1000 training / 1000 test points\\
        - High-D: 8000 training / 1000 test points\\
        - Optimizer: Adam (XNet), L-BFGS (KAN)
        
    \item \textbf{Evaluation Metrics}\, \\
        - Accuracy: MSE, RMSE, MAE\\
        - Efficiency: Training time, Memory usage\\
        - Stability: Loss curves, Convergence rate
\end{itemize}

[Table \ref{tab:function_examples} shows representative test functions and their visualizations]

\subsubsection{Heaviside step function apprxiamtion}\label{sec:1d}

\begin{table}[h!]
\centering
\small
\caption{Performance comparison between XNet and KAN.}
\setlength{\tabcolsep}{3pt}
\begin{tabular}{cccc} 
\toprule
\textbf{Metric} & \textbf{MSE} & \textbf{RMSE} & \textbf{MAE} \\
\midrule
\textbf{XNet [1, 64, 1]} & 8.99e-08 & 3.00e-04 & 1.91e-04 \\
\textbf{KAN [1, 1] with 200 grids} & 5.98e-04 & 2.45e-02 & 3.03e-03 \\
\bottomrule
\end{tabular}
\label{tab:xnet_kan_performance_1d}
\end{table}

\begin{figure}[h!]
    \centering
    \begin{minipage}[b]{0.48\linewidth}
        \centering
        \includegraphics[width=\textwidth]{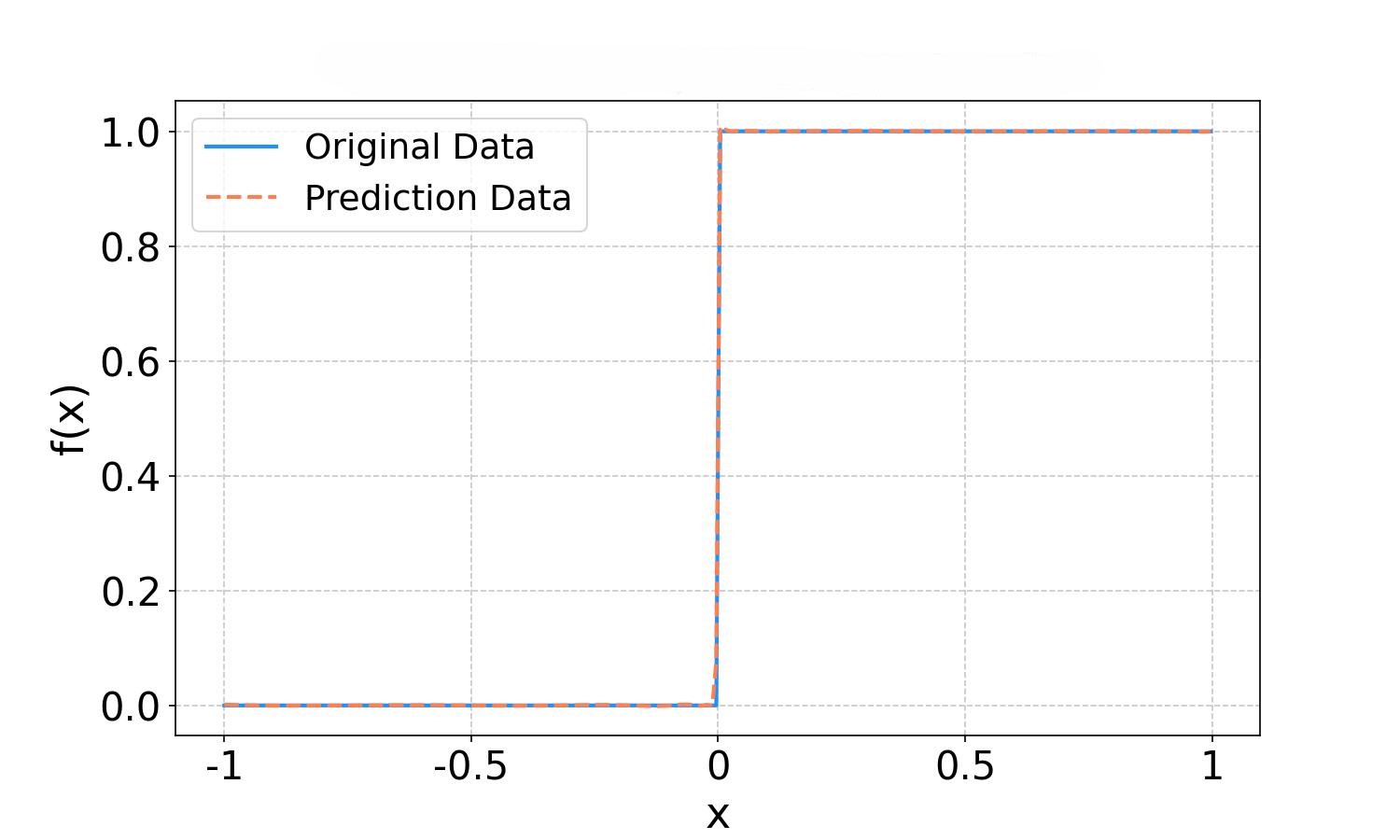}
        \put(-105,-10){\footnotesize (a)}
    \end{minipage}
    \hfill
    \begin{minipage}[b]{0.48\linewidth}
        \centering
        \includegraphics[width=\textwidth]{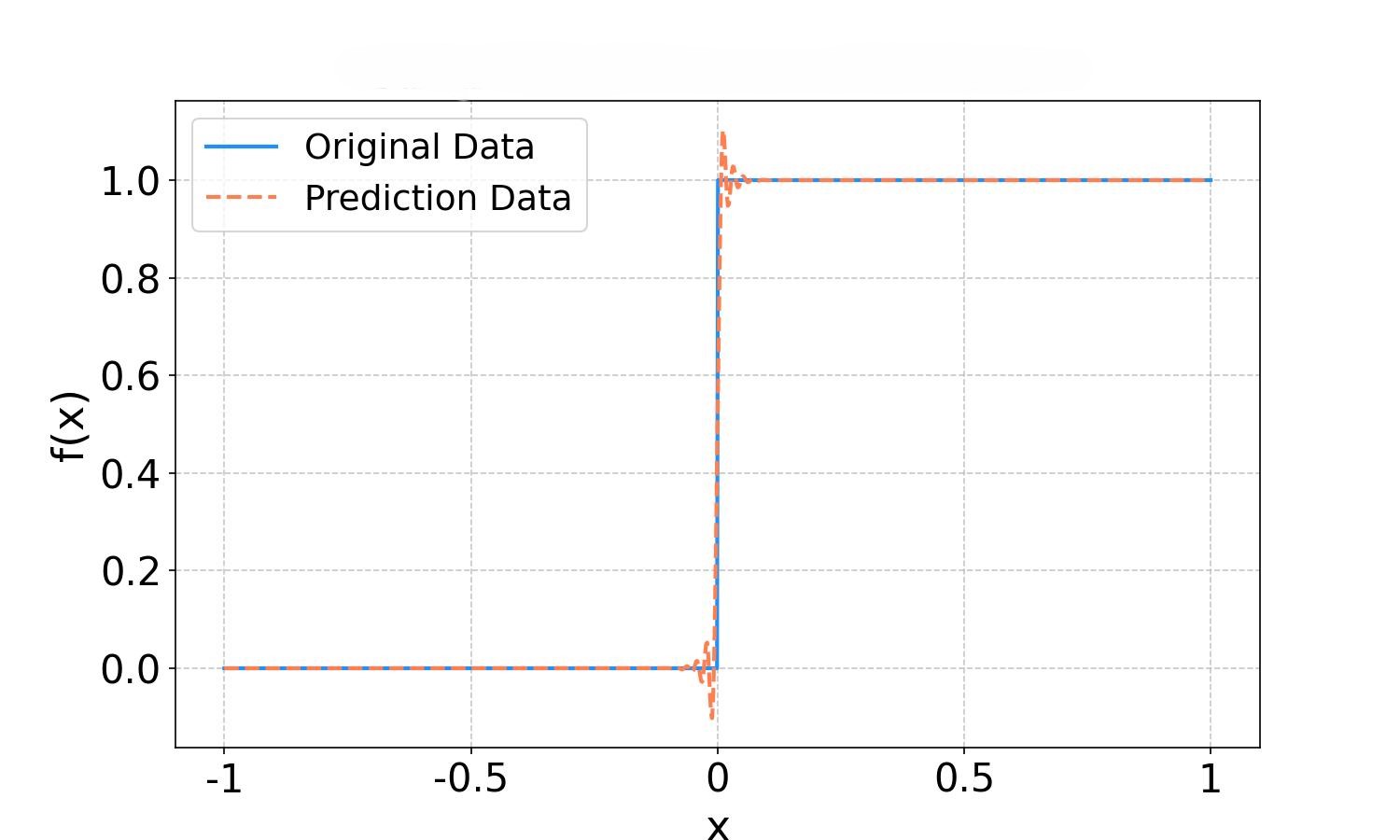}
        \put(-105,-10){\footnotesize (b)}
    \end{minipage}
    \vspace{0.5cm} 
    \begin{minipage}[b]{0.48\linewidth}
        \centering
        \includegraphics[width=\textwidth]{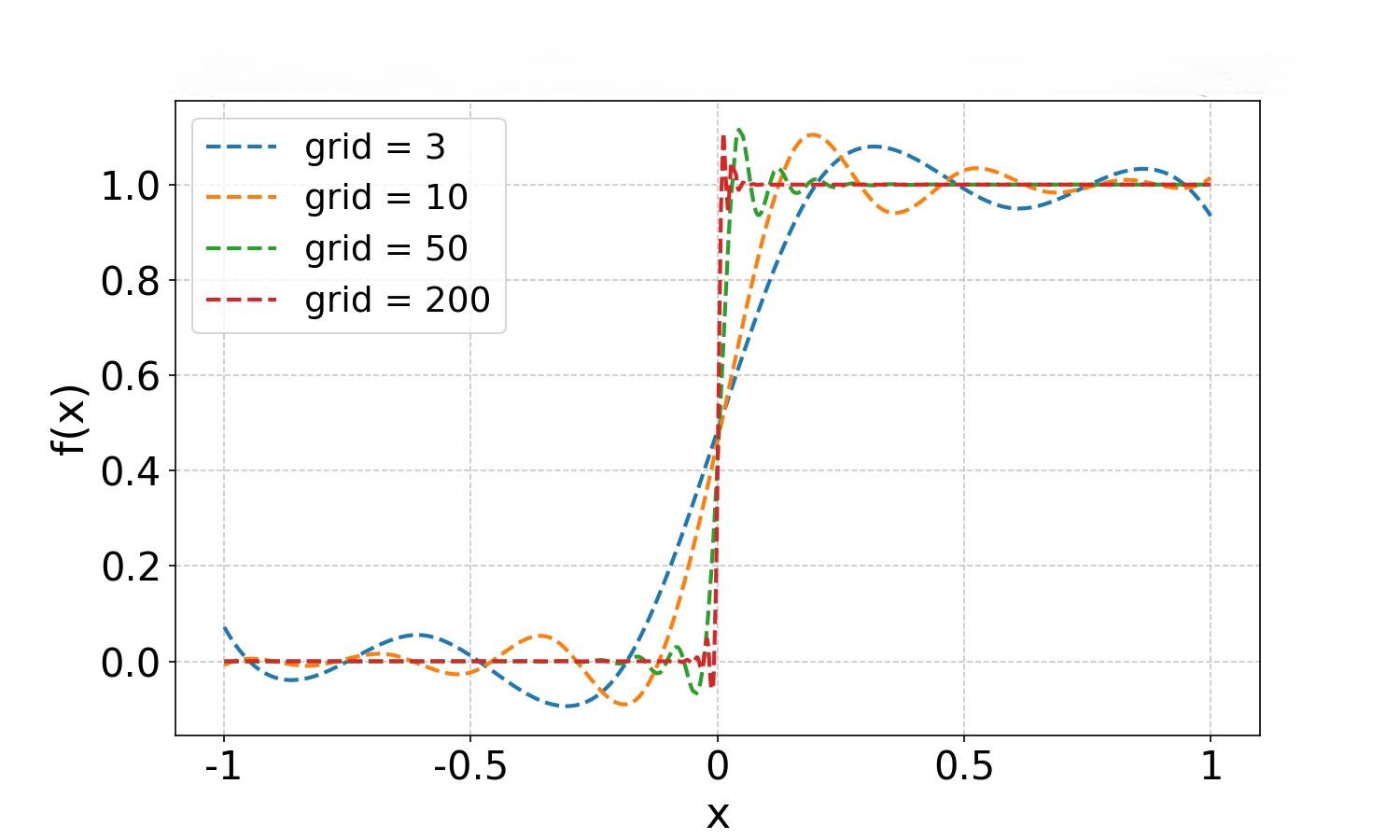}
        \put(-105,-10){\footnotesize (c)}
    \end{minipage}
    \hfill
    \begin{minipage}[b]{0.48\linewidth}
        \centering
        \includegraphics[width=\textwidth]{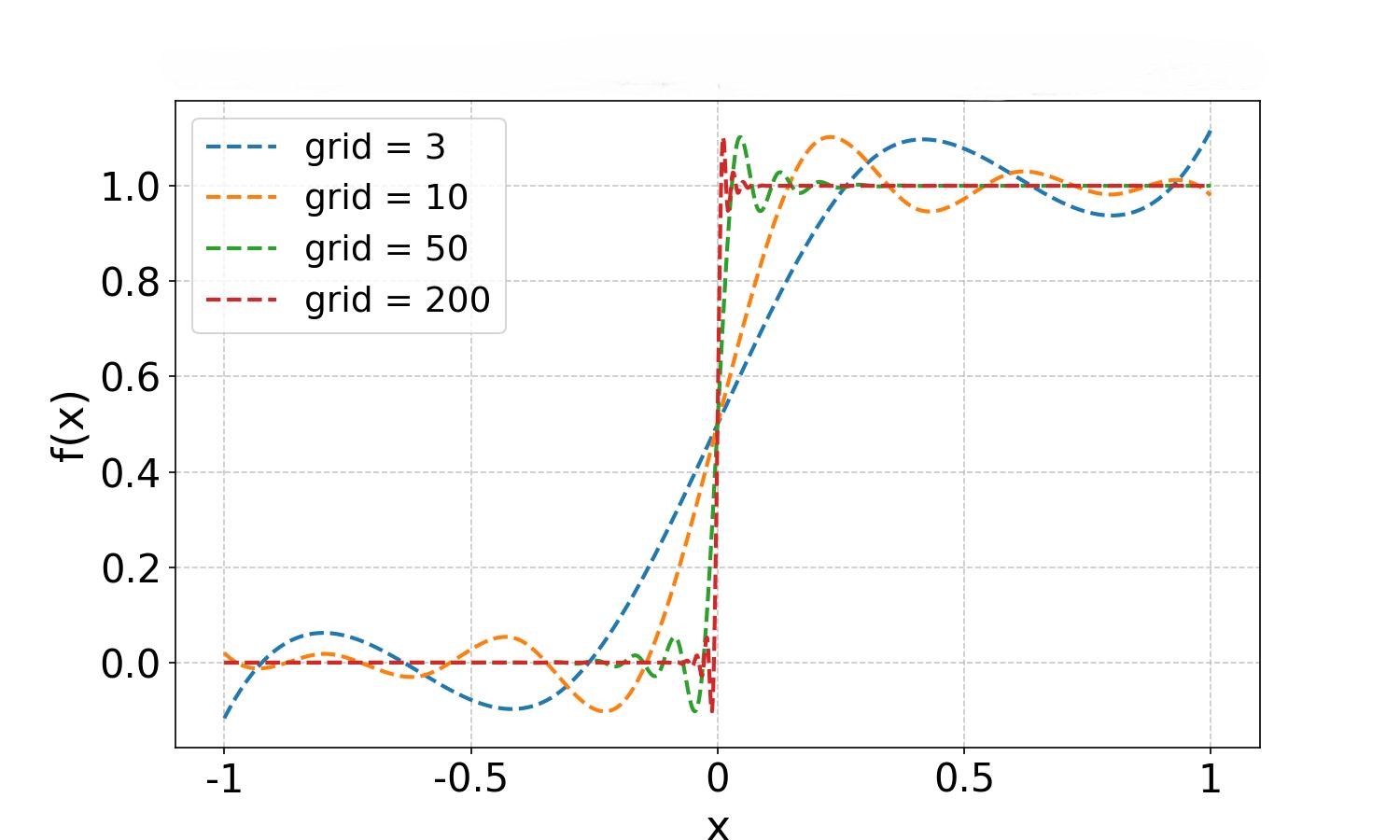}
        \put(-105,-10){\footnotesize (d)}
    \end{minipage}
    \vspace{-4.2pt}
    \caption{Heaviside step function approximation comparison: (a) XNet, with 64 basis functions; (b) KAN [1,1], with $k=3$, grid = 200; (c) B-Spline, with $k=3$; (d)  KAN [1,1], with $k=3$.}
    \label{fig:Heaviside_comparison}
\end{figure}

As shown in Figure \ref{fig:Heaviside_comparison}, both B-Spline and KAN exhibit "overshoot," leading to local oscillations at discontinuities. We speculate that this is due to the fact that a portion of KAN's output is represented by B-Splines. 
While adjusting the grid can alleviate this phenomenon, it introduces complexity in tuning parameters (see Table \ref{table:kan_1d} in appendix A.1). In contrast, XNet demonstrates superior performance, providing smooth transitions at discontinuities. Notably, in terms of fitting accuracy in these regions, XNet's MSE is 1000-fold times smaller than that of KAN.

\subsubsection{Approximation with high-dimensional functions}\label{sec:nd}
We continue to compare the approximation capabilities of KAN and XNet for high-dimensional functions. Following the procedure described in the article \cite{liu2024kan}, 8000 points were used for training and another 1000 points for testing. XNet is trained with adam, while KAN is initialized to have G = 3, trained with LBFGS, with increasing number of grid points every 200 steps to cover G = {3, 5, 10, 20, 50}.

First, we consider the four-dimensional function
\( \exp\left(\frac{1}{2}\left(\sin\left(\pi(x_{1}^{2}+x_{2}^{2})\right) + x_{3}x_{4}\right)\right) \). For this case, the KAN structure is configured as [4,4,2,1], while XNet is equipped with 5000 basis functions. Under the same number of iterations, XNet achieves higher accuracy in less time (see Table \ref{table:4d_1}), the MSE is 1000-fold smaller than that of KAN.

\begin{table}[!ht]
\centering
\small 
\caption{Comparison of XNet and KAN on \(  \exp\left(\frac{1}{2}\left(\sin\left(\pi(x_{1}^{2}+x_{2}^{2})\right) + x_{3}x_{4}\right)\right) \).}
\label{table:4d_1}
\resizebox{0.6\textwidth}{!}{
\begin{tabular}{ccccc} 
\toprule
\textbf{Metric}  & \textbf{MSE} & \textbf{RMSE} & \textbf{MAE} & \textbf{Time (s)} \\
\midrule
\textbf{XNet [2, 5000, 1]} & 2.3079e-06 & 1.5192e-03 & 8.3852e-04 & 78.18 \\ 
\textbf{KAN [4,2,2,1]} & 2.6151e-03 & 5.1138e-02 & 3.6300e-02 & 143.1 \\ 
\bottomrule
\end{tabular}}
\end{table}

Next, we consider the 100-dimensional function \( \exp(\frac{1}{100}\sum_{i=1}^{100}\sin^2(\frac{\pi x_i}{2})) \). For this case, the KAN structure is configured as [100, 1, 1], while XNet has 5000 basis functions. Under the same number of iterations, XNet achieved higher accuracy in less time compared to KAN (see Table \ref{table:100d_1}).

\begin{table}[!ht]
\centering
\small 
\caption{Comparison of XNet and KAN on \resizebox{0.22\textwidth}{!}{\(  \exp\left(\frac{1}{100}\sum_{i=1}^{100}\sin^2\left(\frac{\pi x_i}{2}\right)\right) \)}.}
\label{table:100d_1}
\resizebox{0.6\textwidth}{!}{
\begin{tabular}{ccccc} 
\toprule
\textbf{Metric}  & \textbf{MSE} & \textbf{RMSE} & \textbf{MAE} & \textbf{Time (s)} \\
\midrule
\textbf{XNet [2, 5000, 1]} & 6.8492e-04 & 2.6171e-02 & 2.0889e-02 & 158.69 \\ 
\textbf{KAN [100, 1, 1]} & 6.5868e-03 & 8.1159e-02 & 6.4611e-02 & 556.5 \\ 
\bottomrule
\end{tabular}}
\end{table}

As dimensionality increases, the computational efficiency of KAN decreases significantly, while XNet shows an advantage in this regard. The approximation accuracy of both networks declines with increasing dimensions, which we hypothesize is related to the sampling method and the number of samples used.

\begin{figure}[h!]
    \centering
    \begin{minipage}[b]{0.45\linewidth}
        \centering
        \includegraphics[width=\textwidth]{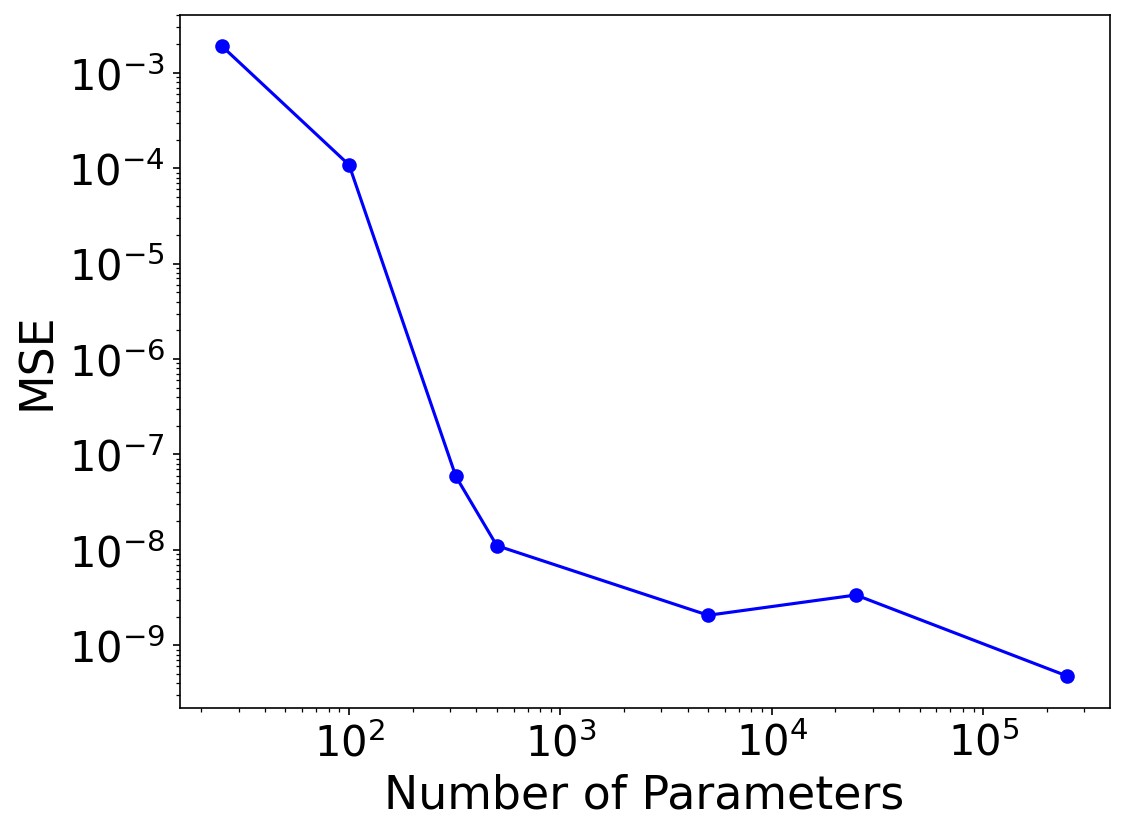}
        \put(-55,-8){\footnotesize (a)}
    \end{minipage}
    \hfill
    \begin{minipage}[b]{0.45\linewidth}
        \centering
        \includegraphics[width=\textwidth]{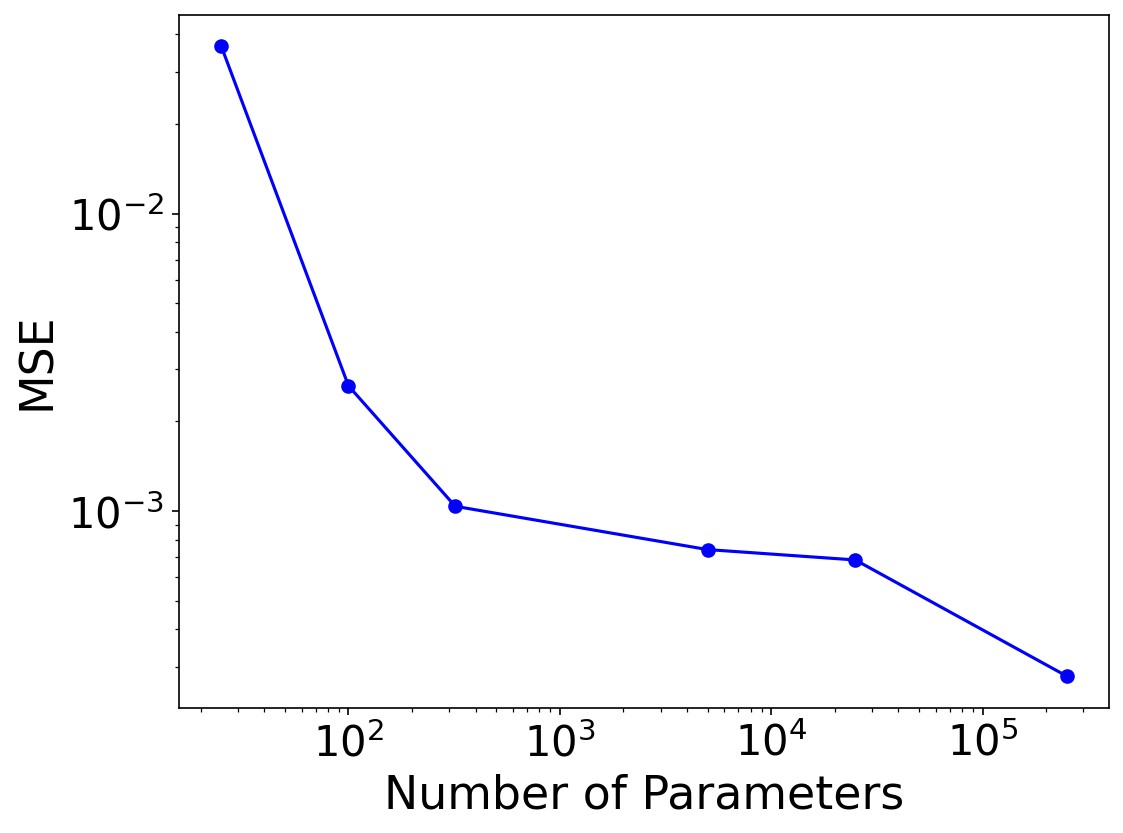}
        \put(-55,-8){\footnotesize (b)}
    \end{minipage}
    \caption{Performance of XNet on approximating different functions with varying numbers of parameters: 
    (a) \(\exp\left(\frac{1}{2}\left(\sin\left(\pi(x_{1}^{2}+x_{2}^{2})\right) + x_{3}x_{4}\right)\right)\); and 
    (b) \(\exp\left(\frac{1}{100}\sum_{i=1}^{100}\sin^2\left(\frac{\pi x_i}{2}\right)\right)\).}
    \label{XNet_params1}
\end{figure}

As shown in Figure \ref{XNet_params1}, XNet achieves high accuracy with relatively few network parameters. Moreover, as the number of parameters increases, XNet can further enhance its accuracy. Given its performance in function approximation tasks, both in terms of computational efficiency and accuracy, we conclude that XNet is a highly efficient neural network with strong approximation capabilities.
Building on this, in the following subsection, we apply PINN, KAN, and XNet to approximate the value function of the Poisson equation.

\subsubsection{Functions with noise} \label{sec:functions_noisy}

The system is generated by the following equations:
$$x_5^i=0.1x_0^ix_1^i+0.5sin(x_2^ix_3^i)+ sin(x_4^i) + \mu^i, $$
where $i=1,2,...,n,$ and
$$x_0^i=x_1^{i-1},x_1^i=x_2^{i-1},x_2^i=x_3^{i-1},x_4^i=x_5^{i-1},  $$
where the initial conditions \( x_0^0, x_1^0, x_2^0, x_3^0, x_4^0 \) are randomly sampled in the range \([0, 0.2]\), and the noise term \( \mu^i \) is sampled from a normal distribution, \( \mu^i \sim N(0, \text{noise}) \). This generates a series \( \{ f^i = x_5^i \}_{i=1,\dots,n} \), with \( n = 300 \). The first 80\% of the data contains noise, while the last 20\% is noise-free.
Clearly, the system is governed by relatively simple functions. The task of predicting the sixth data point using the first five data points becomes a high-dimensional function approximation problem.

Figures [\ref{fig:ts1_noise}] and [\ref{fig:comparison_KAN_XNet_noise}] in \ref{ref:function noise} show a comparison of the predictive performance of [5, 64, 1] KAN and [5, 20, 1] XNet on three scenarios: one with no noise (noise = 0), one with moderate noise (noise = 0.05), and one with high noise (noise = 0.1).

The results indicate that XNet significantly outperforms KAN in both settings, particularly under noisy conditions. When there is no noise, XNet achieves an MSE of \( 4.7099 \times 10^{-7} \), which is lower than that of KAN (\( 1.3209 \times 10^{-5} \)). Similarly, XNet's RMSE, and the MAE are drastically lower than KAN's.
In the presence of moderate noise (noise = 0.05), and high noise (noise = 0.1), XNet show a more significant advantage in metrics, and
it is evident from Figure (\ref{fig:ts1_noise}) that XNet demonstrates superior noise resistance compared to KAN.

\begin{table}[!ht]
\centering
\small 
\caption{Performance comparison of KAN [5, 64, 1] and XNet [5, 20, 1] under different noise levels.}
\label{table:combined}
\resizebox{0.75\textwidth}{!}{
\begin{tabular}{cccccc} 
\toprule
\textbf{Noise Level} & \textbf{Model} & \textbf{MSE} & \textbf{RMSE} & \textbf{MAE} & \textbf{Time (s)} \\
\midrule
\multirow{2}{*}{0}       & \textbf{KAN [5, 64, 1]}   & 1.3209e-05 & 3.6344e-03 & 4.8311e-03 & 17.8863 \\ 
                         & \textbf{XNet [5, 20, 1]}  & 4.7099e-07 & 6.8629e-04 & 5.2943e-04 & 9.6502  \\ 
\midrule
\multirow{2}{*}{0.05}    & \textbf{KAN [5, 64, 1]}   & 1.6784e-03 & 4.0968e-02 & 3.1549e-02 & 17.9535 \\ 
                         & \textbf{XNet [5, 20, 1]}  & 4.5109e-04 & 2.1239e-02 & 1.5797e-02 & 9.4391  \\ 
\midrule
\multirow{2}{*}{0.1}     & \textbf{KAN [5, 64, 1]}   & 9.4653e-03 & 9.7290e-02 & 7.8063e-02 & 17.8863 \\ 
                         & \textbf{XNet [5, 20, 1]}  & 6.5517e-04 & 2.5596e-02 & 2.0233e-02 & 9.0296  \\ 
\bottomrule
\end{tabular}}
\end{table}

In this example of a simple function-driven time series, XNet clearly outperforms KAN, particularly in noisy and noise-free environments. Given these results, we hypothesize that XNet will also exhibit superior performance in highly noisy, real-world datasets. The [5,64,1] KAN model, however, shows signs of overfitting, with excellent performance on the training set but noticeable degradation on the test set.

\subsection{PDE Solving Capability}
\begin{figure}[h!]
    \centering
    \begin{minipage}[b]{0.42\textwidth}
        \centering
        \includegraphics[width=\textwidth]{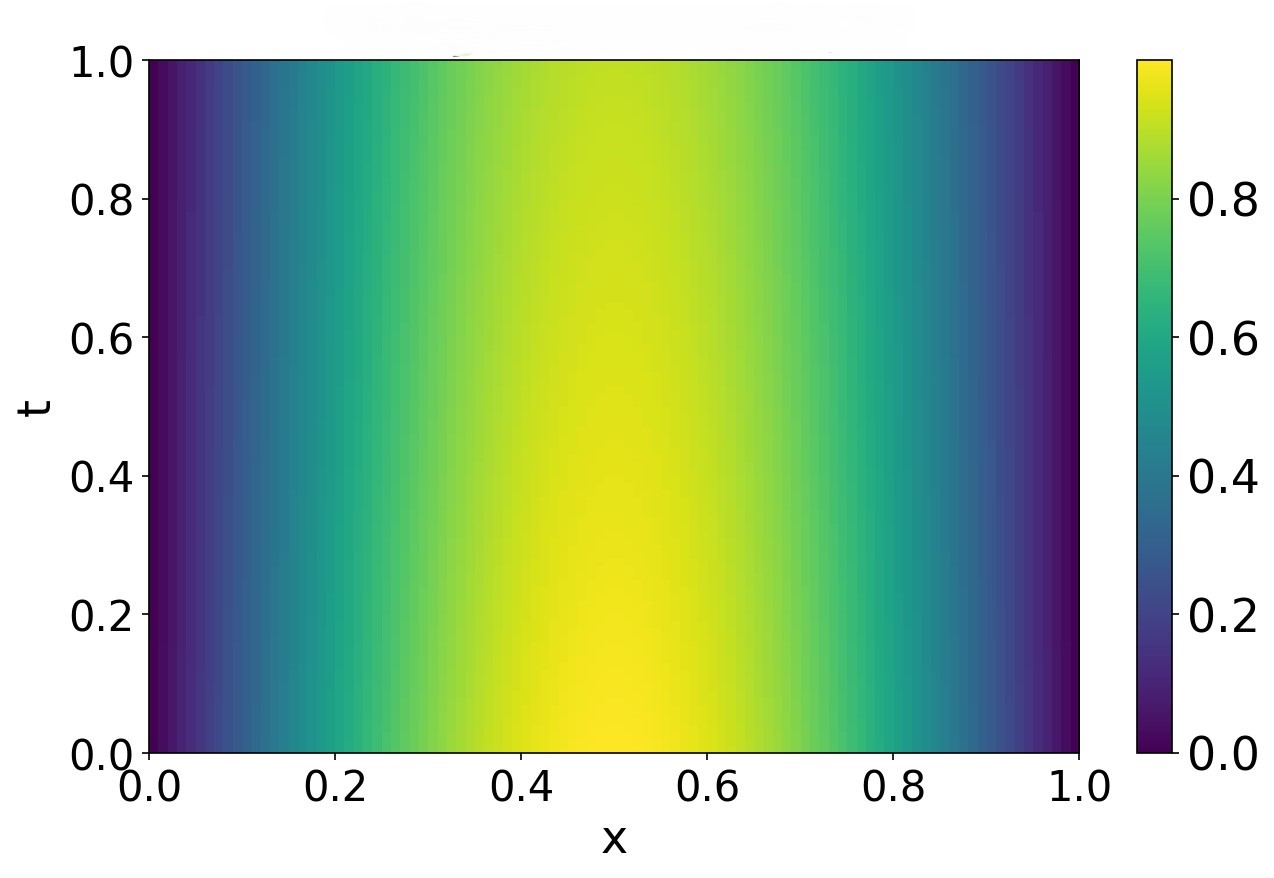}
        \vspace{-20pt}
	\caption{solution of the Heat equation}
	\label{fig:NS_solution}
    \end{minipage}
\end{figure}

We focus on the Heat equation as a benchmark PDE problem:
\begin{equation}\label{NS_equation}
    \begin{cases} 
        \begin{aligned}
            &\frac{{\partial u}}{{\partial t}} = \nu \frac{{{\partial ^2}u}}{{\partial {x^2}}}, \quad (x,t) \in [0,1] \times [0,1],\\
            &u(x,0) = \sin (\pi x), \quad x \in [0,1],\\
            &u(0,t) = u(1,t) = 0,
        \end{aligned}
    \end{cases}
\end{equation}
with known analytical solution $u(x,t) = u(x,t) = {e^{ - \nu {\pi ^2}t}}\sin (\pi x){\rm{ }}$ (Figure \ref{fig:NS_solution}).

\subsubsection{Implementation Details}
\begin{itemize}
    \item \textbf{Network Settings}\, \\
        - MLP: [2, 20, 20, 1] architecture\\
        - XNet: 20 and 200 basis functions\\
        - KAN: [2, 10, 1] structure
        
    \item \textbf{Training Protocol}\, \\
        - Viscosity coefficient $\nu=001$ \\
        - Interior points: 2500 \\
        - Boundary points: 150 \\
        - Loss weight: $\alpha = 0.1$
\end{itemize}

\subsubsection{Heat function} \label{sec:application}
We use the framework of physics-informed neural networks (PINNs) to solve this PDE \eqref{NS_equation}, with the loss function given by
\begin{equation}\label{pinn_loss}
    \begin{aligned}
        \mathrm{loss} = \alpha \, \mathrm{loss}_i &+ \mathrm{loss}_o = \frac{1}{n_o} \sum_{i=1}^{n_o} \left|u^\theta(z_i) - u(z_i)\right|^2
        \\
        &+ \alpha \frac{1}{n_i} \sum_{i=1}^{n_i} \left|u_t^\theta(z_i) - \nu u_{xx}^\theta(z_i)\right|^2 ,
    \end{aligned}
\end{equation}
where \(\mathrm{loss}_i\), referred to as the interior loss, is evaluated by discretizing the domain into \(n_i\) uniformly sampled points \(z_i = (x_i, t_i)\). Similarly, \(\mathrm{loss}_o\), representing the initial and boundary loss, is computed using \(n_o\) uniformly sampled points on both the initial and boundary conditions. The hyperparameter \(\alpha\) balances the contributions of the interior and boundary losses.

\begin{figure}[h!]
    \centering
    \begin{minipage}[b]{0.45\textwidth}
        \centering
        \includegraphics[width=\textwidth]{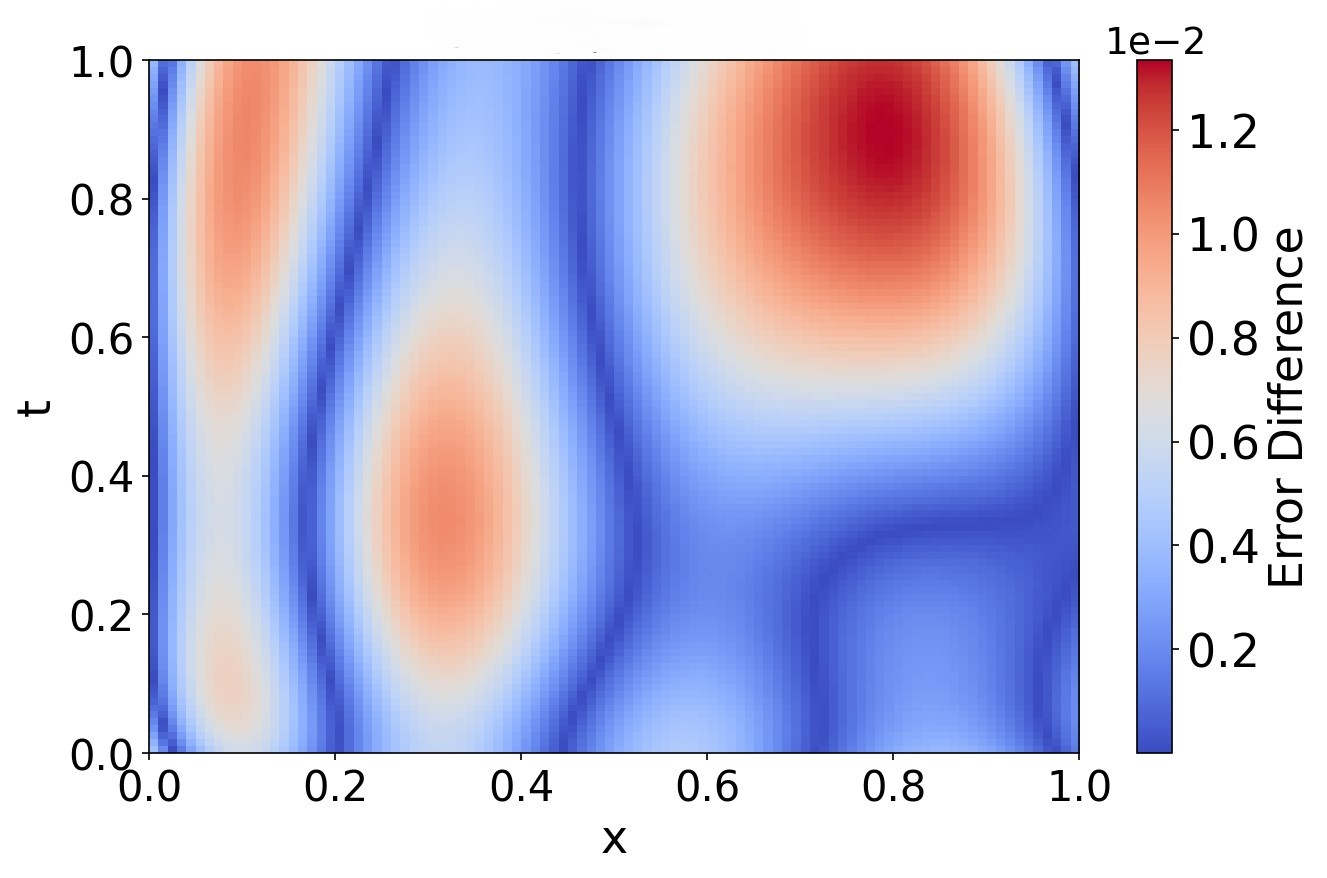}
        \put(-135,-8){\footnotesize (a) MLP [2, 20, 20, 1]}
    \end{minipage}
    \hfill
    \begin{minipage}[b]{0.45\textwidth}
        \centering
        \includegraphics[width=\textwidth]{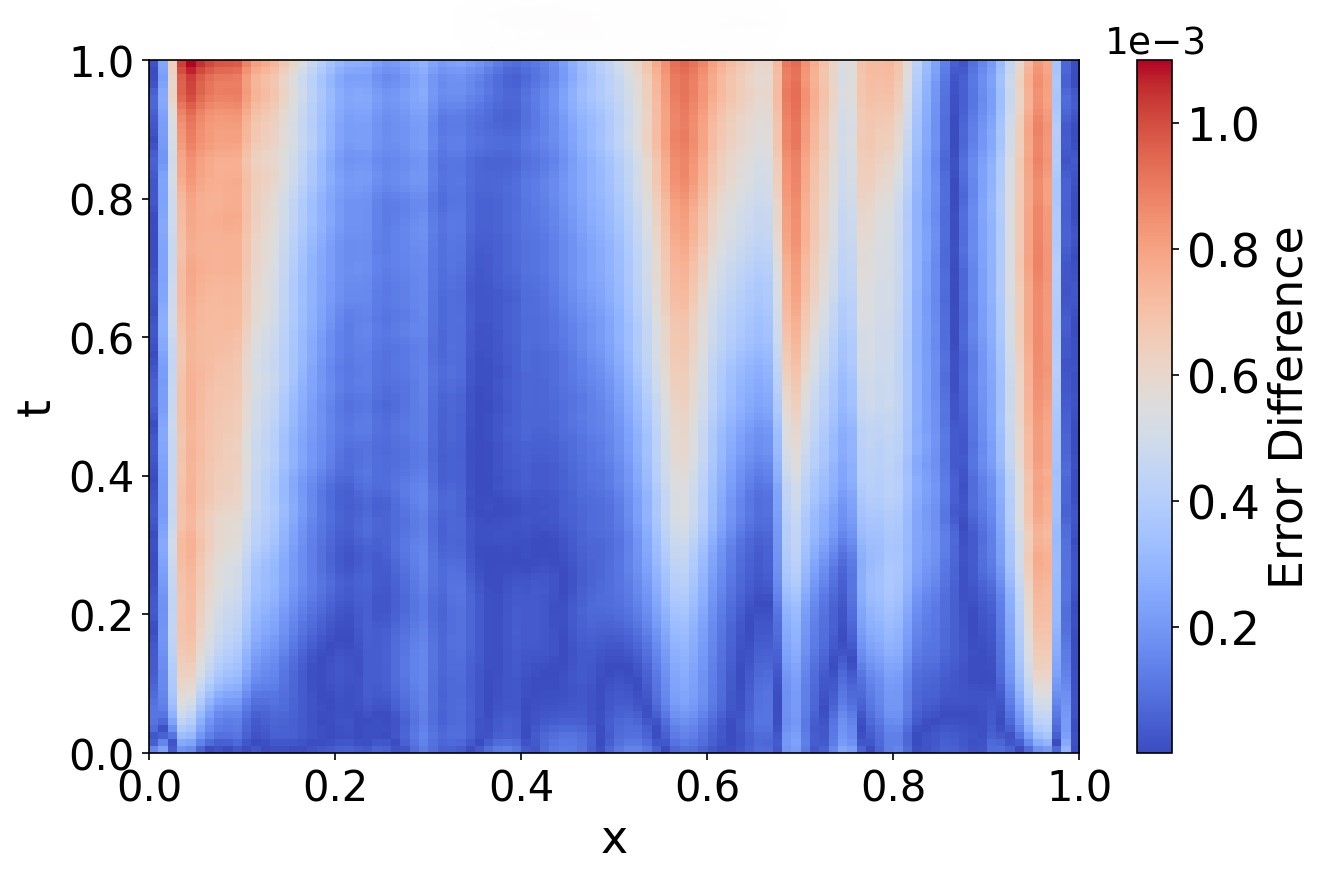}
        \put(-135,-8){\footnotesize (b) KAN [2, 10, 1]}
    \end{minipage}
    \caption{MLP and KAN Performance}
\end{figure}
\vspace{1em}
\begin{figure}[h!]
    \centering
    \begin{minipage}[b]{0.45\textwidth}
        \centering
        \includegraphics[width=\textwidth]{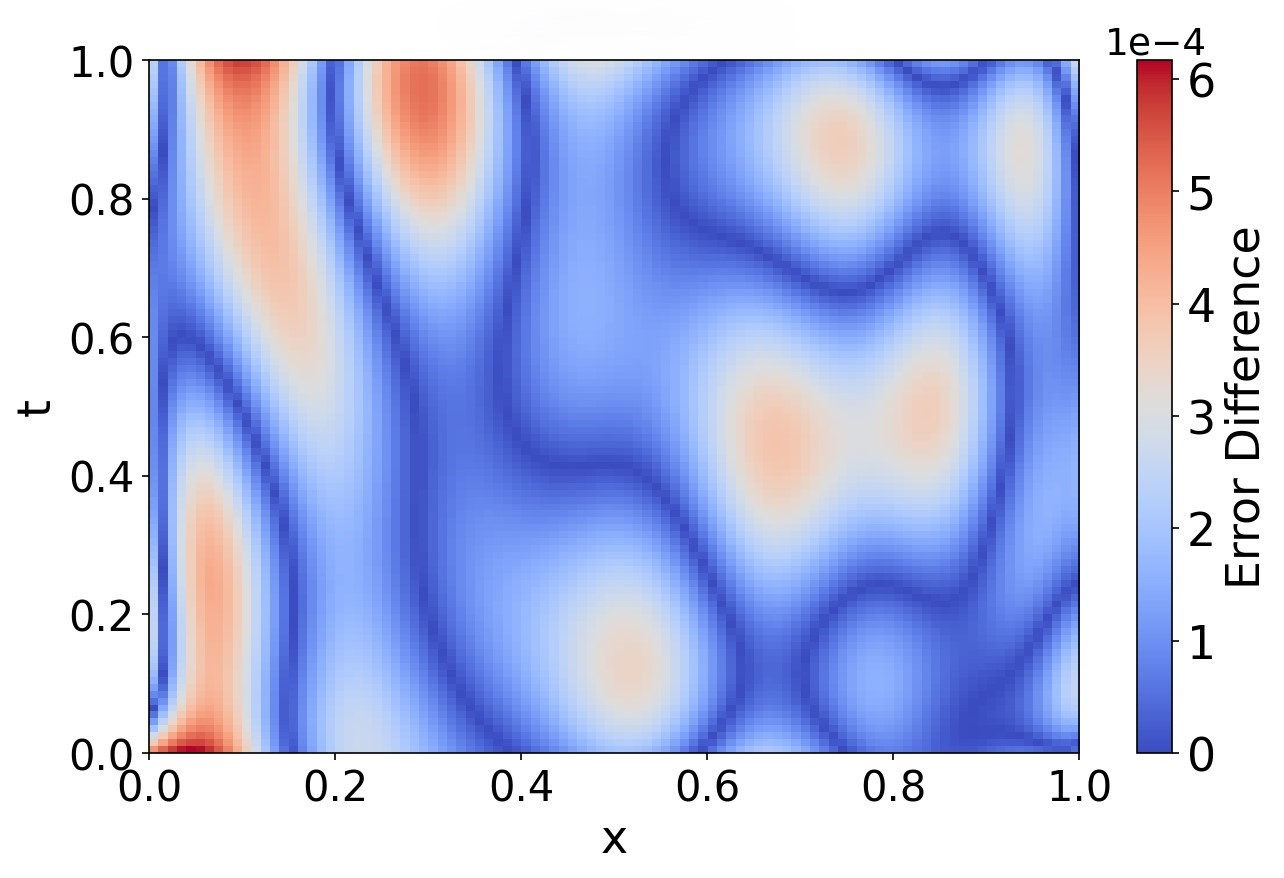}
        \put(-135,-8){\footnotesize (a) XNet [2, 20, 1]}
    \end{minipage}
    \hfill
    \begin{minipage}[b]{0.45\textwidth}
        \centering
        \includegraphics[width=\textwidth]{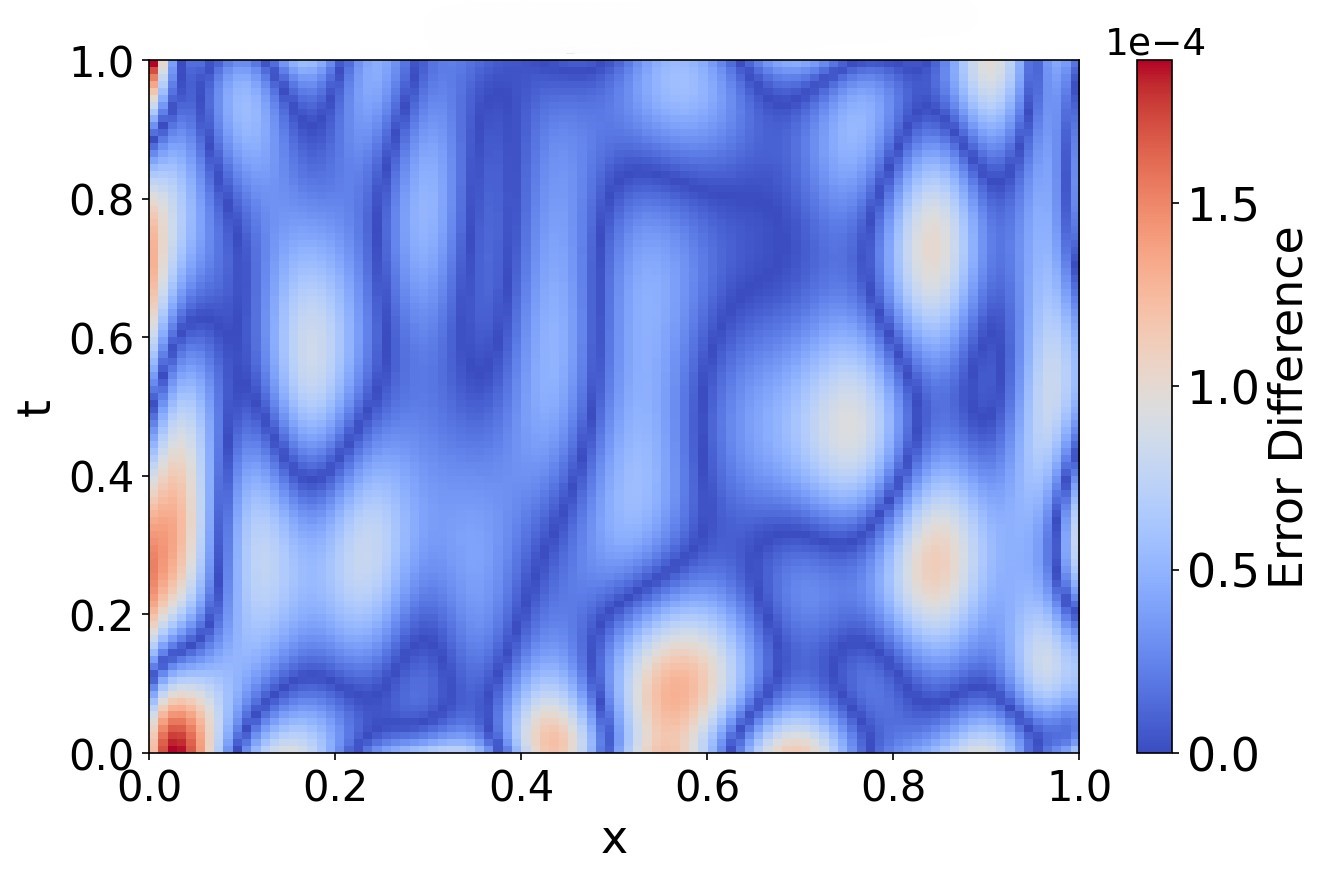}
        \put(-135,-8){\footnotesize (a) XNet [2, 200, 1]}
    \end{minipage}
    \caption{XNet Performance}
\end{figure}

We compare the KAN, XNet and MLP using the same hyperparameters $n_i=2500$, $n_o=150$, and $\alpha=0.1.$ We measured the error in the $L^2$ norm (MSE) and observed that XNet achieved a smaller error, requiring less computational time. 
A width-200 XNet is 100 times more accurate and 2 times faster than a 1-Layer width-10 KAN;
a width-20 XNet is 3 times more accurate and 5 times faster than a 1-Layer width-10 KAN (see Table \ref{table:NS_compare}).
Therefore we speculate that the XNet might have the potential of serving as a good neural network representation for model reduction of PDEs.  In general, KANs and MLPs are good at representing different function classes of PDE solutions, which needs detailed future study to understand their respective boundaries.

\begin{table}[!ht]
\centering
\small 
\caption{Comparison of XNet and KAN on the Heat equation.}
\label{table:NS_compare}
\resizebox{0.65\textwidth}{!}{
\begin{tabular}{ccccc} 
\toprule
\textbf{Metric} & \textbf{MSE} & \textbf{RMSE} & \textbf{MAE} & \textbf{Time (s)} \\
\midrule
\textbf{MLP [2,20,20,1]} &2.4536e-05	&4.9534e-03	&3.8323e-03	&43.8\\ 
\textbf{XNet [2,20,1]} & 3.8936e-08	&1.9732e-04	&1.5602e-04	& 43.5 \\ 
\textbf{KAN [2,10,1]} & 1.5106e-07	&3.8866e-04	&2.9661e-04	&254.6 \\ 
\textbf{XNet [2,200,1]} & 3.6867e-09&6.0718e-05	&5.0027e-05	&108.3 \\ 
\bottomrule
\end{tabular}}
\end{table}

MLP, KAN, and XNet are employed to construct physics-informed machine learning models for solving the 2D Poisson equation. The experimental results are presented in Appendix \ref{ref:exp_poisson}.

\subsection{Reinforcement Learning Applications}\label{online-RF}

We evaluate XNet, KAN, and MLP as function approximators within the Proximal Policy Optimization (PPO) framework on two continuous control tasks from the DeepMind Control Suite: \textit{HalfCheetah-v4} and \textit{Swimmer-v4}. Table \ref{tab:network_configs} summarizes the network configurations for each model.

\begin{table}[h]
    \centering
    \caption{Network configurations for reinforcement learning tasks.}
    \label{tab:network_configs}
    \resizebox{0.45\textwidth}{!}{
    \begin{tabular}{lc}
        \toprule
        \textbf{Model} & \textbf{Configuration} \\
        \midrule
        \textbf{MLP}  & Two hidden layers with 64 units each \\
        \textbf{KAN}  & Order $k = 2$, grid size $g = 3$ \\
        \textbf{XNet} & 64 basis functions \\
        \bottomrule
    \end{tabular}}
\end{table}

Proximal Policy Optimization (PPO) optimizes the stochastic policy $\pi_\theta$ using a clipped surrogate objective to ensure stable updates:
\begin{equation}
L_{\text{PPO}}(\theta) = \mathbb{E}_t \left[ \min \big( r_t(\theta) \hat{A}_t, \text{clip}(r_t(\theta), 1 - \epsilon, 1 + \epsilon) \hat{A}_t \big) \right],
\end{equation}
where $r_t(\theta)$ is the policy ratio and $\epsilon$ is a clipping parameter (set to 0.2). This objective is combined with a value function loss and entropy regularization to balance exploration and exploitation.

\begin{figure*}[h]
    \centering
    \begin{minipage}[b]{0.48\textwidth}
    \centering
    \includegraphics[width=\textwidth]{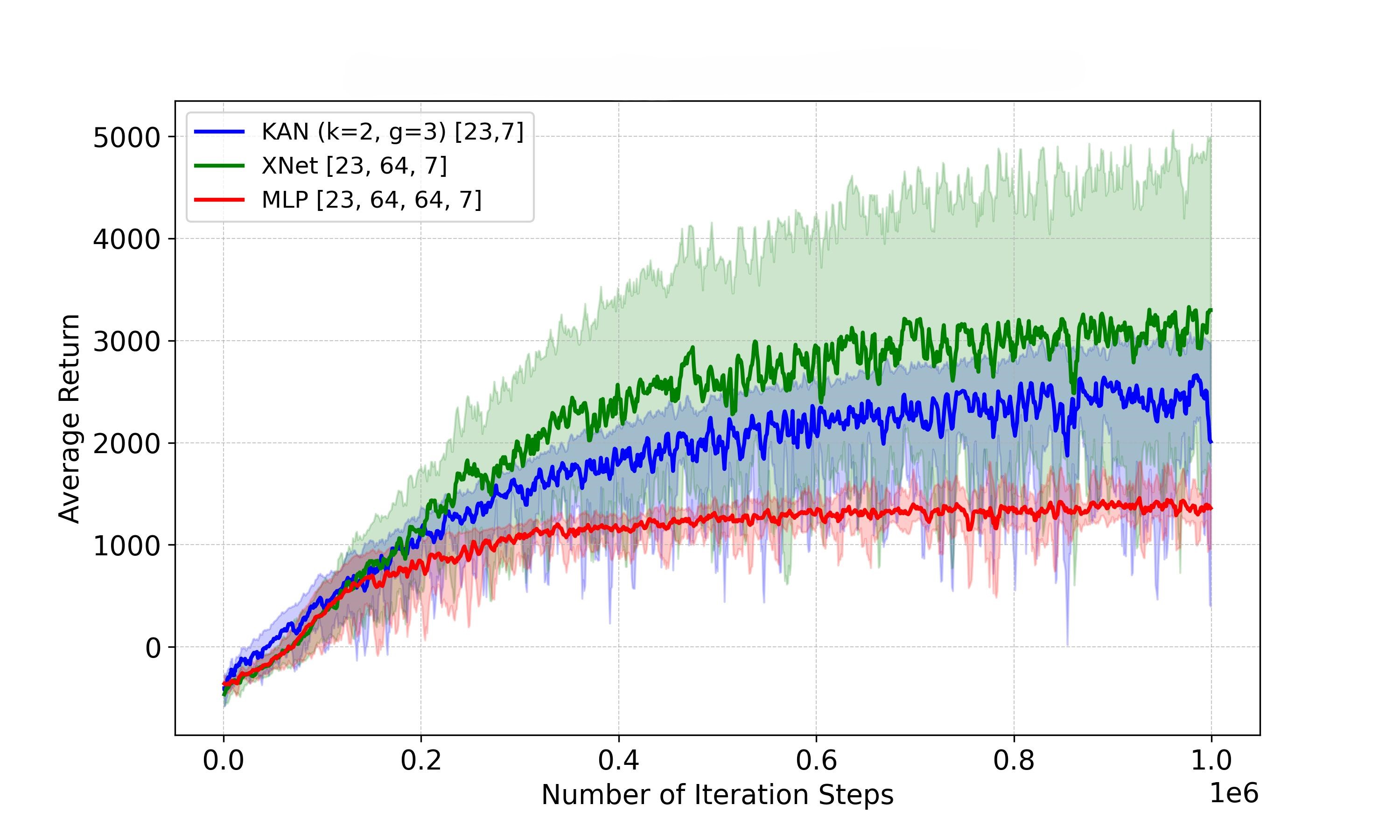}
    \put(-150,-8){\footnotesize (a) HalfCheetah-v4}
    \end{minipage}
    \hfill
    \begin{minipage}[b]{0.48\textwidth}
    \centering
    \includegraphics[width=\textwidth]{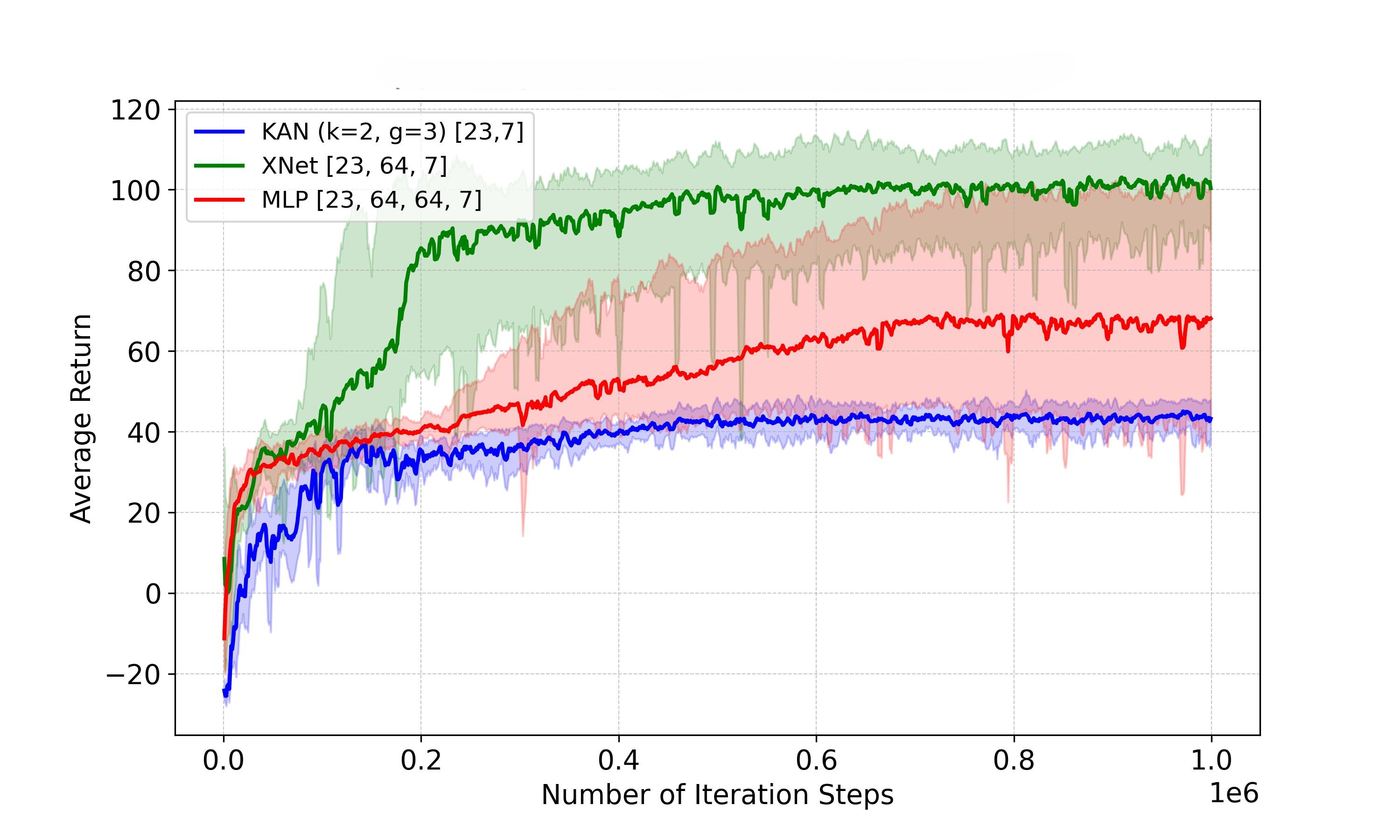}
    \put(-150,-8){\footnotesize (b) Swimmer-v4}
    \end{minipage}
    \caption{Reward comparison for PPO training across environments.}
    \label{fig:Reward_average_comparison}
\end{figure*}

Table \ref{tab:dmc_scores} and Figure \ref{fig:Reward_average_comparison} summarize the results of the three models on the two tasks. XNet consistently outperforms MLP and KAN in both environments, achieving the highest rewards and faster convergence. Notably, in \textit{HalfCheetah-v4}, XNet improves the reward by 64\% compared to KAN and 142\% compared to MLP. In \textit{Swimmer-v4}, XNet achieves a reward improvement of 47\% over MLP and 28\% over KAN.

\begin{table}[h]
    \centering
    \caption{Final PPO performance scores for control tasks after 1M training steps.}
    \label{tab:dmc_scores}
    \resizebox{0.5\textwidth}{!}{
    \begin{tabular}{lccc}
        \toprule
        \textbf{Task} & \textbf{KAN} & \textbf{MLP} & \textbf{XNet} \\
        \midrule
        HalfCheetah-v4 & 2010.52 & 1358.49 & \textbf{3298.52} \\
        Swimmer-v4     &  43.25  &  68.08  & \textbf{100.38} \\
        \bottomrule
    \end{tabular}}
\end{table}

\paragraph{Theoretical Analysis of XNet's Advantage.} 
The superior performance of XNet in reinforcement learning can be attributed to its unique Cauchy kernel activation functions, which possess three critical advantages:

1. \textbf{Localized Response with Fast Decay:} Cauchy kernels exhibit a localized response in their basis functions, which improves the network's ability to adapt to complex policy and value functions with sharp variations in state-action spaces. This property is especially advantageous in continuous control tasks, where such variations are common.

2. \textbf{High-Order Approximation Capability:} As established in Theorem \ref{Cauchy_approximation_theorem}, XNet achieves arbitrarily high-order approximation for analytic functions, allowing it to represent policies and value functions with higher precision compared to MLPs and KANs.

3. \textbf{Reduced Parameter Complexity:} Unlike KAN, which requires a large number of parameters to achieve similar accuracy, XNet leverages its basis function design to achieve better performance with fewer parameters. This reduced complexity directly translates to faster convergence in RL tasks.

These theoretical advantages align well with the policy gradient framework used in PPO, where the policy \( \pi_\theta(a|s) \) and the value function \( V(s) \) need to be accurately approximated for effective optimization. The experimental results in Table \ref{tab:dmc_scores} and Figure \ref{fig:Reward_average_comparison} further validate these theoretical insights.

\section{Summary and Outlook} \label{sec:conclusion}

Through comprehensive theoretical and empirical studies, we have thoroughly evaluated XNet's capabilities across function approximation, physics-informed learning, and reinforcement learning. Our key findings and future directions are summarized below.

\subsection{Key Findings}

\begin{itemize}
    \item \textbf{Superior Function Approximation}  
    Our experiments demonstrate that XNet significantly outperforms the recently proposed KAN in function approximation tasks. Notably, XNet maintains high computational efficiency while achieving superior accuracy, particularly for discontinuous and high-dimensional functions. Across all evaluation metrics (MSE, RMSE, MAE), XNet consistently surpasses KAN while reducing computational time. In high-dimensional function tests, XNet achieves up to a \textbf{1000-fold improvement in MSE} compared to KAN.

    \item \textbf{Enhanced Physics-Informed Learning}  
    Within the PINN framework, XNet exhibits two key advantages over traditional neural networks and KAN:  
    (i) \textbf{Higher solution accuracy}, reducing MSE by a factor of 50 when solving the Heat equation.  
    (ii) \textbf{Lower computational cost}, requiring only \textbf{40\% of KAN's training time}.  
    These results highlight XNet's potential to improve the efficiency of physics-informed neural networks.

    \item \textbf{Effective Reinforcement Learning Integration}  
    Integrating XNet into the PPO framework, we observe superior performance in reinforcement learning tasks on DeepMind Control Suite environments (HalfCheetah-v4 and Swimmer-v4). XNet-based PPO models achieve faster convergence and higher final rewards, reaching scores of \textbf{3298.52} and \textbf{100.38}, respectively, substantially outperforming MLP- and KAN-based implementations. These findings indicate XNet's effectiveness in learning complex control strategies.
\end{itemize}

\subsection{Future Directions}

While XNet demonstrates substantial advantages over existing architectures, several research challenges remain:

\begin{itemize}
    \item \textbf{Extending XNet for Large-Scale Architectures}  
    Investigating its integration with transformer-based models to improve sequence modeling and time-series analysis.
    
    \item \textbf{Expanding Applications in AI and Scientific Computing}  
    Exploring XNet’s potential in generative modeling, computer vision, and physics-informed learning.
    
    \item \textbf{Optimizing Numerical Stability and Scalability}  
    Enhancing stability in extremely high-dimensional settings and improving computational efficiency.
    
    \item \textbf{Theoretical Advancements}  
    Conducting deeper theoretical analysis of XNet’s approximation properties and convergence behavior.
\end{itemize}

By addressing these challenges, XNet can further contribute to advancing neural network architectures and broadening their applications in artificial intelligence and computational mathematics.

\bibliography{final1001}
\bibliographystyle{iclr2025_conference}

\appendix
\section{Appendix}

\subsection{Mathematical Analysis}\label{ref:th_proof}
\paragraph{Derivation of Approximation Rate.}

For any fixed integer $p$, if $f$ is $C^p$ smooth, the error bound follows from Cauchy approximations theorem:
\begin{equation}
\| f - f_N \|_{\infty} \leq C(p) N^{-p}, \quad \forall p > 0,
\end{equation}
where \( C(p) \) is a constant dependent on \( p \), which accounts for the smoothness of \( f \) and the domain geometry. The Cauchy activation functions are derived from one-dimensional Cauchy approximations, therefore, XNet has high order approximation properties. 

To ensure an approximation error of at most \( \epsilon \), we require
\begin{equation}
C(p) N^{-p} \leq \epsilon.
\end{equation}
Rearranging, we obtain
\begin{equation}
N^{-p} \leq \frac{\epsilon}{C(p)}.
\end{equation}
Taking the reciprocal and raising both sides to the power of \( 1/p \), we derive
\begin{equation}
N \geq \left( \frac{C(p)}{\epsilon} \right)^{1/p}.
\end{equation}
Thus, the number of required basis functions satisfies
\begin{equation}
N = O(1/\epsilon^{1/p}).
\end{equation}

For B-splines of degree \( k \), approximation theory gives:
\begin{equation}
\| f - f_N \|_{\infty} \leq C N^{-k}.
\end{equation}

Rearranging for \( N \) to achieve error \( \epsilon \):
\begin{equation}
N \geq \left( \frac{C}{\epsilon} \right)^{1/k} = O(1/\epsilon^{1/k}).
\end{equation}
This confirms that B-splines require polynomial growth in \( 1/\epsilon \), whereas Cauchy kernels allow arbitrarily fast convergence depending on \( p \).

\paragraph{Derivation of Computational Efficiency}

For Cauchy kernels, the approximation theorem directly states that \( \varepsilon = O(N^{-p}) \), implying that:
\begin{equation}
N_{\text{Cauchy}} = O(\epsilon^{-1/p}).
\end{equation}
For B-splines of degree \( k \), classical approximation theory suggests that for functions with \( k \) continuous derivatives, the approximation error is \( \varepsilon = O(N^{-k}) \), leading to:
\begin{equation}
N_{\text{B-spline}} = O(\epsilon^{-1/k}).
\end{equation}
This direct comparison highlights the inherent flexibility and potential efficiency of Cauchy kernels over B-splines, especially in scenarios requiring rapid convergence.

\paragraph{Proof of Numerical Characteristics:}

\textbf{Matrix Structure and Stability:}
\begin{itemize}
    \item \textbf{Cauchy Kernels:}
    The interaction matrix \( A \) formed by Cauchy kernels is dense and its elements are given by \( A_{ij} = \frac{1}{\xi_i - z_j} \). Using Gershgorin's Circle Theorem, we estimate the eigenvalues of \( A \) by:
    \begin{equation}
    |\lambda - A_{ii}| \leq \sum_{j \neq i} |A_{ij}| = \sum_{j \neq i} \left|\frac{1}{\xi_i - z_j}\right|,
    \end{equation}
    which typically grows with the number of nodes \( N \), leading to a condition number \( \kappa(A) = O(N) \). This indicates a potential increase in numerical instability as problem size increases, though this can often be mitigated with modern computational techniques such as preconditioning.

    \item \textbf{B-Splines:}
    The matrix structure for B-splines is sparse and banded due to their local support. This structure naturally leads to a lower condition number, typically \( O(1) \), enhancing numerical stability and making B-splines particularly well-suited for large-scale problems.
\end{itemize}

\textbf{Derivative Properties:}
\begin{itemize}
    \item The derivatives of Cauchy kernels with respect to \( d^2 \) are given by:
    \begin{equation}
    \frac{d^n}{d(d^2)^n} \left( \frac{1}{x^2 + d^2} \right) = (-1)^n \frac{(n-1)!}{(x^2 + d^2)^{n+1}}.
    \end{equation}
    This explicit formula enables efficient and precise computation of derivatives with respect to \( d^2 \), which is particularly beneficial in optimization and adaptive learning settings.

    \item B-splines, on the other hand, involve piecewise polynomial functions, where computing higher-order derivatives requires tracking piecewise regions. Each derivative operation reduces the degree of the spline, leading to increased computational complexity when multiple derivatives are required.
\end{itemize}

Detailed mathematical proofs and additional theoretical discussions on these characteristics are presented in Appendix \ref{ref:th_proof}, providing a rigorous foundation for the discussed numerical properties and their implications in practical applications.

\subsection{ADDITIONAL EXPERIMENT DETAILS} 
The numerical experiments presented below were performed in Python using the Tensorflow-CPU processor on a Dell computer equipped with a 3.00 Gigahertz (GHz) Intel Core i9-13900KF.
When detailing grids ans k for KAN models, we always use values provided by respective authors (Kan).

\subsection{FUNCTION APPROXIMATION} \label{A.1}
For 1d heaciside function, we set different configurations. The results are shown as follows
\begin{table}[h!]
\centering
\caption{B-Spline Performance metrics comparison for different G and K values. reference}
\resizebox{0.5\textwidth}{!}{
\begin{tabular}{|c|c|c|c|}
\hline
\multicolumn{4}{|c|}{B-Spline} \\
\hline
\textbf{k, G} & \textbf{MSE} & \textbf{RMSE} & \textbf{MAE} \\
\hline
k=50, G=200 & 5.8477e-01 & 7.6470e-01 & 6.1076e-01 \\
k=3, G=10 & 9.2871e-03 & 9.6369e-02 & 4.7923e-02 \\
k=3, G=50 & 2.3252e-03 & 4.8221e-02 & 1.2255e-02 \\
k=10, G=50 & 1.9881e-03 & 4.4588e-02 & 1.0879e-02 \\
k=3, G=200 & 1.1252e-03 & 3.3544e-02 & 4.4737e-03 \\
k=10, G=200 & 1.1029e-03 & 3.3210e-02 & 5.1904e-03 \\
\hline
\end{tabular}}

\end{table}

\begin{table}[htbp]
	\centering
	\caption{KAN  reference}
	\label{table:kan_1d}
	\resizebox{0.7\textwidth}{!}{%
		\begin{tabular}{ccccccc}
			\hline \hline
			& \multicolumn{3}{c}{[1,1]KAN} & \multicolumn{3}{c}{[1,3,1]KAN} \\
			\midrule
			k,G  & MSE & RMSE  & MAE & MSE & RMSE  & MAE \\
			\midrule
			k=3, G=3 & 2.20E-02 & 1.48E-01 & 9.89E-02  & 3.50E-04 & 1.87E-02 & 5.56E-03 \\
			k=3, G=10 & 1.22E-02 & 1.10E-01 & 5.91E-02 & 1.84E-04 & 1.36E-02 & 2.54E-03  \\
			k=3, G=50 & 2.44E-03 & 4.94E-02 & 1.22E-02 & 4.28E-05 & 6.55E-03 & 2.71E-03 \\
			k=3, G=200 & 5.98E-04 & 2.45E-02 & 3.03E-03 & 3.79E-04 & 1.95E-02 & 1.24E-02 \\
			\hline \hline
		\end{tabular}%
	}
	\label{dt_diffrate_results}
\end{table}

For 2d functions, loss function
\begin{figure}[h!]
    \centering
    \begin{minipage}[b]{0.48\textwidth}
        \centering
        \includegraphics[width=0.48\textwidth]{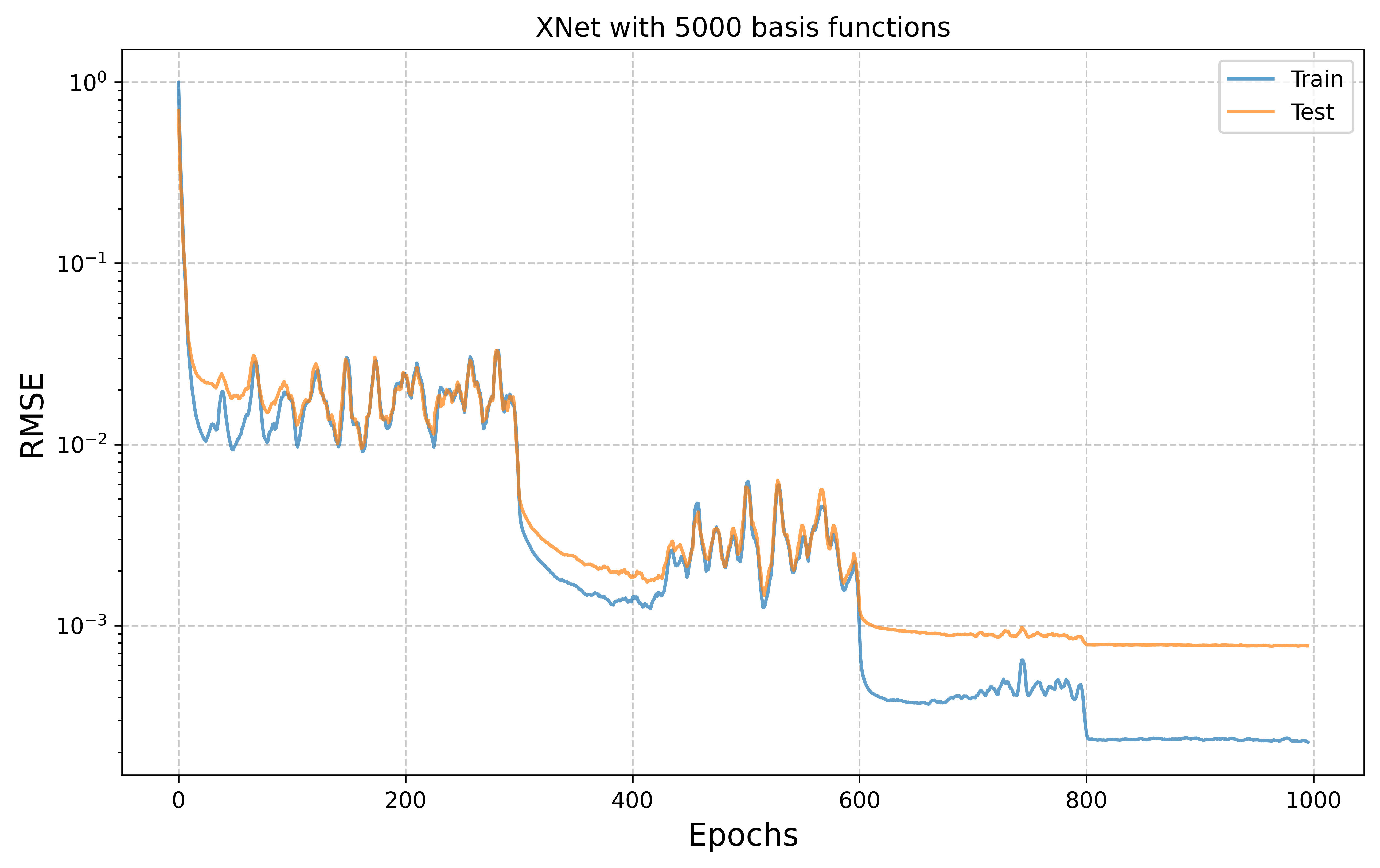}
        \hfill
        \includegraphics[width=0.48\textwidth]{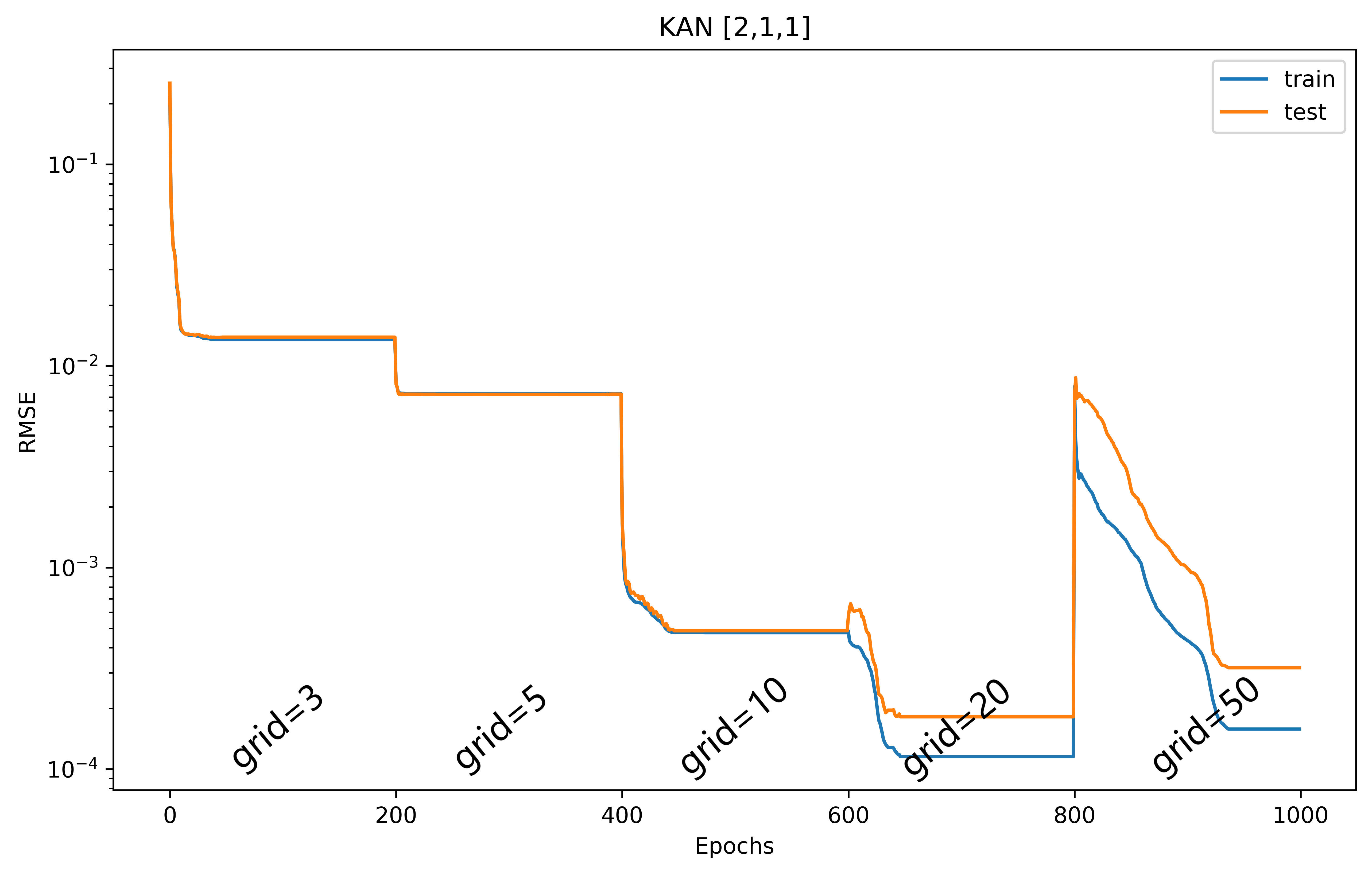}
        \captionof{figure}{Loss on \(\exp(\sin(\pi x) + y^2)\)}
    \end{minipage}
    \hfill
    \begin{minipage}[b]{0.48\textwidth}
        \centering
        \includegraphics[width=0.48\textwidth]{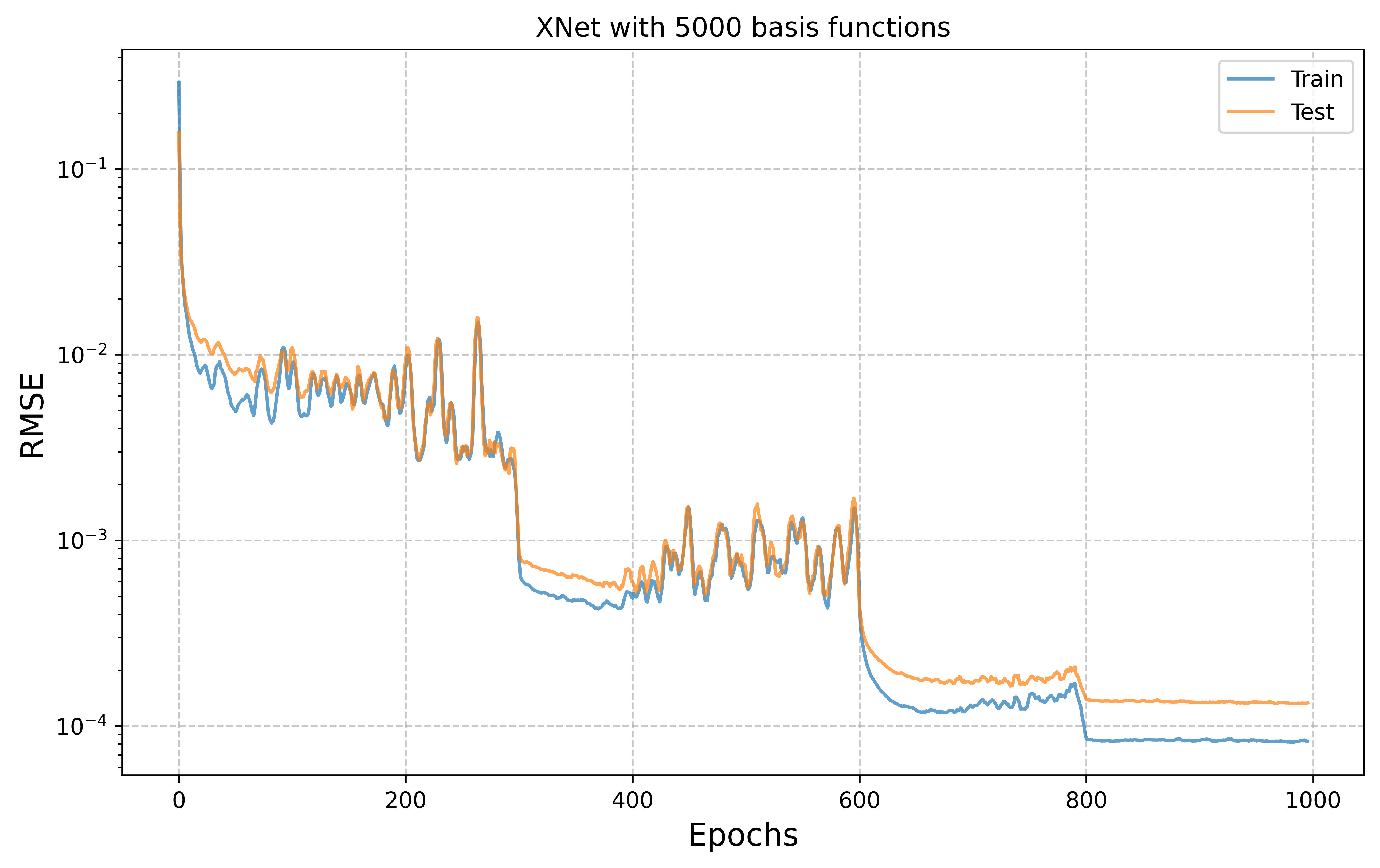}
        \hfill
        \includegraphics[width=0.48\textwidth]{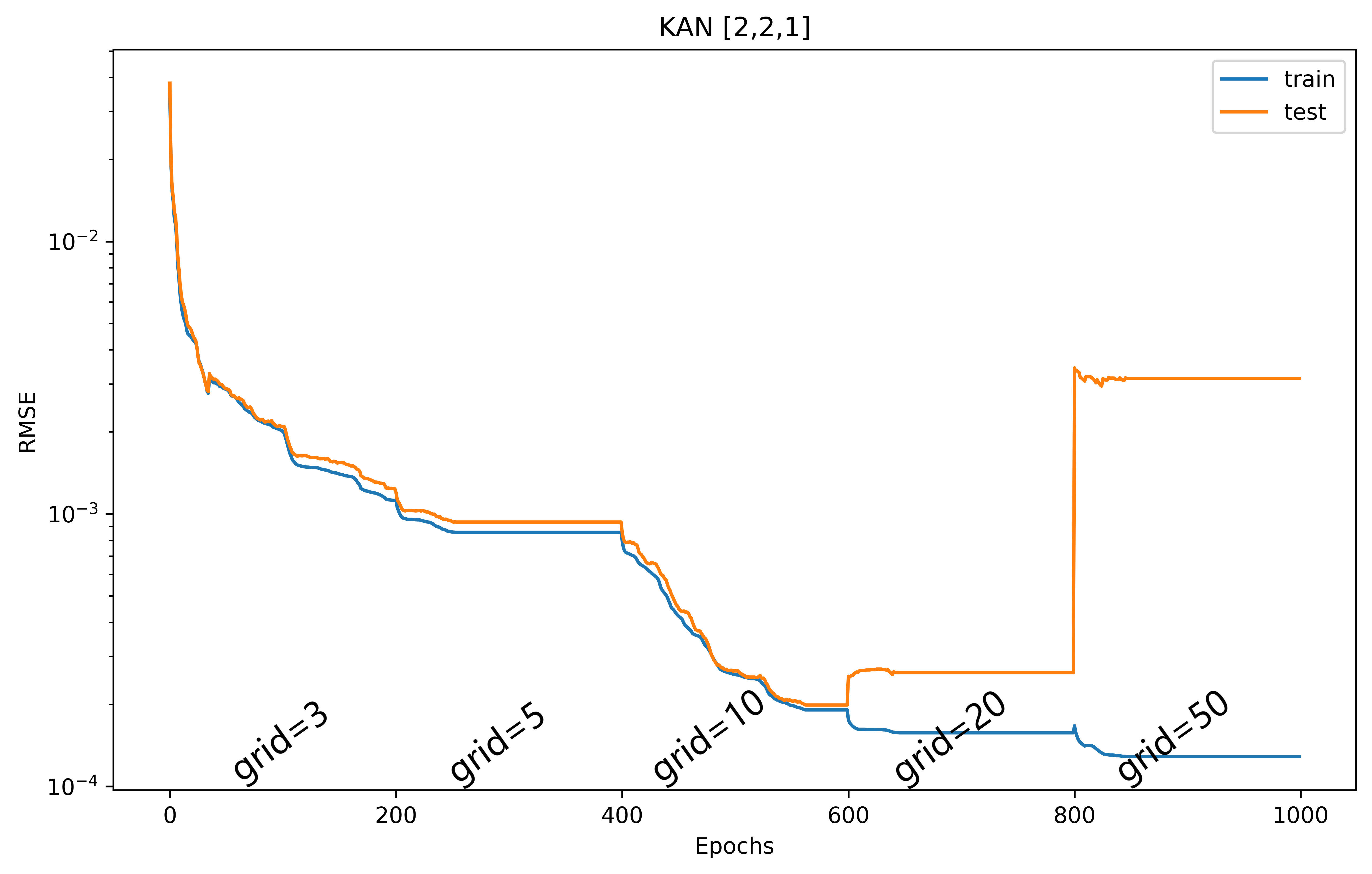}
        \captionof{figure}{Loss on \(xy\)}
    \end{minipage}
\end{figure}

for high-dimensional functions, loss functions
\begin{figure}[h!]
    \centering
    \begin{minipage}[b]{0.48\textwidth}
        \centering
        \includegraphics[width=0.48\textwidth]{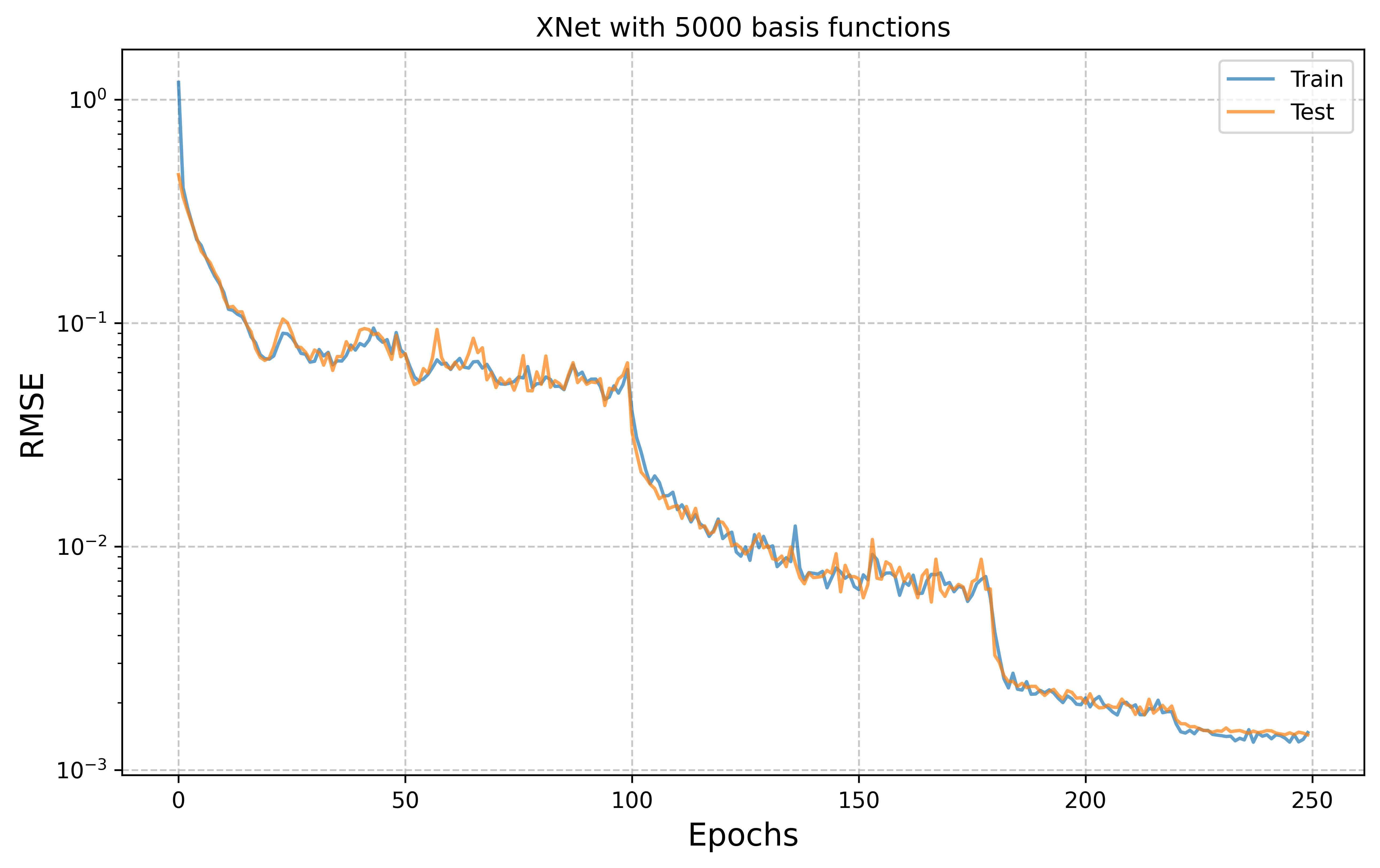}
        \hfill
        \includegraphics[width=0.48\textwidth]{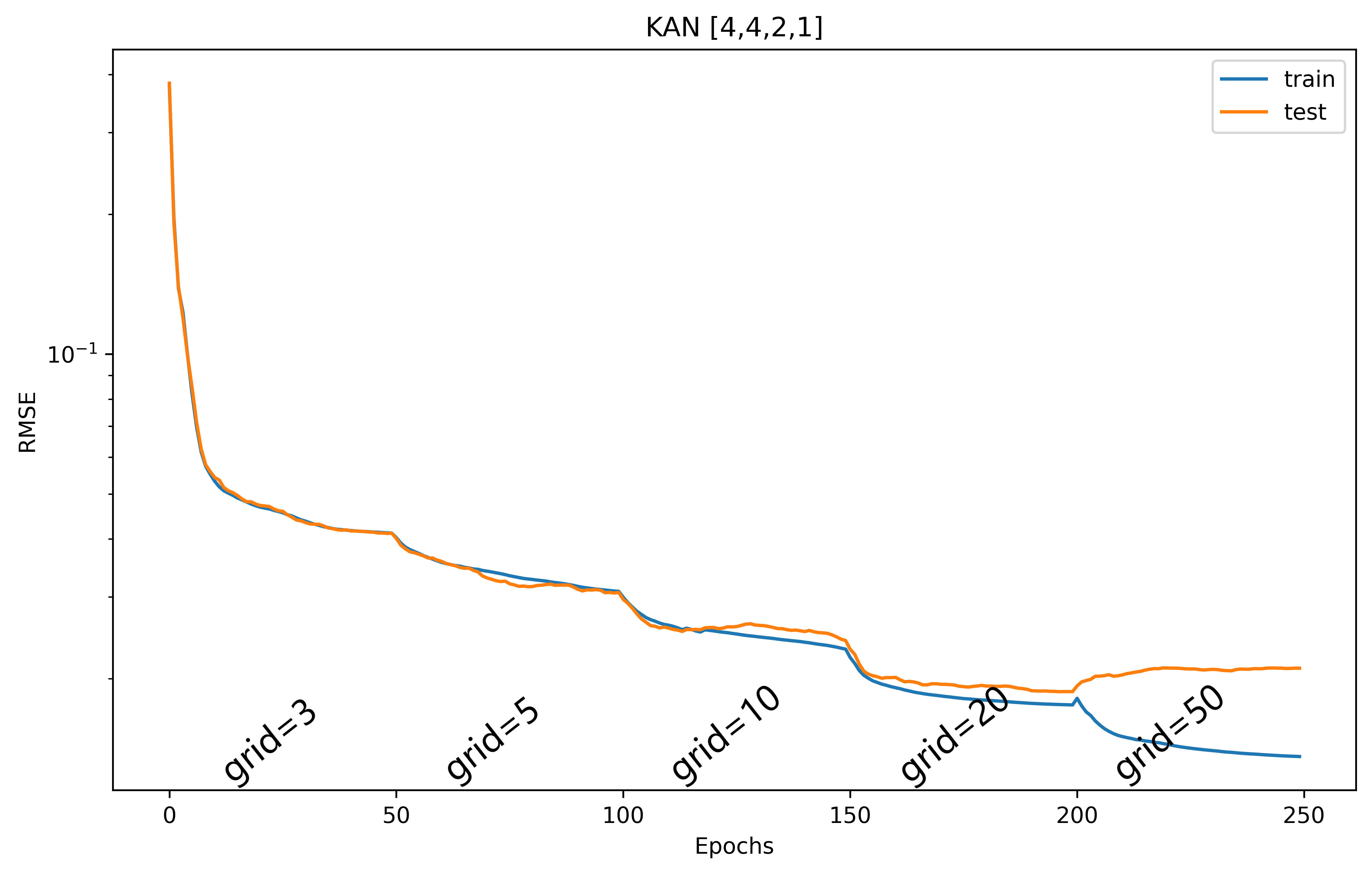}
        \caption*{ \( \exp\left(\frac{1}{2}\left(\sin\left(\pi(x_{1}^{2}+x_{2}^{2})\right) + x_{3}x_{4}\right)\right) \)}
    \end{minipage}
    \hfill
    \begin{minipage}[b]{0.48\textwidth}
        \centering
        \includegraphics[width=0.48\textwidth]{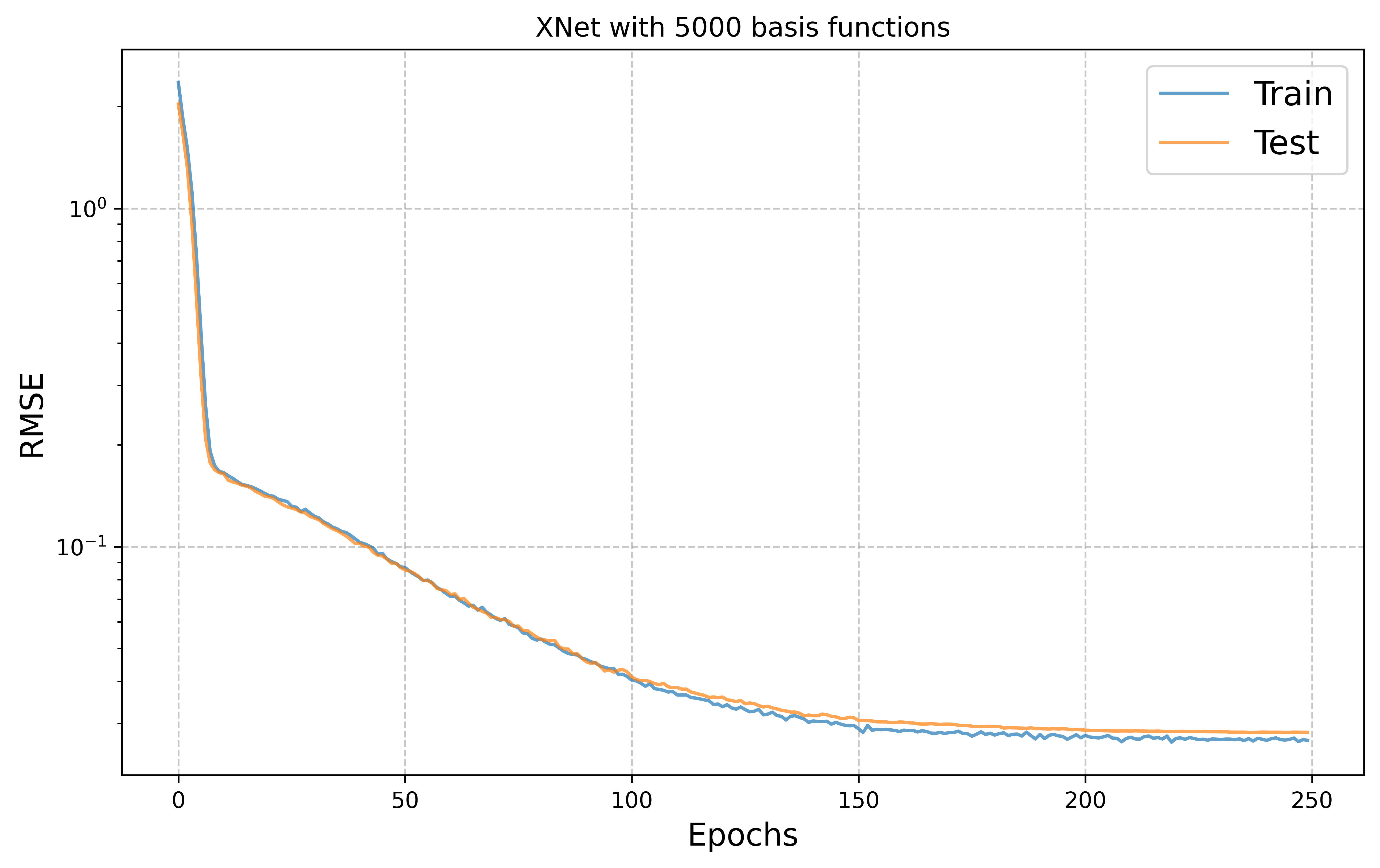}
        \hfill
        \includegraphics[width=0.48\textwidth]{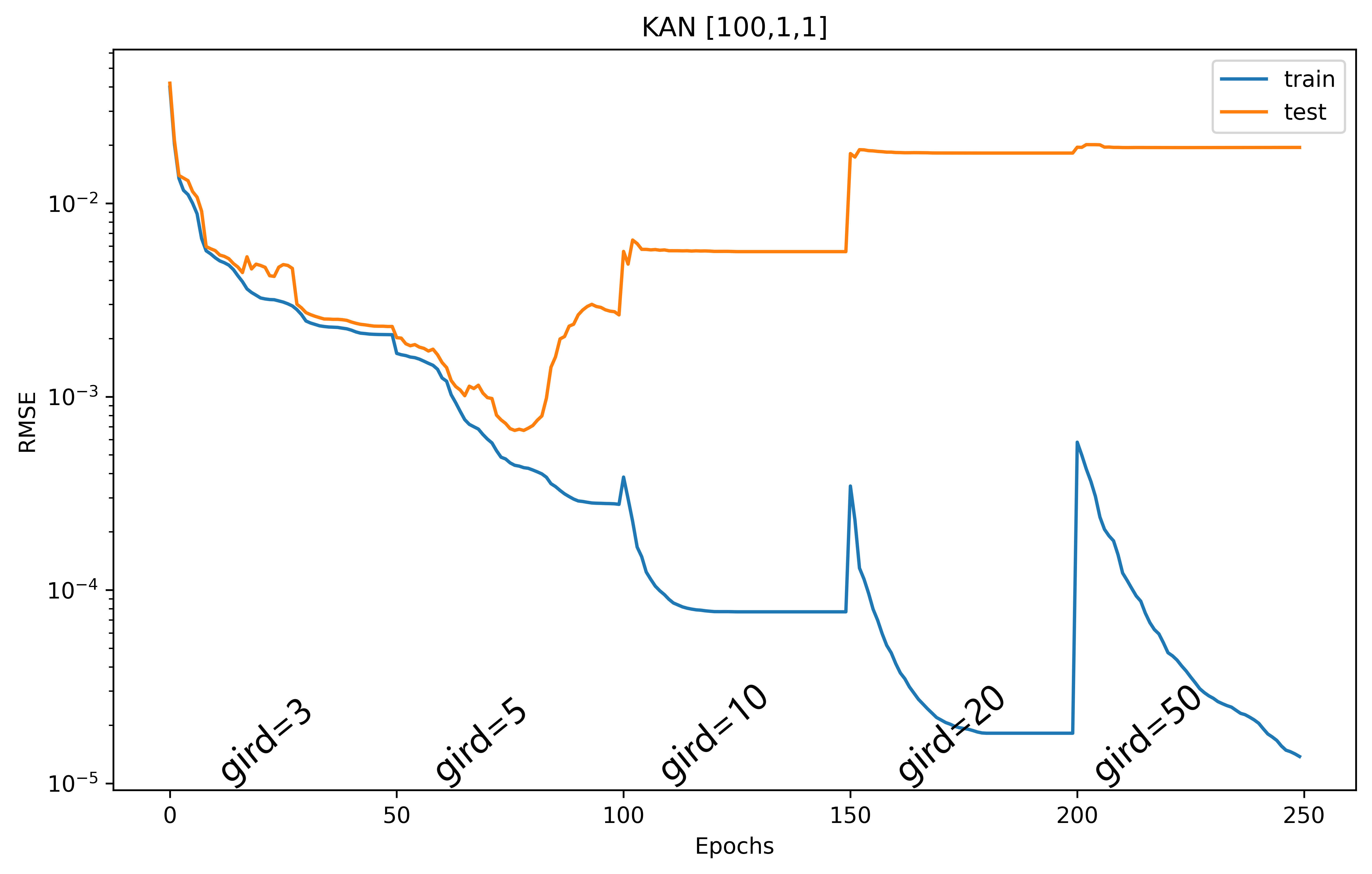}
        \caption*{\( \exp(\frac{1}{100}\sum_{i=1}^{100}\sin^2(\frac{\pi x_i}{2})) \)}
    \end{minipage}
	\caption{Loss on high-dimensional functions}
\end{figure}

\subsection{Function Approximation with noise}\label{ref:function noise}
\begin{figure*}[h!]
    \centering
    \begin{minipage}[b]{0.3\textwidth}
        \centering
        \includegraphics[width=\textwidth]{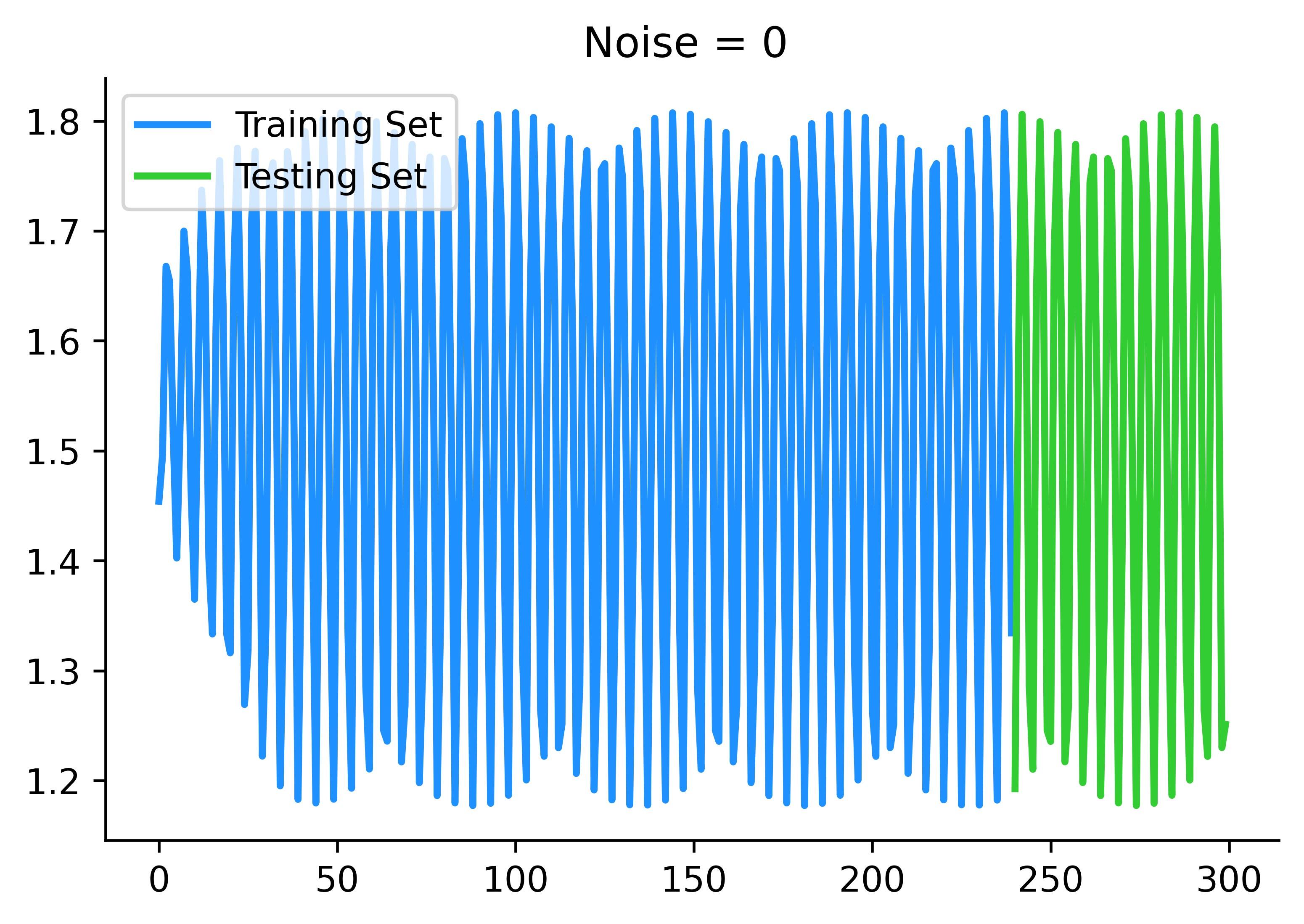}
        \put(-75,-10){\footnotesize (a) Noise = 0}
    \end{minipage}
    \hfill
    \begin{minipage}[b]{0.3\textwidth}
        \centering
        \includegraphics[width=\textwidth]{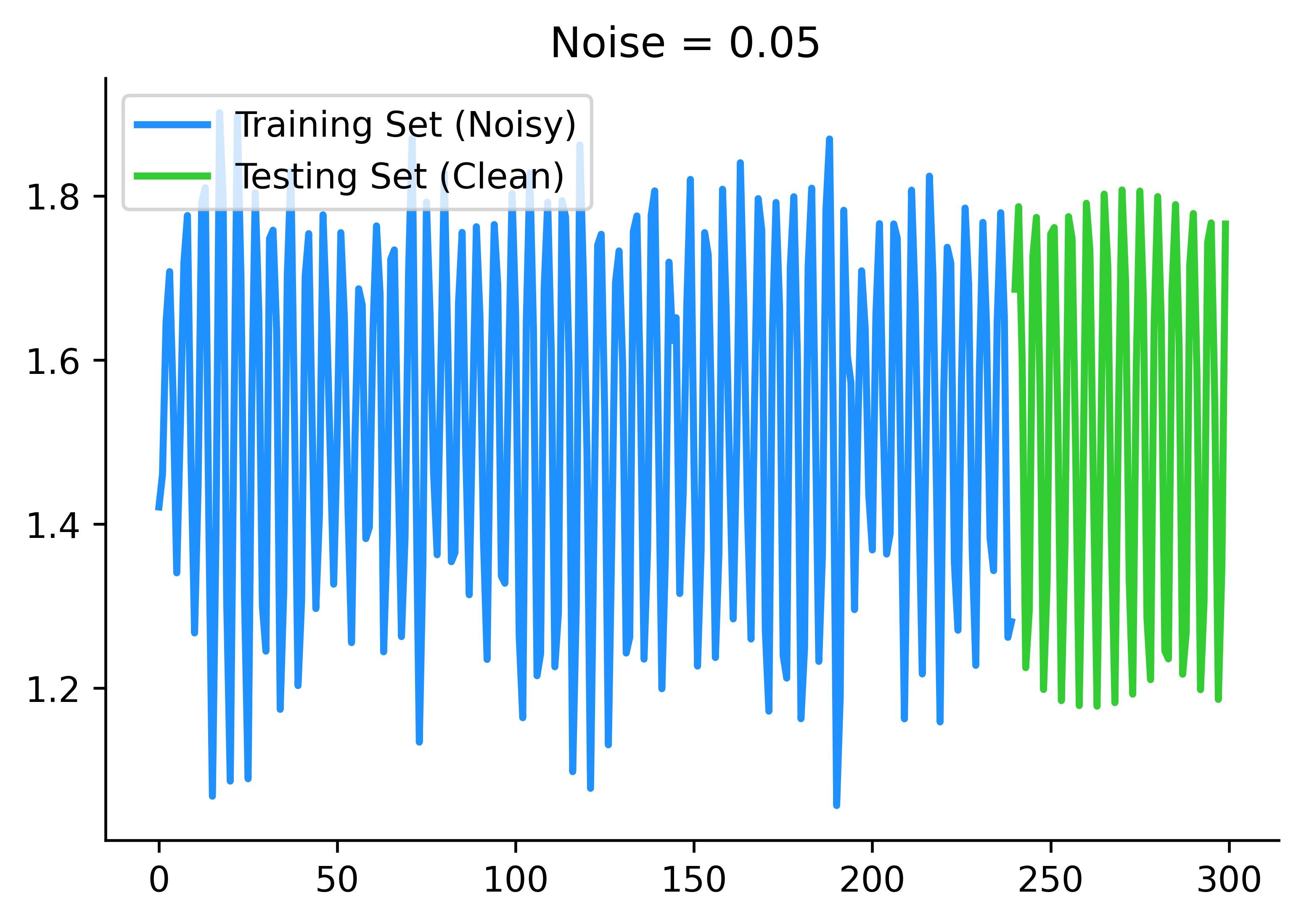}
        \put(-75,-10){\footnotesize (b) Noise = 0.05}
    \end{minipage}
    \hfill
    \begin{minipage}[b]{0.3\textwidth}
        \centering
        \includegraphics[width=\textwidth]{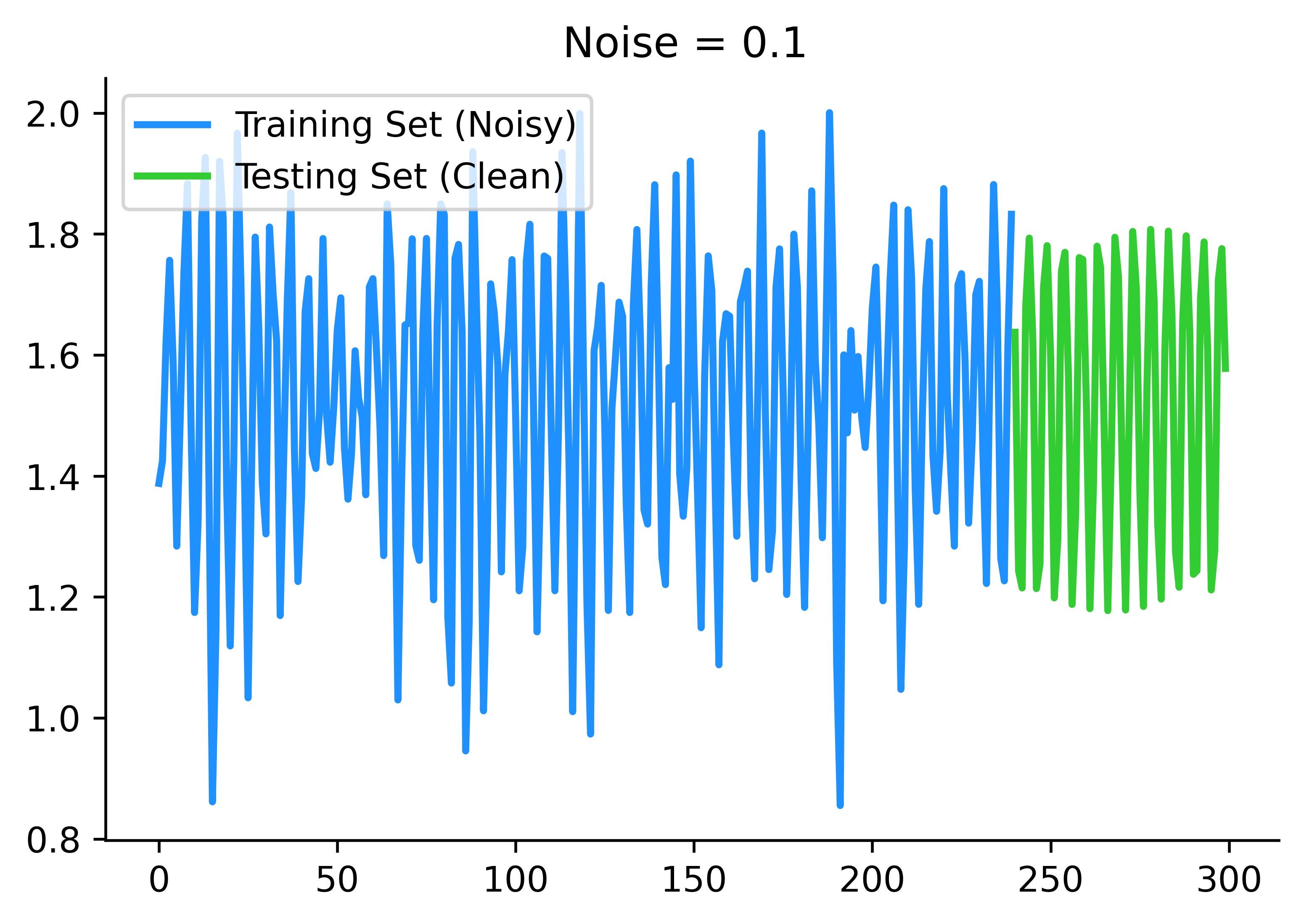}
        \put(-75,-10){\footnotesize (c) Noise = 0.1}
    \end{minipage}
    \caption{Function fitting tested on datasets with different noise levels: (a) Noise = 0, (b) Noise = 0.05, and (c) Noise = 0.1.}
    \label{fig:ts1_noise}
\end{figure*}

The system is governed by relatively simple functions under different noise levels (see figure \ref{fig:ts1_noise}). 

\begin{figure*}[h!]
    \centering
    \begin{minipage}[b]{0.30\textwidth}
        \centering
        \includegraphics[width=\textwidth]{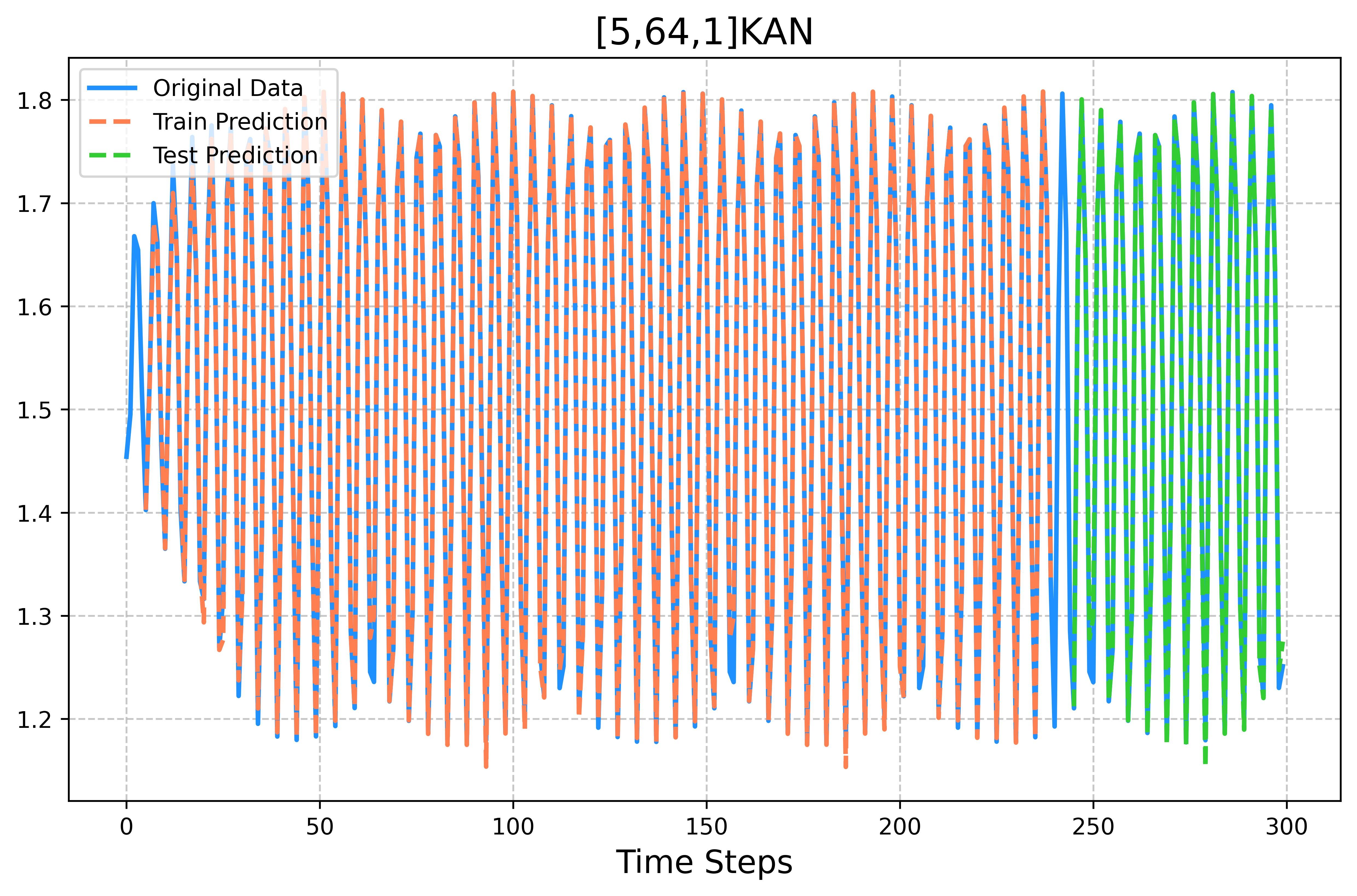}
        \put(-105,-10){KAN (Noise = 0)}
    \end{minipage}
    \hfill
    \begin{minipage}[b]{0.30\textwidth}
        \centering
        \includegraphics[width=\textwidth]{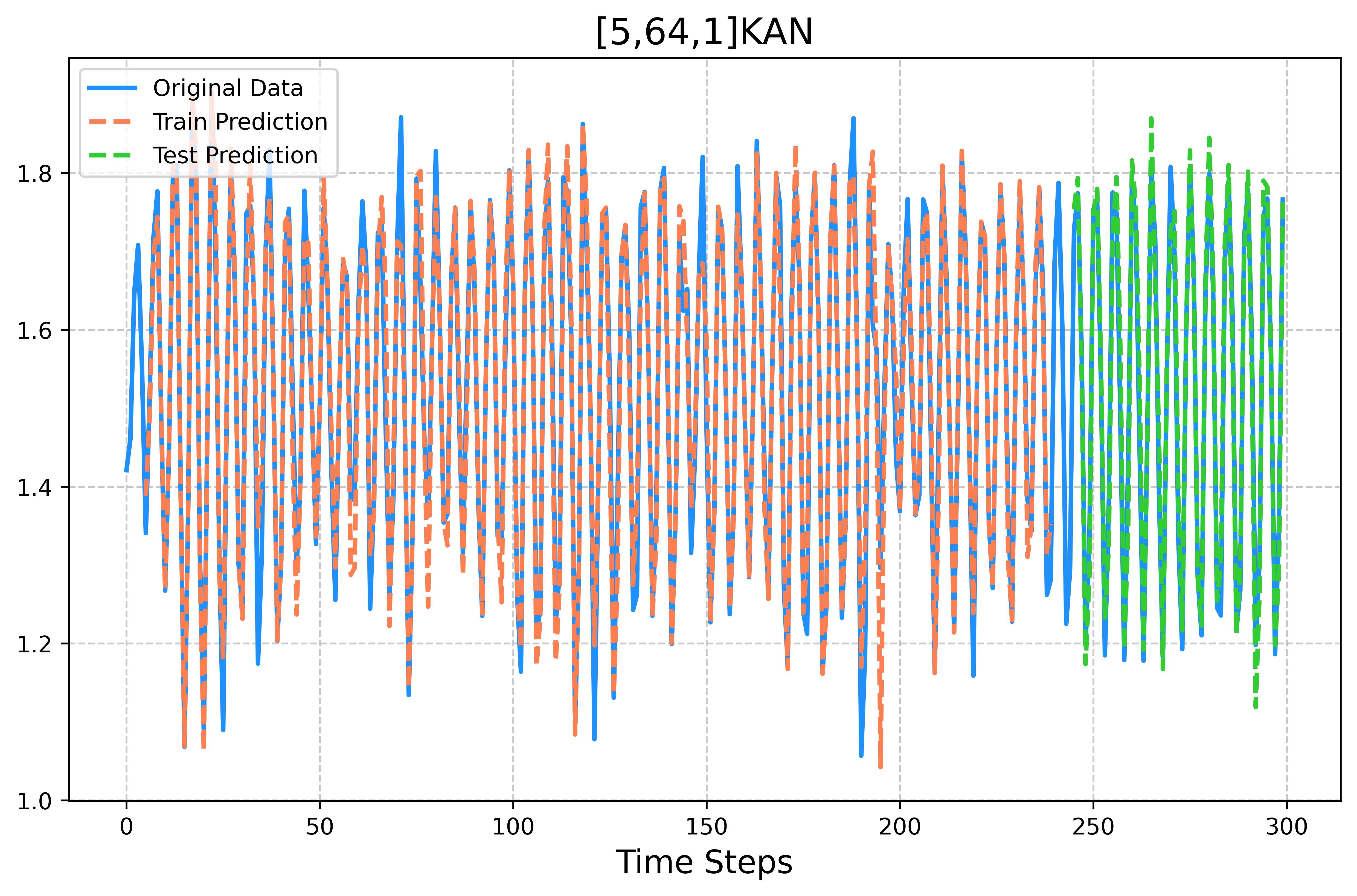}
        \put(-105,-10){KAN (Noise = 0.05)}
    \end{minipage}
    \hfill
    \begin{minipage}[b]{0.30\textwidth}
        \centering
        \includegraphics[width=\textwidth]{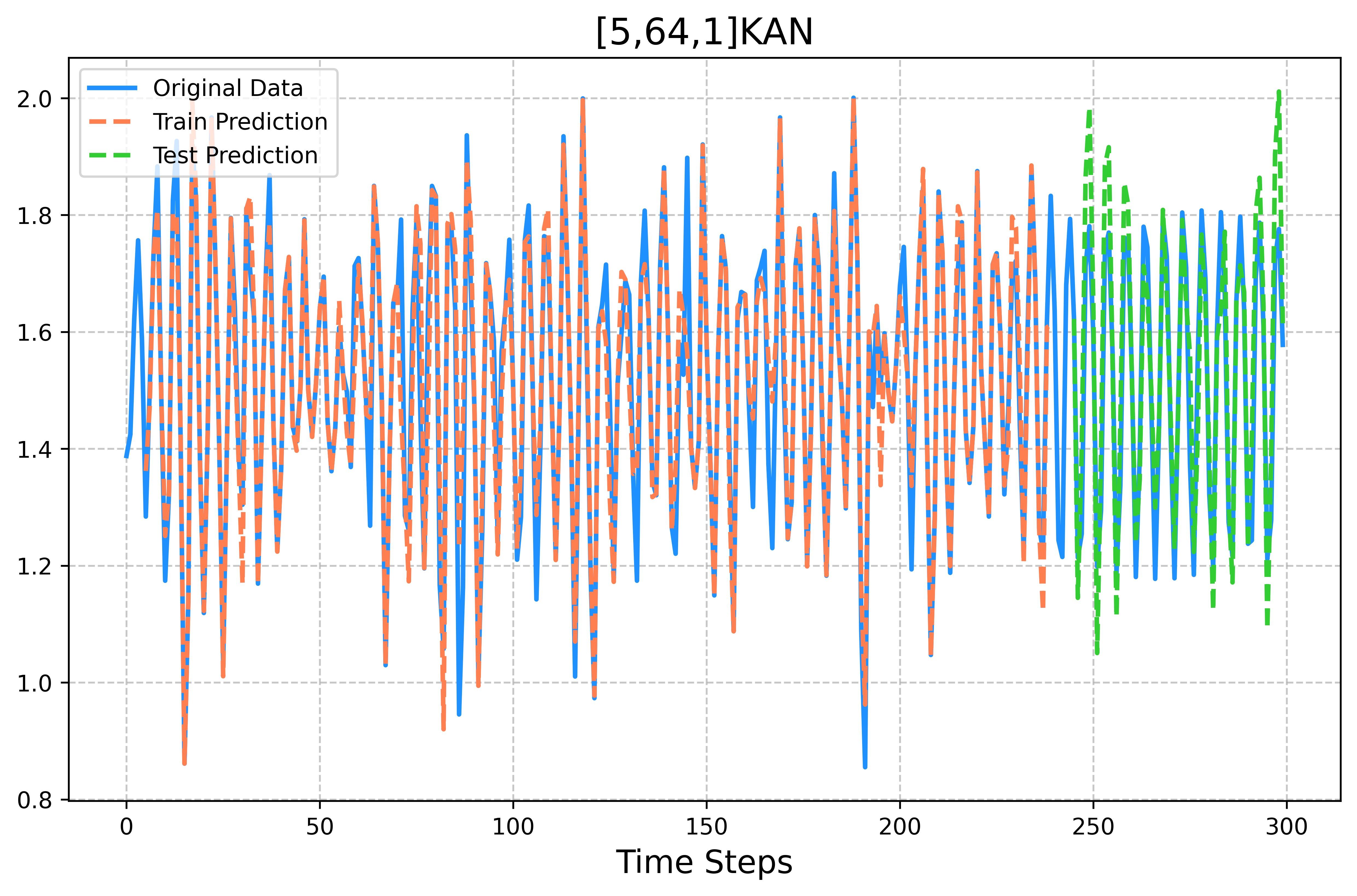}
        \put(-105,-10){KAN (Noise = 0.1)}
    \end{minipage}
    
    \vspace{0.5cm}
    \begin{minipage}[b]{0.30\textwidth}
        \centering
        \includegraphics[width=\textwidth]{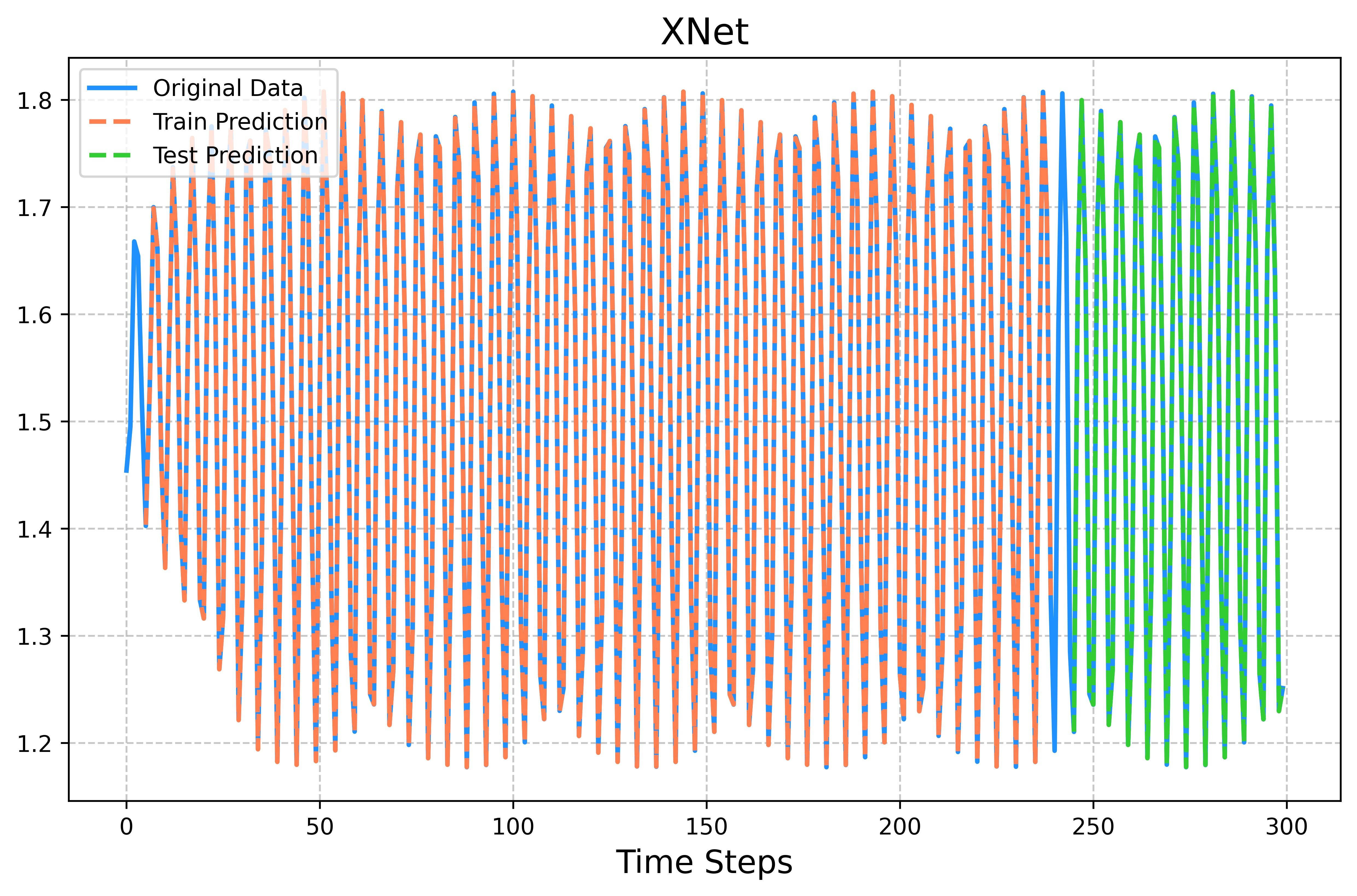}
        \put(-105,-10){XNet (Noise = 0)}
    \end{minipage}
    \hfill
    \begin{minipage}[b]{0.30\textwidth}
        \centering
        \includegraphics[width=\textwidth]{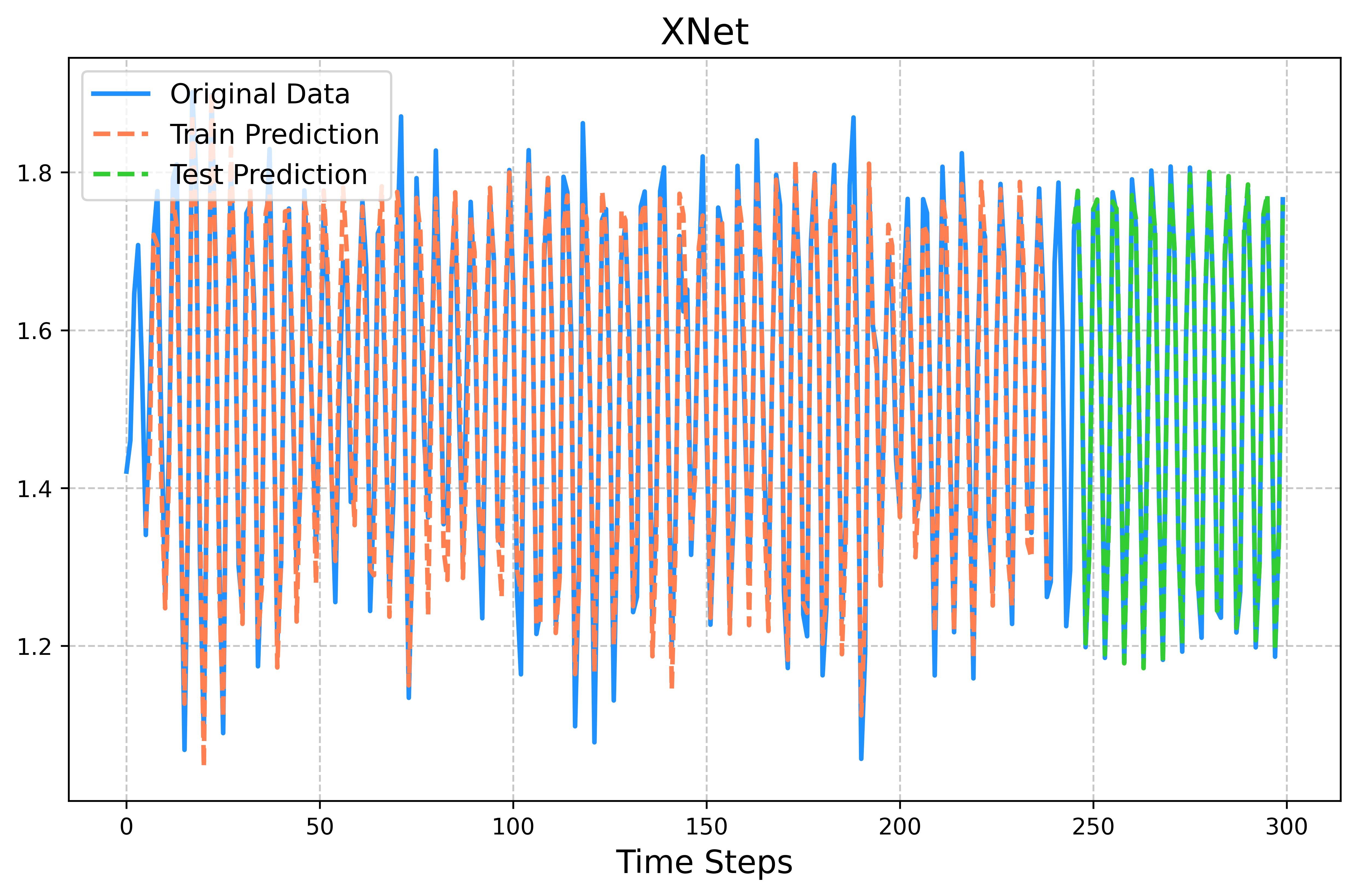}
        \put(-105,-10){XNet (Noise = 0.05)}
    \end{minipage}
    \hfill
    \begin{minipage}[b]{0.30\textwidth}
        \centering
        \includegraphics[width=\textwidth]{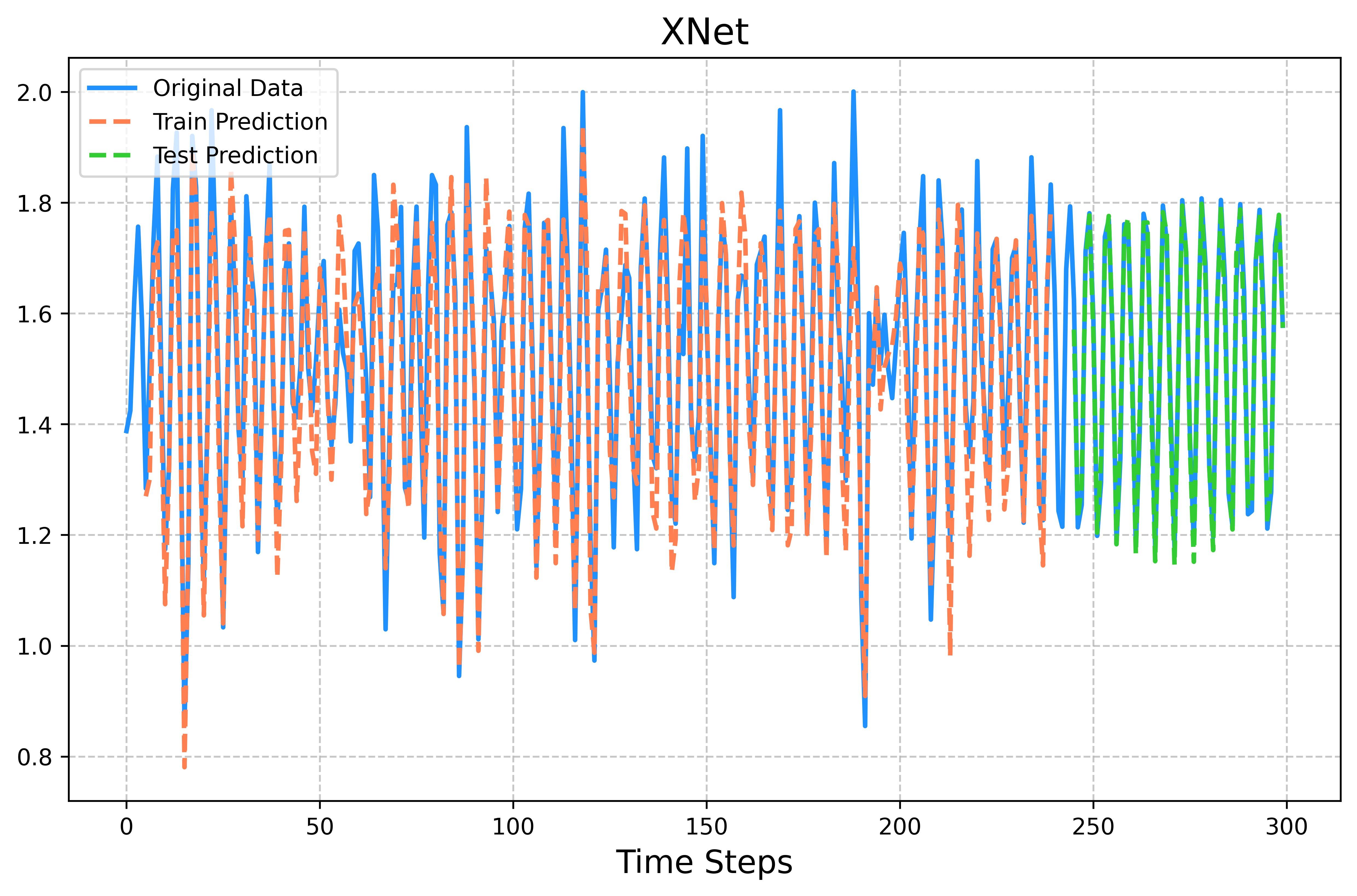}
        \put(-105,-10){XNet (Noise = 0.1)}
    \end{minipage}
    
    \caption{Comparison of the performance of KAN and XNet under different noise levels.}
    \label{fig:comparison_KAN_XNet_noise}
\end{figure*}

Figures \ref{fig:comparison_KAN_XNet_noise} show a comparison of the predictive
performance of [5, 64, 1] KAN and [5, 20, 1] XNet on three
scenarios: one with no noise (noise = 0), one with moderate
noise (noise = 0.05), and one with high noise (noise = 0.1).

\subsection{Heat equation}\label{ref:exp_NS_LOSS}
\begin{figure}[h!]
    \centering
    \begin{minipage}[b]{0.23\textwidth}
        \centering
        \includegraphics[width=\textwidth]{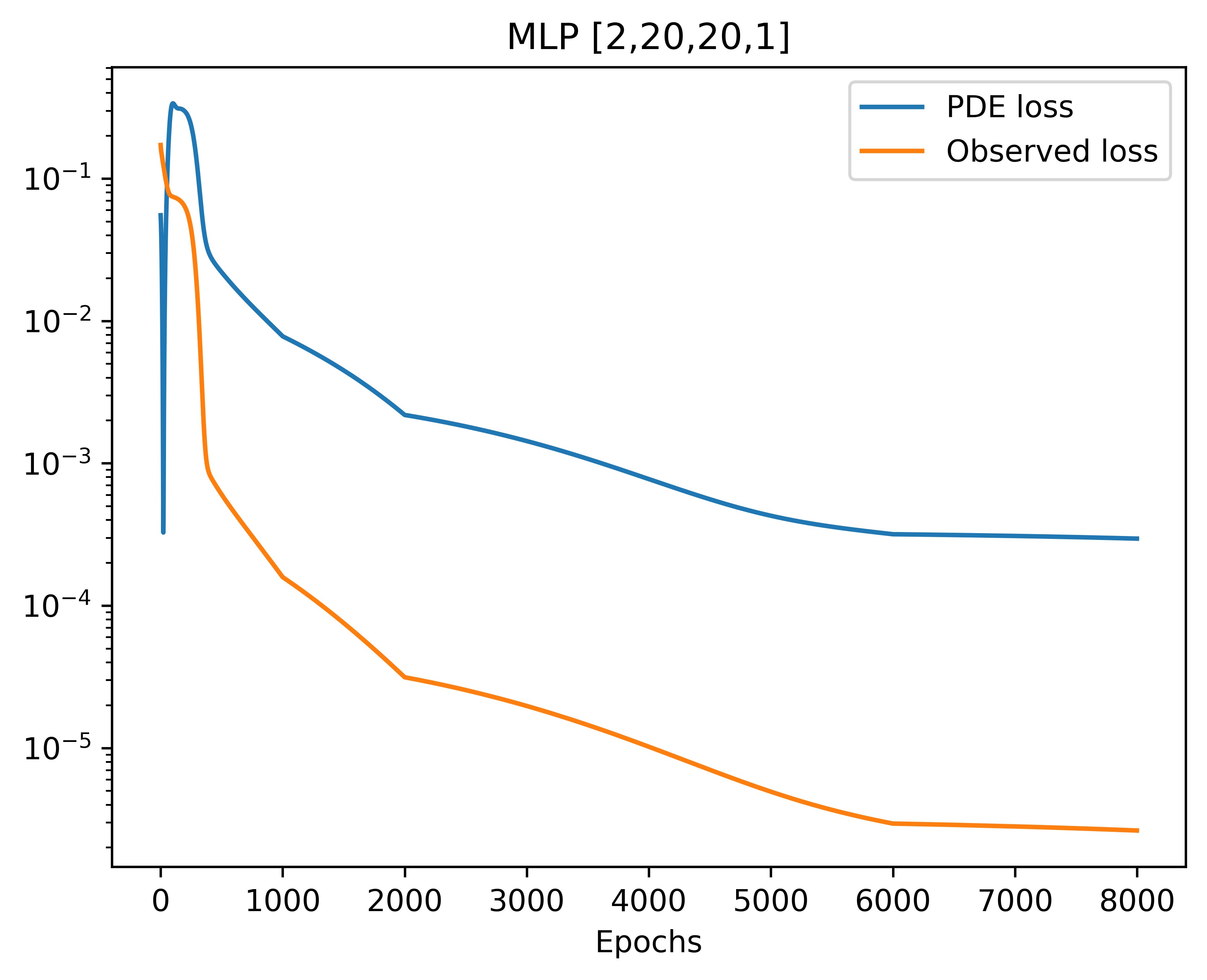}
    \end{minipage}
    \hfill
    \begin{minipage}[b]{0.23\textwidth}
        \centering
        \includegraphics[width=\textwidth]{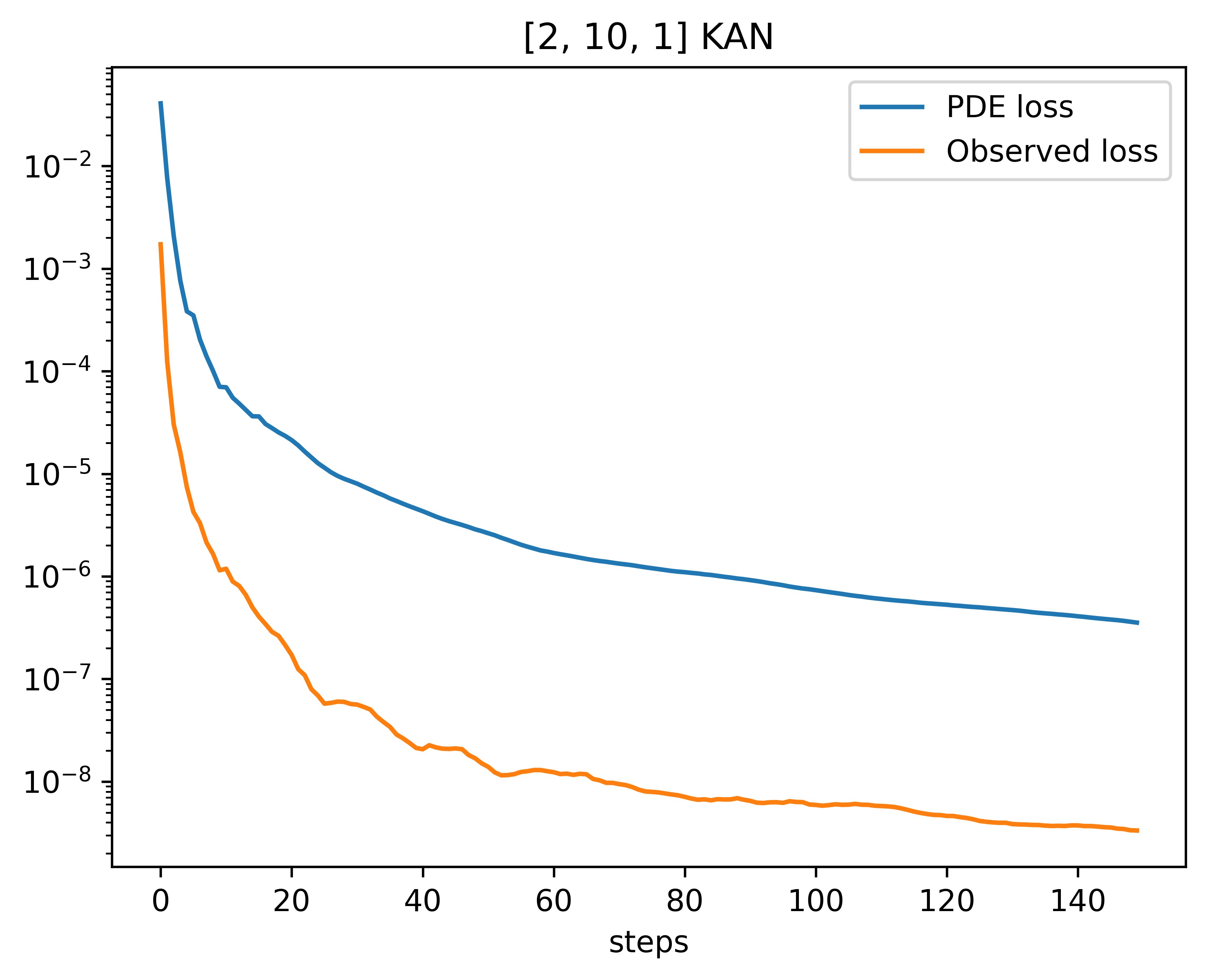}
    \end{minipage}
    \hfill
    \begin{minipage}[b]{0.23\textwidth}
        \centering
        \includegraphics[width=\textwidth]{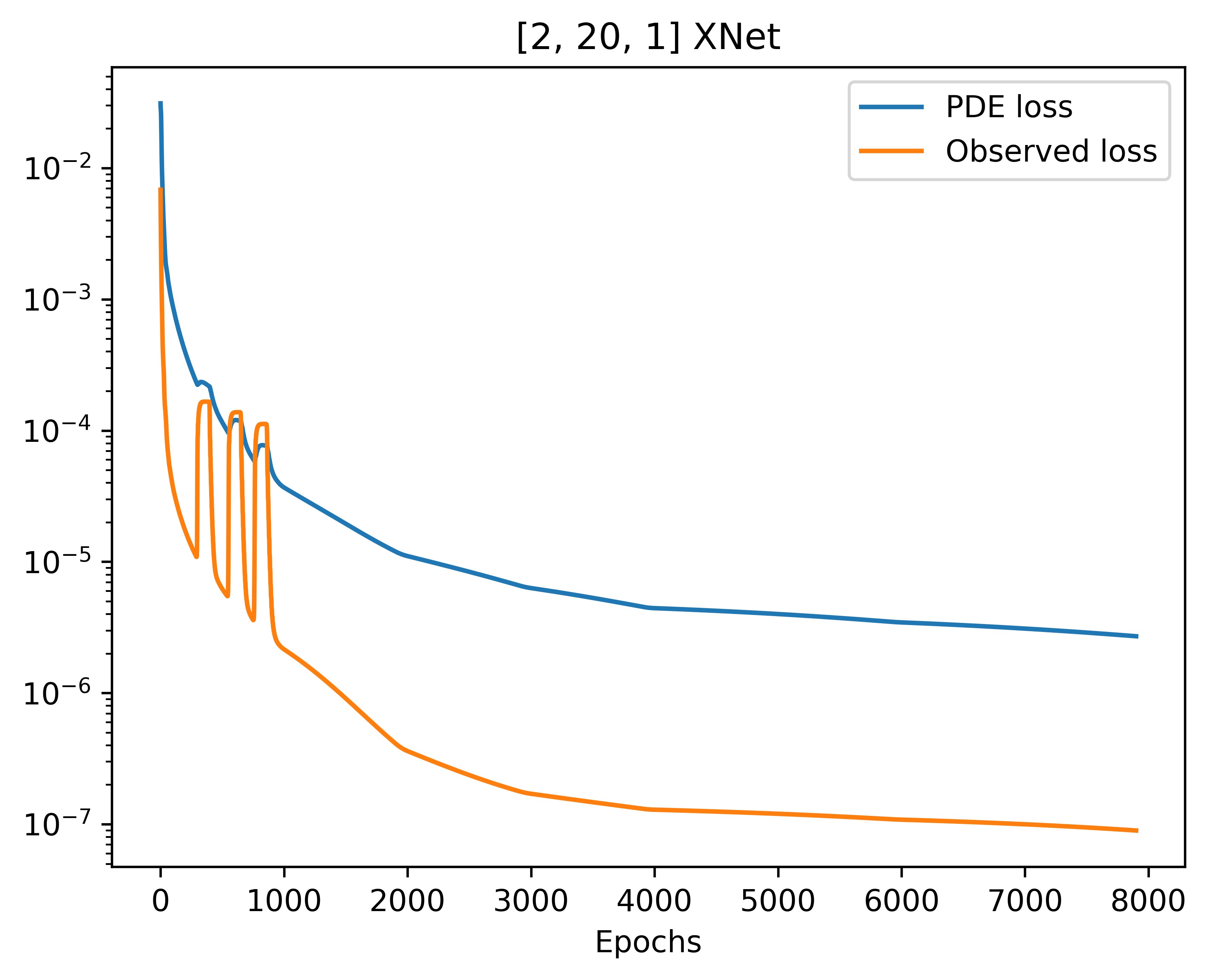}
    \end{minipage}
    \hfill
    \begin{minipage}[b]{0.23\textwidth}
        \centering
        \includegraphics[width=\textwidth]{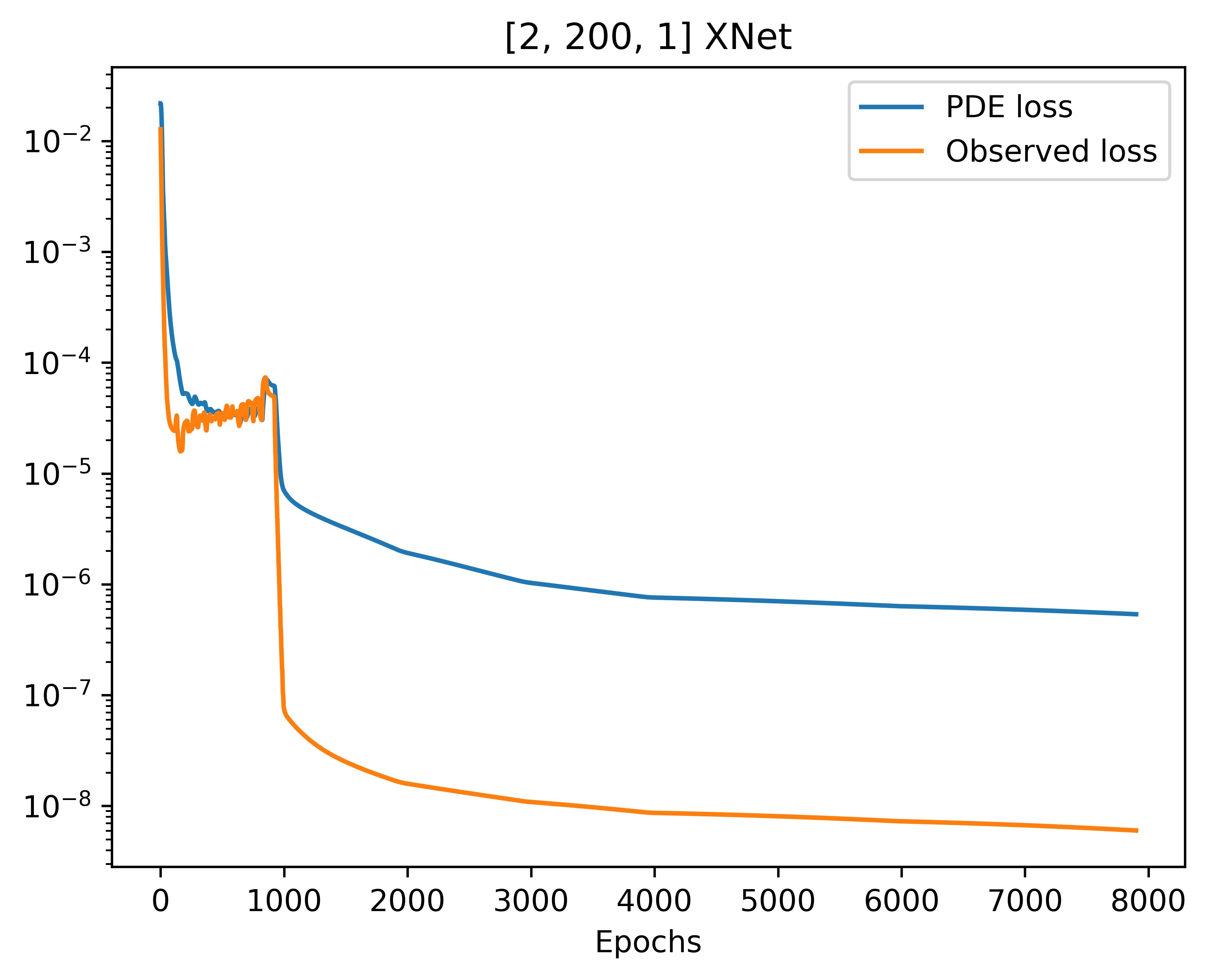}
    \end{minipage}
    \caption{Comparison of KAN, PINN and XNet approximations on PDE loss.}
    \label{fig:comparison_NS_loss}
\end{figure}

For solving the Heat equation under the framework of PINN, the loss function plots of various models (see Figure \ref{fig:comparison_NS_loss}).

\subsection{Poisson equation}\label{ref:exp_poisson}
We aim to solve a 2D Poisson equation $\nabla^2 v(x,y) = f(x,y)$, $f(x,y)=-2\pi^2{\rm sin}(\pi x){\rm sin}(\pi y)$, with boundary condition $v(-1,y)=v(1,y)=v(x,-1)=v(x,1)=0$. The ground truth solution is $v(x,y)={\rm sin}(\pi x){\rm sin}(\pi y)$. 
We use the framework of physics-informed neural networks (PINNs) to solve this PDE, with the loss function given by
\begin{equation}
    \begin{aligned}
\mathrm{loss}_{\mathrm{pde}}&=\alpha\mathrm{loss}_i+\mathrm{loss}_b\\
    &:=\alpha\frac{1}{n_i}\sum_{i=1}^{n_i}|v_{xx}(z_i)+v_{yy}(z_i)-f(z_i)|^2\\
    &\;\;\;\;\;+\frac{1}{n_b}\sum_{i=1}^{n_b}v^2\:,
    \end{aligned}
\end{equation}

where we use loss$_i$to denote the interior loss, discretized and evaluated by a uniform sampling of $n_i$ points $z_i=(x_i,y_i)$ inside the domain, and similarly we use loss$_b$ to denote the boundary loss, discretized and evaluated by a uniform sampling of $n_b$ points on the boundary. $\alpha$ is the hyperparameter balancing the effect of the two terms.

\begin{figure}[h!]
    \centering
    \begin{minipage}[b]{0.23\textwidth}
        \centering
        \includegraphics[width=\textwidth]{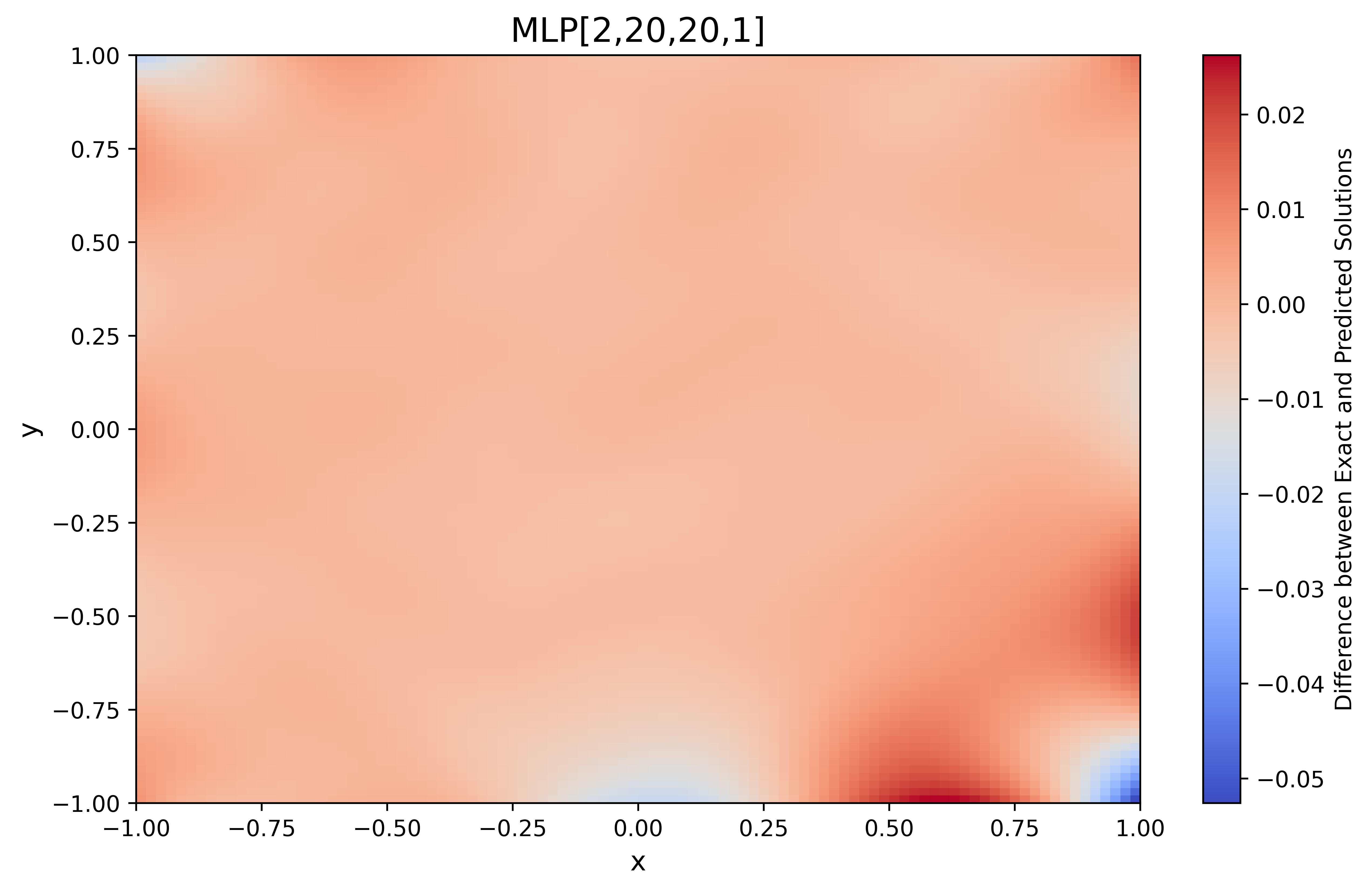}
    \end{minipage}
    \hfill
    \begin{minipage}[b]{0.23\textwidth}
        \centering
        \includegraphics[width=\textwidth]{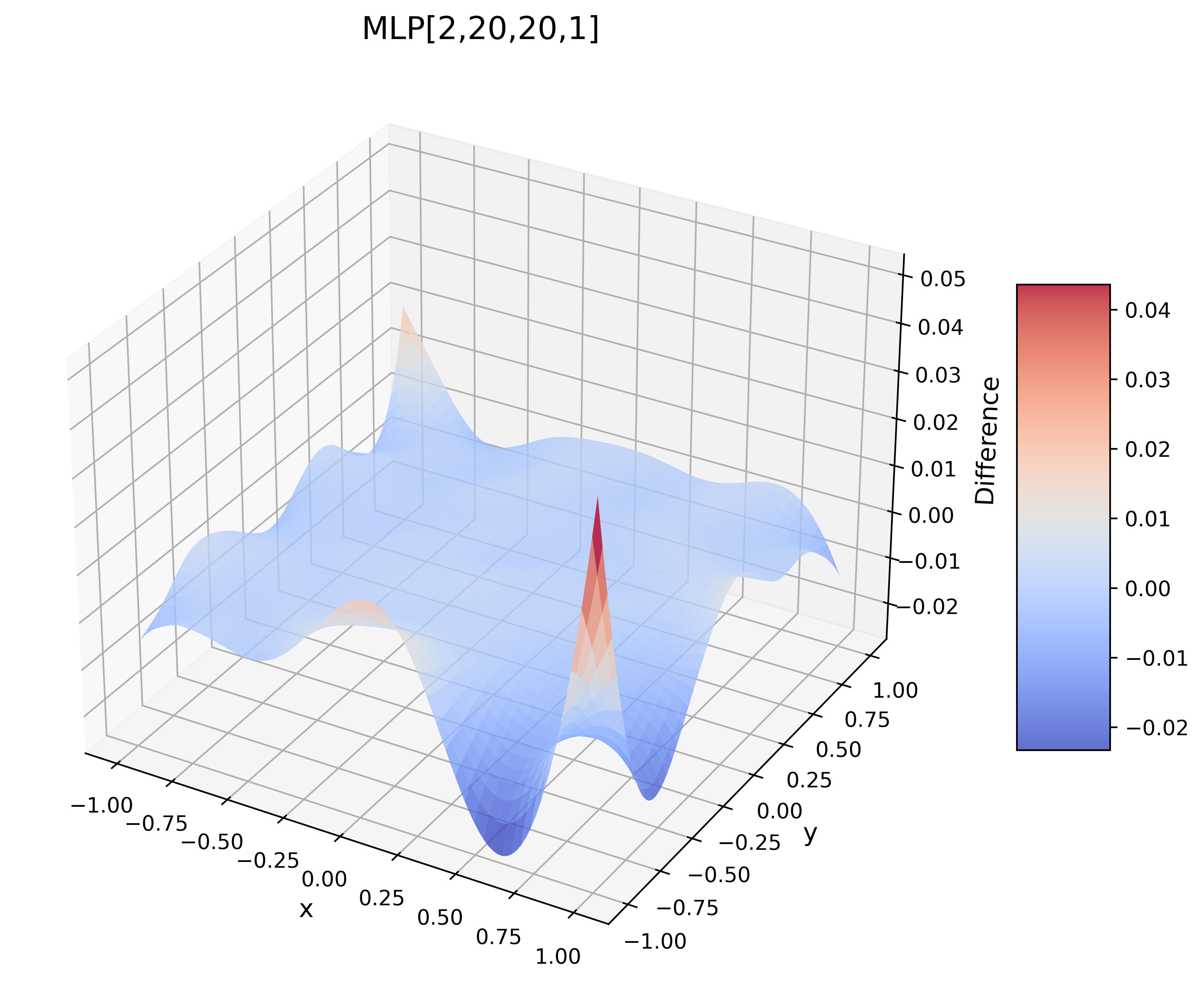}
    \end{minipage}
    \hfill
    \begin{minipage}[b]{0.23\textwidth}
        \centering
        \includegraphics[width=\textwidth]{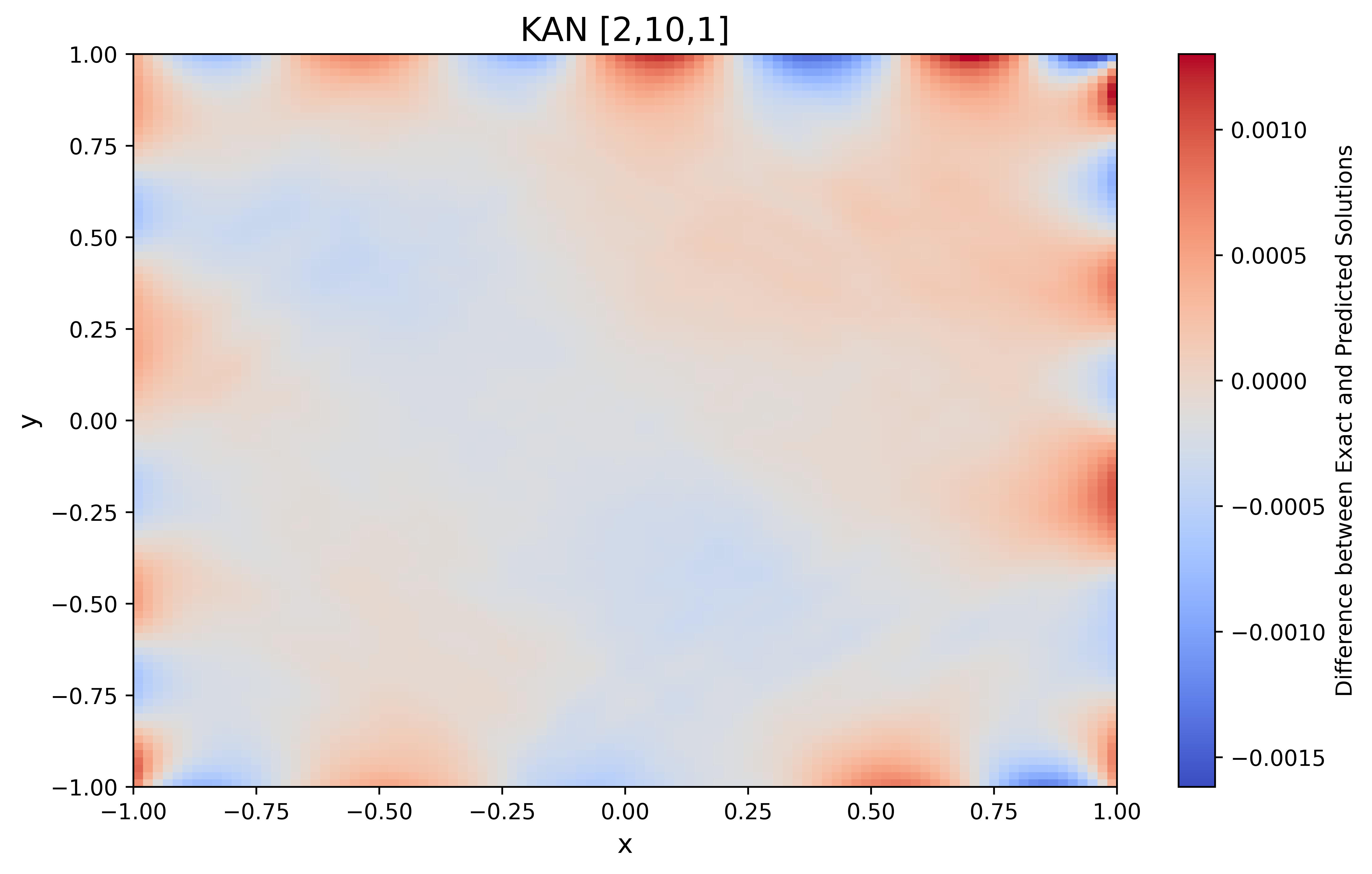}
    \end{minipage}
    \hfill
    \begin{minipage}[b]{0.23\textwidth}
        \centering
        \includegraphics[width=\textwidth]{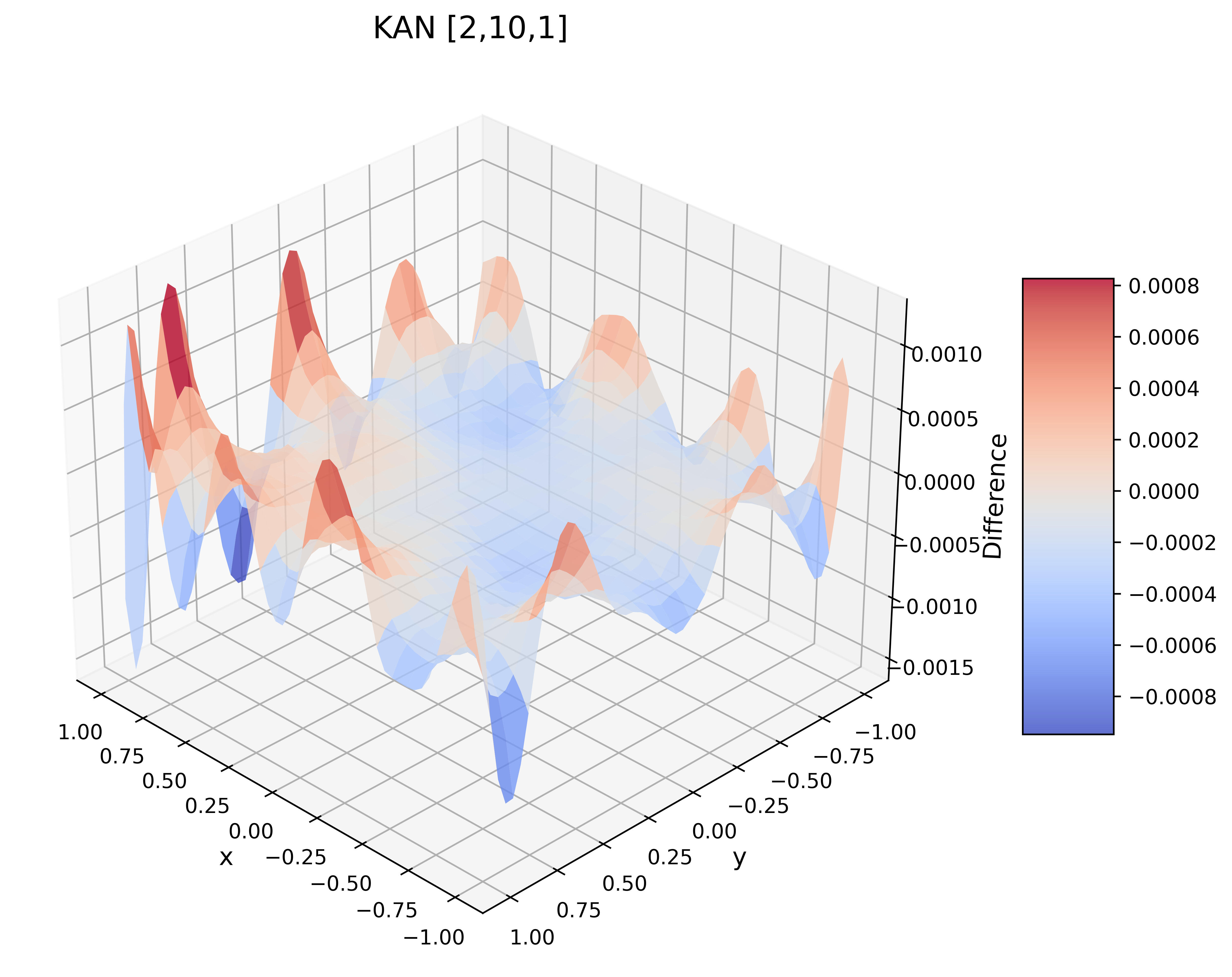}
    \end{minipage}
    \caption{PINN and KAN Performance}
\end{figure}

\begin{figure}[h!]
    \centering
    \begin{minipage}[b]{0.23\textwidth}
        \centering
        \includegraphics[width=\textwidth]{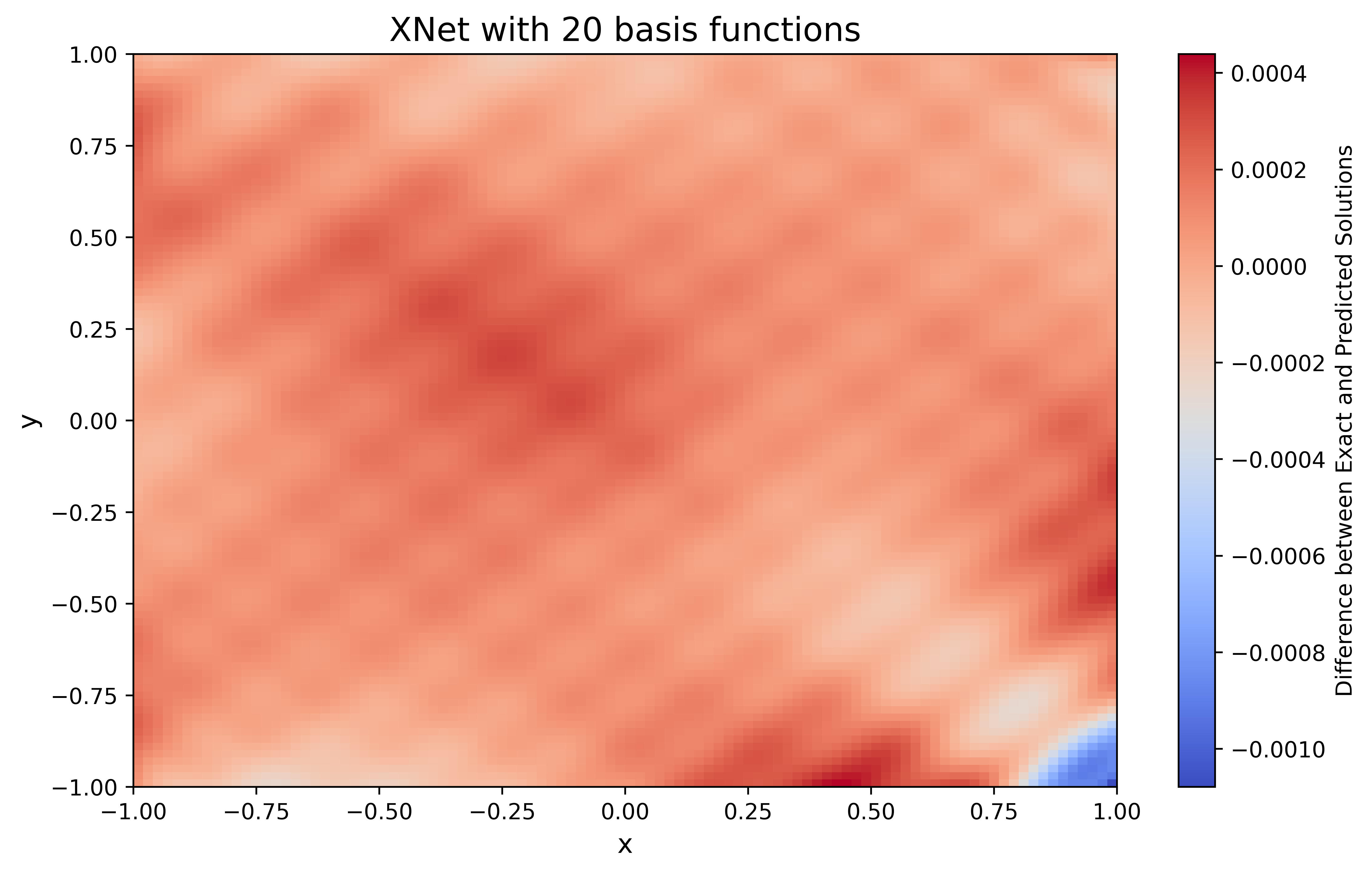}
    \end{minipage}
    \hfill
    \begin{minipage}[b]{0.23\textwidth}
        \centering
        \includegraphics[width=\textwidth]{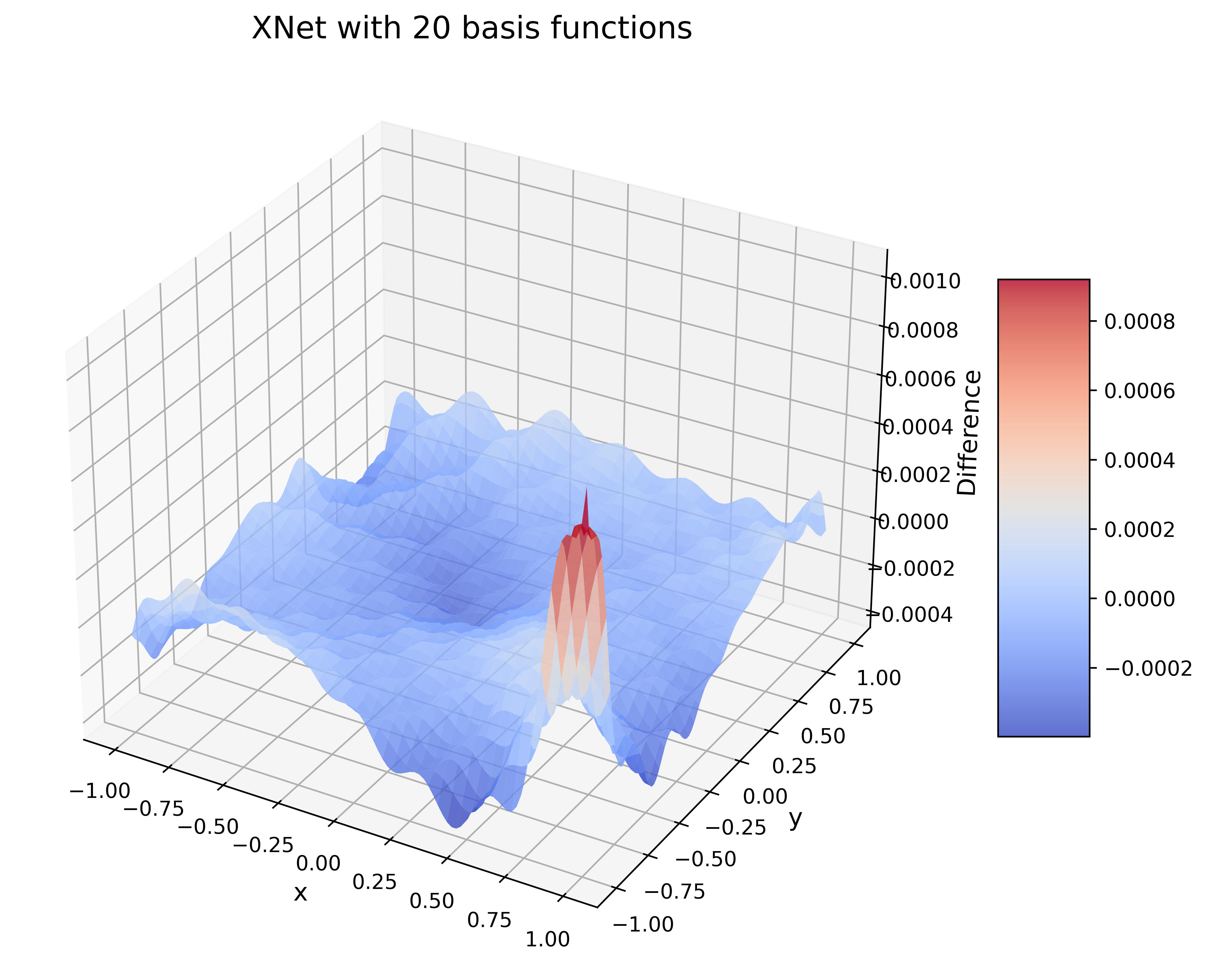}
    \end{minipage}
    \hfill
    \begin{minipage}[b]{0.23\textwidth}
        \centering
        \includegraphics[width=\textwidth]{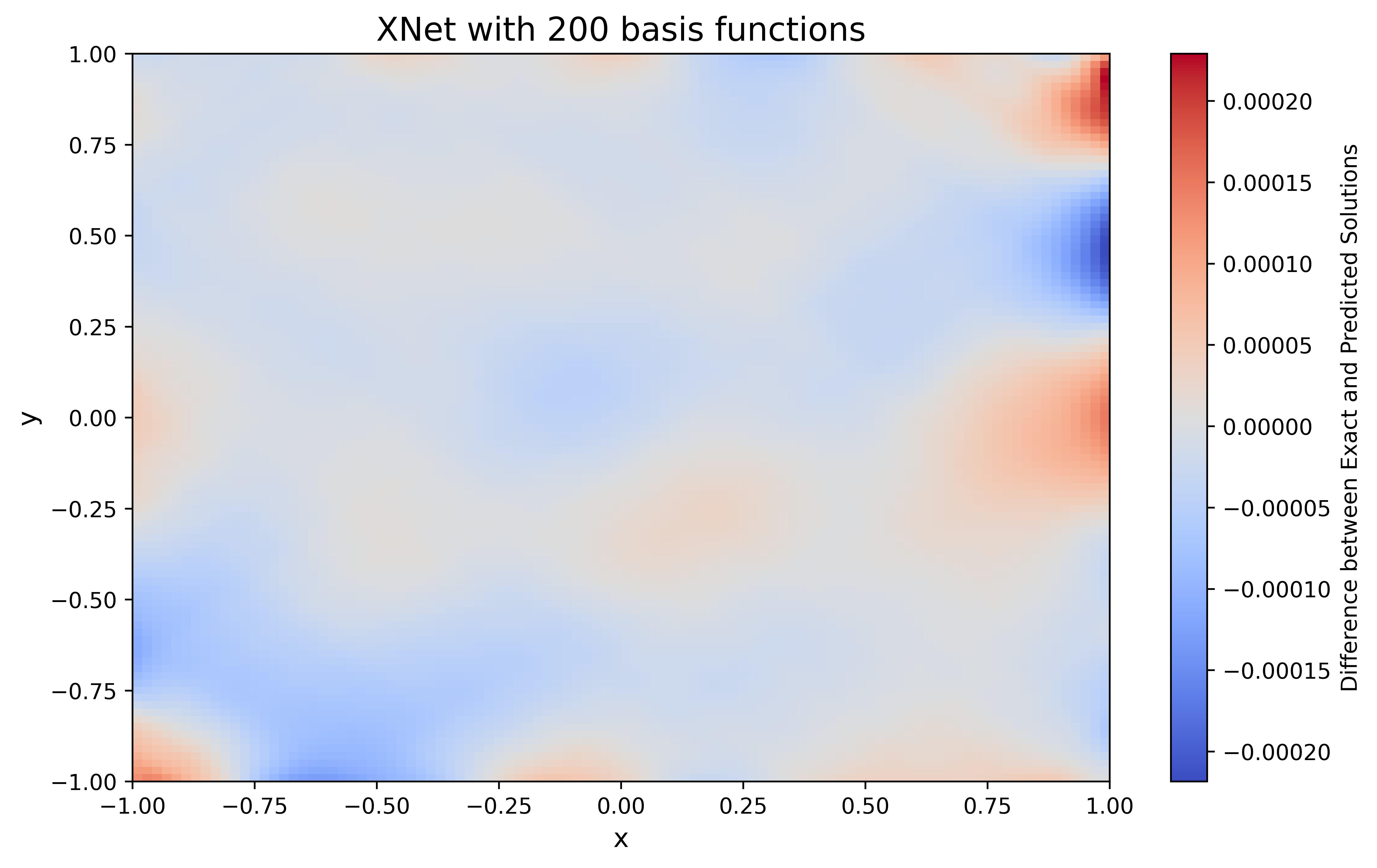}
    \end{minipage}
    \hfill
    \begin{minipage}[b]{0.23\textwidth}
        \centering
        \includegraphics[width=\textwidth]{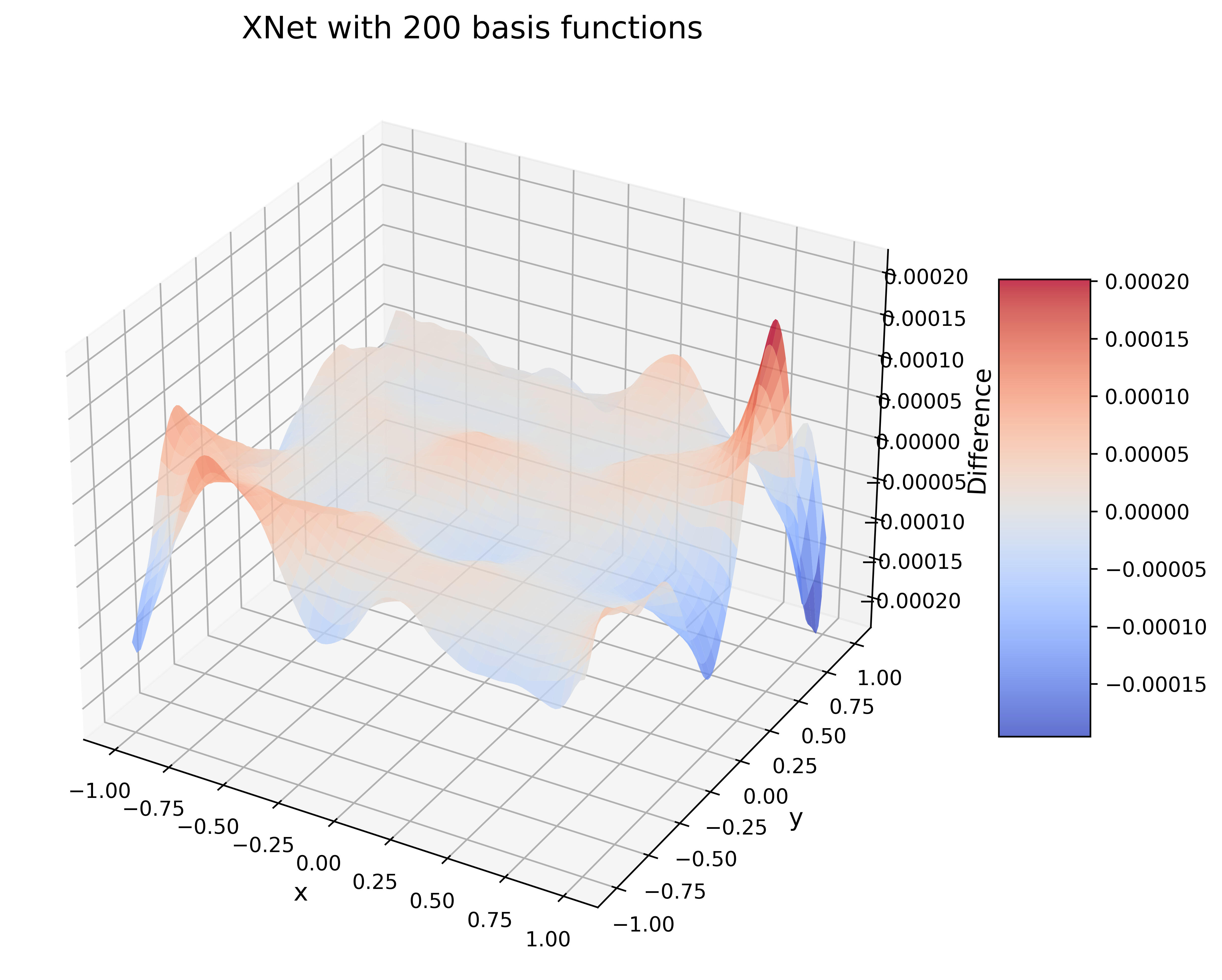}
    \end{minipage}
    \caption{XNet Performance}
\end{figure}

\begin{figure}[h!]
    \centering
    \begin{minipage}[b]{0.23\textwidth}
        \centering
        \includegraphics[width=\textwidth]{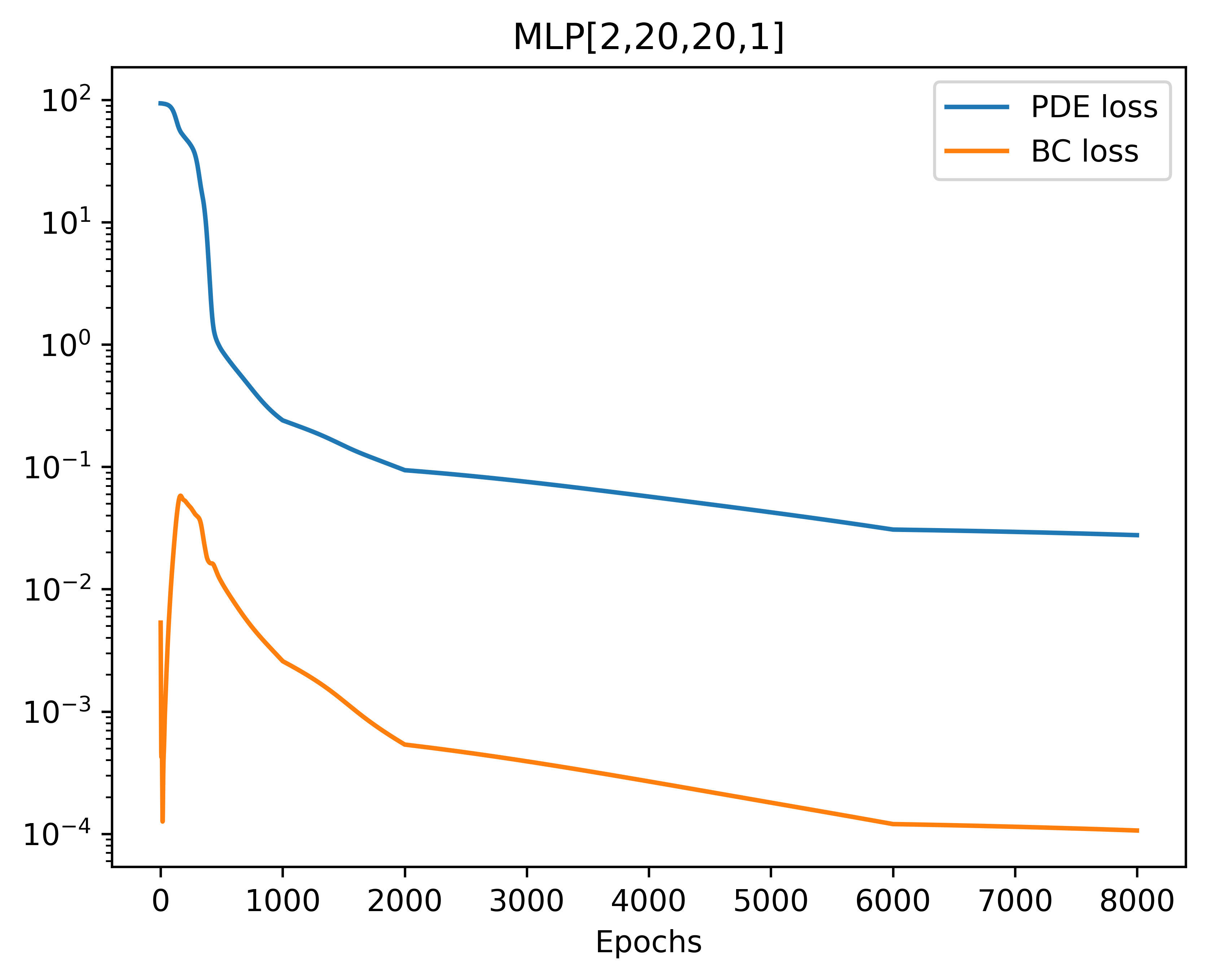}
    \end{minipage}
    \hfill
    \begin{minipage}[b]{0.23\textwidth}
        \centering
        \includegraphics[width=\textwidth]{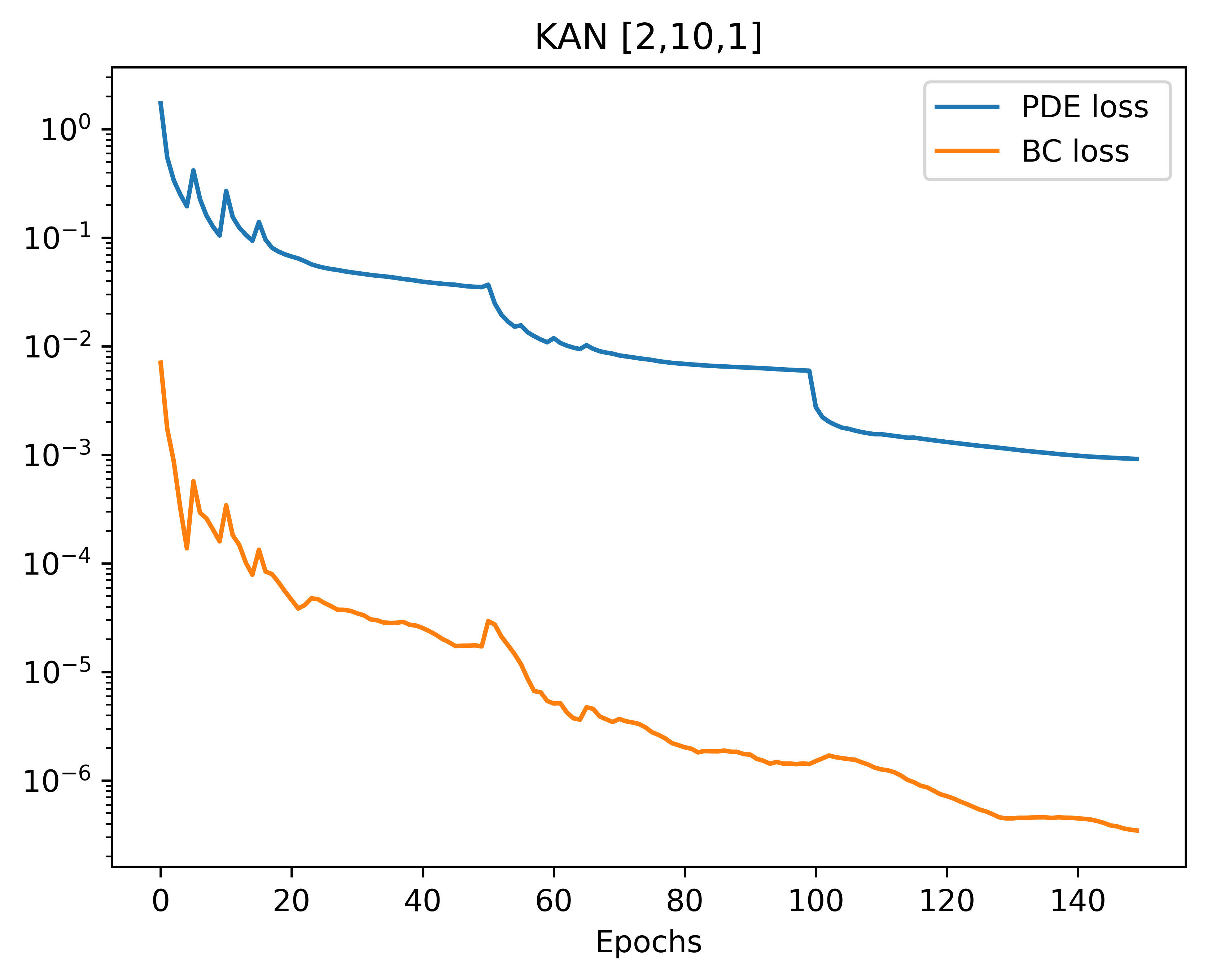}
    \end{minipage}
    \hfill
    \begin{minipage}[b]{0.23\textwidth}
        \centering
        \includegraphics[width=\textwidth]{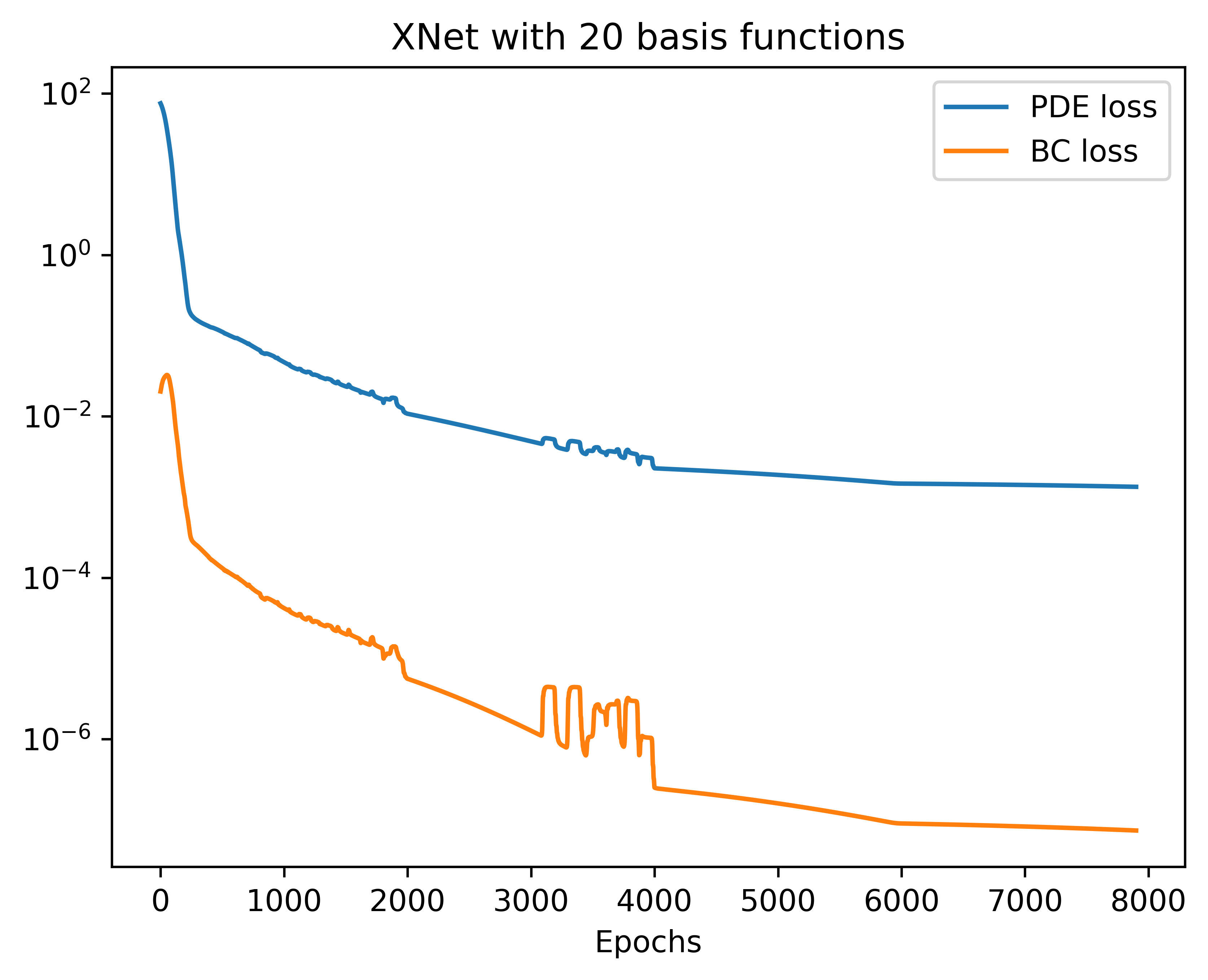}
    \end{minipage}
    \hfill
    \begin{minipage}[b]{0.23\textwidth}
        \centering
        \includegraphics[width=\textwidth]{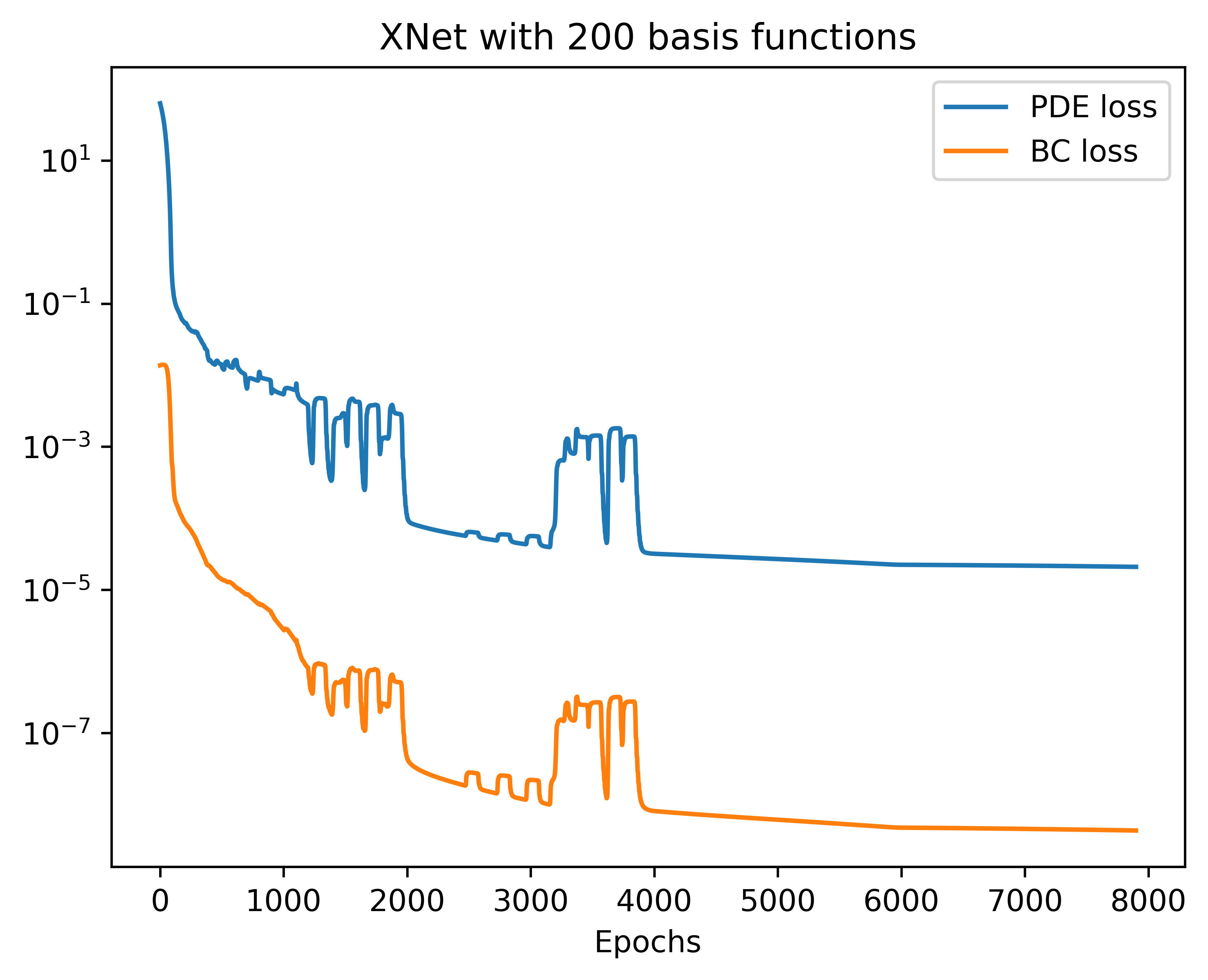}
    \end{minipage}
    \caption{Comparison of KAN, PINN and XNet approximations on PDE loss.}
    \label{fig:comparison}
\end{figure}

We compare the KAN, XNet and PINNs using the same hyperparameters $n_i=2500$, $n_b=200$, and $\alpha=0.01.$ We measured the error in the $L^2$ norm (MSE) and observed that XNet achieved a smaller error, requiring less computational time, as shown in Figure \ref{fig:comparison}. 
A width-200 XNet is 50 times more accurate and 2 times faster than a 2-Layer width-10 KAN;
a width-20 XNet is 3 times more accurate and 5 times faster than a 2-Layer width-10 KAN (see Table \ref{table:poison_compare}).
Therefore we speculate that the XNet might have the potential of serving as a good neural network representation for model reduction of PDEs.  In general, KANs and PINNs are good at representing different function classes of PDE solutions, which needs detailed future study to understand their respective boundaries.

\begin{table}[!ht]
\centering
\small 
\caption{Comparison of XNet and KAN on the Poisson equation.}
\label{table:poison_compare}
\resizebox{0.45\textwidth}{!}{
\begin{tabular}{ccccc} 
\toprule
\textbf{Metric} & \textbf{MSE} & \textbf{RMSE} & \textbf{MAE} & \textbf{Time (s)} \\
\midrule
\textbf{PINN [2,20,20,1]} & 1.7998e-05 & 4.2424e-03 & 2.3300e-03 & 48.9 \\ 
\textbf{XNet (20)} & 1.8651e-08 & 1.3657e-04 & 1.0511e-04 & 57.2 \\ 
\textbf{KAN [2,10,1]} & 5.7430e-08 & 2.3965e-04 & 1.8450e-04 & 286.3 \\ 
\textbf{XNet (200)} & 1.0937e-09 & 3.3071e-05 & 2.1711e-05 & 154.8 \\ 
\bottomrule
\end{tabular}}
\end{table}

\begin{figure}[h!]
    \centering
    \begin{minipage}[b]{0.5\textwidth}
        \centering
        \includegraphics[width=\linewidth]{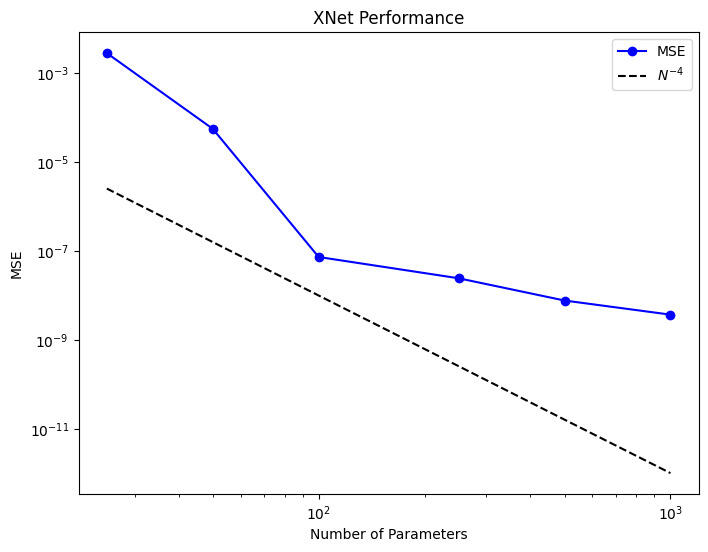}
    \end{minipage}\hfill
\caption{XNet Performance with Number of Parameters}
\label{fig:XNet_pde_params}
\end{figure}

\end{document}